\documentclass[10pt,twocolumn,letterpaper]{article}

\usepackage[pagenumbers]{cvpr} 

\definecolor{cvprblue}{rgb}{0.21,0.49,0.74}
\usepackage[pagebackref,breaklinks,colorlinks,citecolor=cvprblue]{hyperref}
\usepackage{graphicx}
\usepackage{booktabs}
\usepackage{dsfont}
\usepackage{amsmath, amssymb}
\usepackage{algorithm}
\usepackage{algpseudocode}
\usepackage{float}
\usepackage{booktabs} 
\usepackage{multirow} 
\usepackage[export]{adjustbox}
\usepackage{array}
\usepackage{comment}
\usepackage[normalem]{ulem}
\usepackage{wrapfig}
\usepackage{orcidlink}
\usepackage{xcolor}
\usepackage{colortbl}

\definecolor{cvprblue}{rgb}{0.21,0.49,0.74}
\definecolor{darkorange}{rgb}{1.0, 0.45, 0.0}
\definecolor{darkred}{rgb}{1, 0.7, 0.7}
\definecolor{midred}{rgb}{1, 0.85, 0.7} 
\definecolor{lightred}{rgb}{1, 1, 0.7} 

\title{Outlier-Robust Multi-Model Fitting
on Quantum Annealers}

\author{Saurabh Pandey$^{1,2}$ \quad\; Luca Magri$^{3}$ \quad\; Federica Arrigoni$^{3}$ \quad\; Vladislav Golyanik$^{1}$ 
\\
\vspace{10pt} 
$^1$MPI for Informatics, SIC \quad\; $^2$Saarland University \quad\; $^3$Politecnico di Milano
}

\begin{document}
\maketitle

\begin{abstract}
Multi-model fitting (MMF) presents a significant challenge in Computer Vision, particularly due to its combinatorial nature.
While recent advancements in quantum computing offer promise for addressing NP-hard problems, existing quantum-based approaches for model fitting are either limited to a single model or consider multi-model scenarios within outlier-free datasets.
This paper introduces a novel approach, the robust quantum multi-model fitting (R-QuMF) algorithm, designed to handle outliers effectively. 
Our method leverages the intrinsic capabilities of quantum hardware to tackle combinatorial challenges inherent in MMF tasks, and it does not require prior knowledge of the exact number of models, thereby enhancing its practical applicability. 
By formulating the problem as a maximum set coverage task for adiabatic quantum computers (AQC), R-QuMF outperforms existing quantum techniques, demonstrating superior performance across various synthetic and real-world 3D datasets. 
Our findings underscore the potential of quantum computing in addressing the complexities of MMF, especially in real-world scenarios with noisy and outlier-prone 
data\footnote{Project page: \url{https://4dqv.mpi-inf.mpg.de/RQMMF/}}. 
\end{abstract}


\section{Introduction}
\label{sec:intro}

\begin{figure}[htbp] 
  \centering 
  \includegraphics[width=\columnwidth] 
  {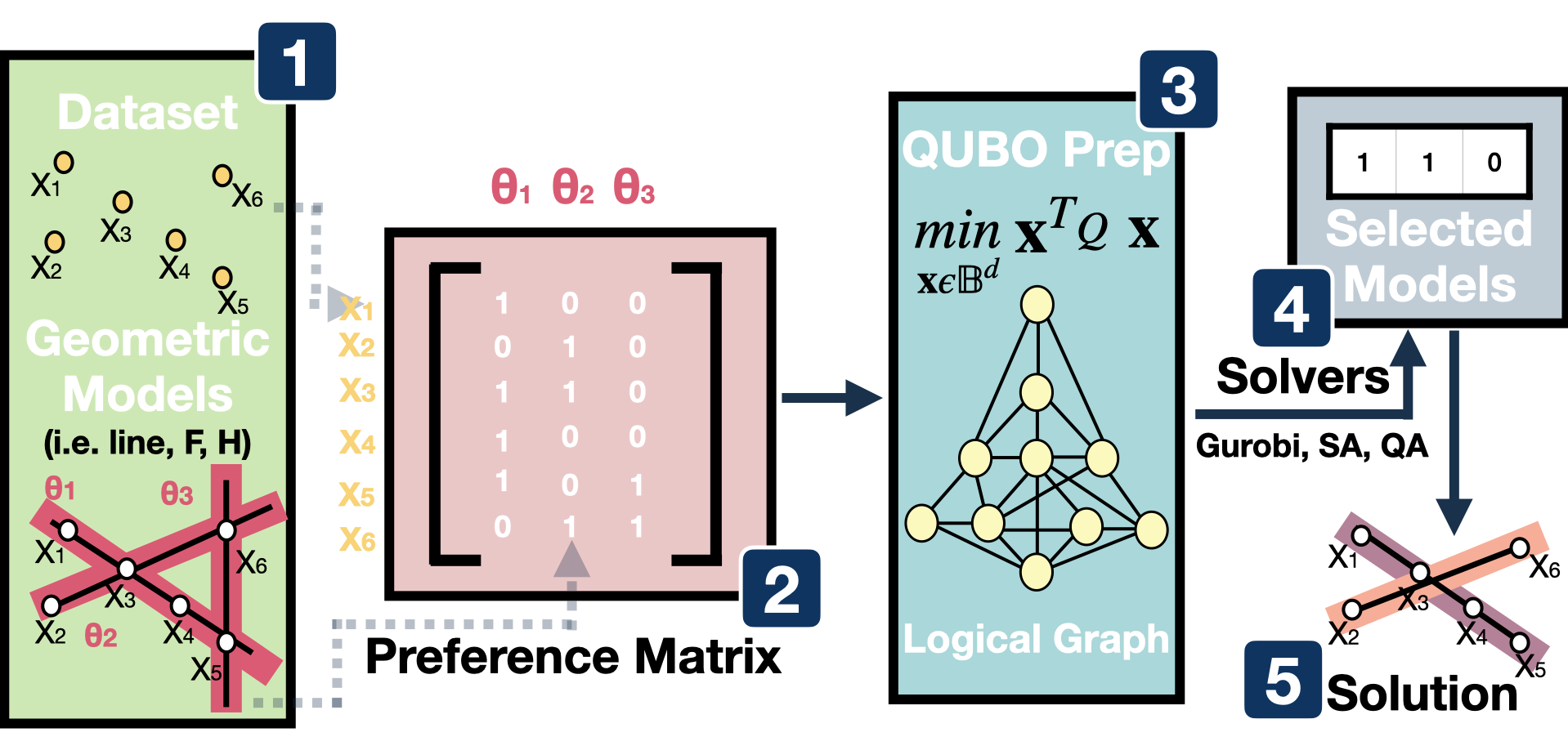} 
  \caption{\textbf{Overview of our R-QuMF}, a multi-model fitting approach that is robust to outliers and admissible to modern quantum annealers. We first sample models that along with the data define the preference matrix $P$. Next, a QUBO problem is prepared that can be minimised by quantum annealing (after a minor embedding of the logical problem on quantum hardware) or other solvers. Finally, the best solution is selected. R-QuMF outperforms previous quantum-admissible model fitting approaches. 
  } 
  \label{fig:method_overview}
\end{figure}

Model fitting is a fundamental and challenging problem in computer vision, with applications such as 3D reconstruction, scene layout estimation, motion segmentation, and image stitching. 
Its objective 
is to explain input data (e.g., 2D or 3D point sets) using a non-redundant number of parametric models. However, challenges arise with multiple models, whose exact number is typically unknown, and the need for robustness against outliers. These issues are particularly critical in homography and fundamental matrix estimations, where errors can significantly impact downstream tasks. Ultimately, multi-model fitting is an ill-posed problem with a combinatorial nature, where data clustering and model estimation must be solved simultaneously.

Multi-model fitting (MMF) has been actively researched during the last decades, following many different principles (\emph{e.g.}, optimisation and clustering-based approaches, see Sec.~\ref{sec:Related_Work}).
One of the latest and newest research directions in the field focuses on adopting quantum computational paradigms, either gate-based or quantum annealing \cite{Chin2020,Doan2022,Farina2023}. 
While gate-based quantum machines are universal in the sense they 
complement classical computers with a set of additional operations, quantum annealers can be thought of as samplers of a specific type of optimisation problems, \emph{i.e.}, quadratic unconstrained binary optimisation (QUBO) objectives. 
The latter has recently gained a lot of attention in the community since many computer vision tasks can be rephrased as a QUBO, including matching problems \cite{seelbach20quantum,SeelbachBenkner2021,BirdalGolyanikAl21,bhatia2023CCuantuMM,meli2025qucoop}, object detection \cite{LiGhosh2020}, multi-object tracking \cite{Zaech_2022_CVPR}, motion segmentation \cite{arrigoni2022quantum} and neural network weight optimisation \cite{Sasdelli2021QuantumAF, Krahn2024}. The main motivation for using quantum computers lies in their promise to accelerate the solution search for combinatorial optimisation problems while returning globally optimal solutions with a certain non-zero probability \cite{Farhi2001}. Advantages in adopting a quantum approach have also been shown for non-combinatorial problems like point-set registration \cite{golyanik2020quantum, Meli_2022_CVPR}. 

In the context of model fitting, Chin \textit{et al.}~\cite{Chin2020} introduced a method based on gate-based quantum hardware, while other works rely on the quantum annealing paradigm~\cite{Doan2022, Farina2023}.  
These methods were shown to provide improvements compared to classical methods due to the quantum effects both from the theoretical and practical perspectives; they are compatible with current and upcoming generations of quantum hardware. 
 
\begin{table}

    \centering
    	\resizebox{0.7\columnwidth}{!}{
    \begin{tabular}{l|c|c}
        \toprule
        \textbf{Method} & \textbf{Multi-Model} & \textbf{Outlier-Robust} \\
        \midrule
        HQC-RF \cite{Doan2022}  & $\times$ & \checkmark \\
        QuMF \cite{Farina2023}  & \checkmark & $\times$ \\
        Ours & \checkmark & \checkmark \\
        \bottomrule   
    \end{tabular}
    }
    \caption{Comparison of method characteristics.}
    \label{tab:method_comparison}
\end{table}
 
Among these approaches, \cite{Chin2020,Doan2022} address the case of a single model. The QuMF method of Farina \textit{et al.}~\cite{Farina2023} for quantum annealers, instead, not only outperforms previous quantum single-model fitting approaches but also supports multiple models and competes with classical state-of-the-art. 
However, the main drawback of QuMF \cite{Farina2023} is that it assumes outlier-free data, which is unrealistic in most practical scenarios. The naive way to extend such an approach to managing outliers is by post-processing, \emph{i.e.}, only the $k$ largest models are selected at the end among all candidate models obtained by random sampling (with $k$ equal to the true number of models explaining the data). Besides requiring the knowledge of $k$, or equivalent ancillary information, this approach is sub-optimal and prone to inaccuracies, as demonstrated by our experiments on standard datasets. 

This paper addresses the challenge of outlier handling in multi-model fitting and tailors for quantum annealers a new multi-model fitting approach, which we call Robust Quantum Multi-Model Fitting (R-QuMF); see Fig.~\ref{fig:method_overview}. 
In contrast to previous quantum work \cite{Farina2023}, it accounts for outliers explicitly in the formulation, resulting in a more general approach while exhibiting superior results on real data. 
Another advantage of our method is that it does not require any prior information about the optimal number of models explaining the data, which is convenient in practical applications. More details on the differences with respect to previous quantum papers are reported in Tab.~\ref{tab:method_comparison}.
To summarise, the contributions of this paper are two-folds: 
\begin{itemize} 
\item[i)] R-QuMF, a new approach for outlier-robust fitting of multiple models; 
\item[ii)] A formulation compatible with quantum annealers that accounts for outliers explicitly: Our method does not need any post-processing steps or a number of optimal models in advance as input, which are highly advantageous properties in practice.
\end{itemize}

We also apply to R-QuMF the decomposition principle similar to De-QuMF~\cite{Farina2023}. It addresses the limitations of current AQC hardware by iteratively decomposing the original large problem into smaller QUBO sub-problems until the final sub-problem selects the solutions among the most promising models. Our approach significantly outperforms previous quantum techniques in the experiments with various multi-model fitting scenarios with different outlier ratios such as geometric model fitting, homography estimation and fundamental matrix estimation. The source code of our method for all solver versions (including D-Wave and demo examples) will be made available.


\section{Related Work}
\label{sec:Related_Work}

\paragraph{Classical Approaches.} 
Multi-model fitting has been addressed since the 1960s, with effective techniques ranging from the Hough transform \cite{XuOjaAl90} to more recent approaches based either on clustering or optimising an objective function. 
Clustering-based methods \cite{ToldoFusiello08,MagriFusiello14,ChinWangAl09,MagriFusiello17,TepperSapiro17,AgarwalJongowooAl05,Govindu05, JainGovindu13,ZassShashua05,PurkaitChinAl14,MagriLeveniAl21, WangGupbao18,BarathRozumnyiAl23} focus on data segmentation and offer procedural, easy-to-implement solutions that produce promising results in most cases. However, hard clustering of data does not always produce optimal results when models overlap. On the contrary, optimisation methods prioritise the refinement of a precise objective function, offering a quantitative measure to assess the quality of the derived solution. 
The most common objective functions are typically based on consensus, \emph{i.e.}, they aim at maximising the number of inliers of each model, so optimisation-based methods \cite{VincentLaganiere01,ZulianiKenneyAl05,Magri2016,IsackBoykov12,BarathMatas17,BarathMatas19} can be considered as sophisticated extensions of the popular RanSaC paradigm \cite{FischlerBolles1981} to the case of multiple models. 

In this respect, the classical work that is mostly related to our approach is RanSaCov \cite{Magri2016}, which casts multi-model fitting as a coverage problem. Given a collection of models with their consensus sets, RanSaCov extracts either the minimum number of models that explain all the points (set cover formulation) or, if the number $k$ of the sought models is known in advance, it selects the $k$ models that explain most of the data (maximum coverage formulation). The latter is effective in dealing with data contaminated by outliers that can be recognised as uncovered points. 

{We borrow from RanSaCov the maximum coverage formulation. However, }
RanSaCov \cite{Magri2016} solves the coverage problems via integer linear programming and branch and bound, hence, it either resorts to approximations or falls back to enumerating all candidate solutions. 
Instead, our approach exploits quantum effects to optimise the objective directly in the space of qubits, where global optimality is expected with high probability after multiple anneals. 
{In addition, we directly minimize the number of models as done in several traditional MMF frameworks \cite{BarathMatas17, BarathMatas19, IsackBoykov12}, without requiring the knowledge of the true number of models in advance. Contrary to RanSaCov, we do not deal natively with intersecting models, but inliers belonging to multiple models can be identified after the models have been extracted by inspecting the point-model residuals}.

\paragraph{Approaches Compatible with Quantum Hardware.} 
While many classical methods have been developed, the first methods based on quantum computing have only recently attempted to exploit the capabilities of quantum hardware to tackle the combinatorial nature of the problem. 
The quantum solutions presented so far do not address the multi-model fitting problem under outliers. 
Moreover, other approaches---both theoretical and practical---start to address closely related problems such as linear regression \cite{Schuld2016, DatePotok2021}, clustering and segmentation \cite{Willsch2020, arrigoni2022quantum} to capitalise on the advantages of quantum computing.
The first attempts to address model fitting with the help of quantum hardware concentrate on single-model fitting \cite{Chin2020,Doan2022}. 
Chin \textit{et al.}~and Yang \textit{et al.}~\cite{Chin2020, Yang2024} introduced a single-model fitting method based on gate-based quantum hardware, while Doan \textit{et al.}~\cite{Doan2022} rely on quantum annealing. 
Although both Doan \textit{et al.}~\cite{Doan2022} and our method are based on linear programming, their approaches diverge significantly. Doan \textit{et al.}~\cite{Doan2022} employ a hypergraph formalism that relies on multiple QUBOs within an iterative framework, while our method is more streamlined, using a single QUBO. Additionally, while they focus on single-model fitting, our approach extends to multi-model fitting. 

Farina \textit{et al.}~\cite{Farina2023} take this further with their QuMF method for quantum annealers, which considers multiple models and achieves results on par with classical state-of-the-art techniques. However, QuMF is primarily limited by its reliance on outlier-free data. While post-processing can improve outlier robustness, it usually requires prior knowledge of the number $k$ of models and often leads to sub-optimal results, as evidenced by our experiments.


\section{Background on Quantum Annealers}
\label{sec:Background} 

Modern quantum annealers (QAs) can sample Quadratic Unconstrained Binary Optimisation (QUBO) problems, which in a general form can be written as 
\begin{equation} 
    \arg\,\min_{\mathbf{y}\epsilon \mathbb{B}^{d}} \quad \mathbf{y}^{T} Q \mathbf{y} + \mathbf{s}^{T}\mathbf{y} + \sum_i \lambda_i {||A_i\mathbf{y} - \mathbf{b}_i||}^{2}_{2}, \label{eq2} 
\end{equation} 
where $\mathbf{y} \in \mathbb{B}^{d}$ is a vector of $d$ binary variables, $Q \in \mathbb{R}^{d\times{d}}$ is a real symmetric matrix, $A_i \in \mathbb{R}^{d\times{d}}$ are real matrices, $\mathbf{s}, \mathbf{\mathbf{b}}_i \in \mathbb{R}^{d}$, and $\lambda_i$ are scalar weights. 
The terms under the $\ell_2$-norm are rectifiers expressing soft linear constraints. 
They preserve the QUBO problem type since Eq.~\eqref{eq2} can be written without constraints as follows: 
\begin{equation} 
    \arg\,\min_{\mathbf{y}\epsilon \mathbb{B}^{d}} \quad 
    \mathbf{y}^{T} \widetilde{Q} \mathbf{y} + \mathbf{\tilde{s}}^{T}\mathbf{y} \label{eq3}, 
\end{equation} 
with $\widetilde{Q} = Q + \sum_i A_i^T A_i$ and $\mathbf{\tilde{s}} = \mathbf{s} - 2 \sum_i A_i \lambda_i \mathbf{b}_i$. 
Eq.~\eqref{eq3} is the combinatorial QUBO form admissible to modern QAs; 
the elements of $\widetilde{Q}$ and $\mathbf{\tilde{s}}$ along with the number of binary variables to be optimised have to be provided. 
During quantum annealing, $\mathbf{y}_i$ are modelled as qubits weighted by $\mathbf{\tilde{s}}_i$ with the strength of the mutual influence defined by $\widetilde{Q}$. 
The optimisation takes place in the $2^d$-dimensional Hilbert space and involves quantum-mechanical effects of qubit superposition, entanglement and quantum tunnelling. 

We will call the connectivity pattern between the binary variables in Eq.~\eqref{eq3} the \textit{logical problem graph} (i.e.~in which qubits are represented by vertices and edges between the vertices are present if $Q_{i, j}$ entries are non-zero). 
Since direct interactions between only a small subset of all possible $d(d-1)/2$ pairs of binary variables are enabled by the hardware, mapping the logical problem graph into the physical hardware is necessary for most QUBO problems. 
This mapping is called \textit{minor embedding}, as the logical problem graph is interpreted as a graph minor of a larger graph, i.e. hardware graph of physical qubits. 
This means the hardware graph contains qubit chains representing a single logical qubit from the logical problem graph.


\section{The Proposed R-QuMF Method}
\label{sec:Method} 

This section presents our robust quantum multi-model fitting (R-QuMF) approach and details of its implementation. The method can be summarised as in Fig.~\ref{fig:method_overview}: Steps 1 and 2 are described in Secs.~\ref{ssec:Preliminaries} and \ref{ssec:Objective}. 
Steps 3 and 5 correspond to Sec.~\ref{ssec:ourQUBO}, whereas details of Step 4 (implementation/solvers) are in Sec.~\ref{ssec:Implementation}. 
\subsection{Preliminaries, Definitions and Notations}\label{ssec:Preliminaries} 
\label{subsec:problem} 
We frame the multi-model fitting problem in the presence of outliers as follows. We are given in input a set of points $X = (x_{1}, x_{2}, ..., x_{n})$ and a collection of models $\Theta = (\theta_{1}, \theta_{2}, ..., \theta_{m})$ generated through random sampling akin to those of the RANSAC algorithm \cite{FischlerBolles1981}. 
The desidered output is a subset $\{\theta_{i1},\ldots \theta_{ik}\}\subset \Theta$  of \emph{non-redundant} models that  explain the data. Having non-redundant models is a form of regularisation to make the ill-posed multi-model fitting problem tractable. Echoing Occam Razors' principle, we favour the interpretation of the data that minimizes the number of models required.
Moreover, we assume that: \emph{i}) $X$ can be corrupted by  high outlier percentages (up to $50\%$), and that \emph{ii}) the number $k$ of true models explaining the underlying data is unknown. 
The above-mentioned assumptions are very desirable in practical applications.

The problem can be equivalently formulated in terms of a preference-consensus matrix defined as 
    \begin{equation}
    \label{eq7}
        {P}[i,j]=
        \begin{cases}
            1 & \text{if } \operatorname{err}(x_{i}, \theta_{j}) < \epsilon, \\
            0 & \text{otherwise}, 
        \end{cases}
    \end{equation} 
where $P \in \mathbb{B}^{{n}\times{m}}$ is a $n\times m$ binary matrix; $n$ and $m$ being the number of points and models, respectively. 
The $i$-th data point $x_i$ is assigned to the $j$-th sampled model $\theta_j$ if its residual is below an inlier threshold $\epsilon$. 
Operator $\operatorname{error(\cdot, \cdot)}$ quantifies the point-to-model distance. 
Following \cite{Magri2016}, 
the rows of $P$ can be interpreted as preference sets, while
the columns of $P$---denoted by $S_1, \dots, S_m$---represent consensus sets (w.r.t.~$\epsilon$).  MMF reduces to selecting from  $P$ the columns that correspond to the sought models $\{\theta_{i1},\ldots \theta_{ik}\}$.

\subsection{Revisiting Maximum-Set Coverage Objective}\label{ssec:Objective} 

In order to gain robustness against outliers,  we generalise the QUBO formulation presented in QuMF \cite{Farina2023} approach -- which, in turn, is based on the Set Cover formulation presented in \cite{Magri2016}.
Specifically, we revisit the maximum-set coverage (MSC) objective for MMF \cite{Magri2016}. 
The MSC task is to select at most $k$ subsets from $\Theta$ such that the coverage of the data points contained in the set $X$ is maximum (or as complete as possible); all the uncovered points are considered outliers. 
Intuitively, with reference to the matrix $P$, we want to select $k$ columns that explain most of the points.
 
Let us introduce $n$ binary variables $y_{1}, \dots, y_n$ such that: $y_{i} = 1 $ if $x_{i}$ is covered by one of the $\theta_{j}$, or, in other terms, it is part of the selected subsets (i.e.~the point is an inlier); $y_{i} = 0 $ otherwise (i.e.~the point is an outlier).
Let us consider additional auxiliary variables $z_1, \dots, z_m$ such that: $z_j=1$ if model $\theta_j$ is selected; $z_j=0$ otherwise.
Using this notation, the MSC problem is formulated as an integer linear programming:
 \begin{equation}
    \max \sum_{i=1}^{n} y_{i} \label{eq8}
   \quad s.t. \ \sum_{j=1}^{m} z_{j} \leq k, \
   \sum_{j : S_j \ni x_i}  z_j \geq y_i \quad \forall x_i \in X  
 \end{equation}
where both $ y_i \in \{0, 1\} $ and $ z_j \in \{0, 1\} $. 
In this context, the first constraint imposes that at most $k$ models are selected (with $k$ known in advance); the second constraint ensures that, if $y_i =1 $, then at least one set $\theta_j$ containing $x_i$ must be chosen. 
Recall that $n$ corresponds to the cardinality of the set $X$ and $m$ denotes the number of candidate models. Note that, within this formulation, uncovered points are considered outliers. Therefore, outliers and uncovered points are used interchangeably and no artificial models for outliers need to be introduced.

The combinatorial nature of the problem makes it a suitable choice for designing a QUBO method compatible with a quantum annealer. To accomplish such a task, we make some changes with respect to the MSC objective in \eqref{eq8}. First, we replace the inequality in the second constraint with an equality, therefore looking for \emph{disjoint models}.
Secondly, instead of demanding the approach to select $k$ models, we opt for directly minimising the number of selected models. This choice is motivated by the fact that, in many practical situations, assuming known $k$ is restrictive. 
Therefore our final objective, converted into a minimisation problem, is:
 \begin{equation}
    \min -\sum_{i=1}^{n} y_{i} + \lambda_1 \sum_{j=1}^{m} z_{j}
   \quad s.t.  \ 
   \sum_{j : S_j \ni x_i}  z_j = y_i \quad \forall x_i \in X  
   \label{eq_MSC_sum}
 \end{equation}
where $\lambda_1$ is a regularisation parameter.

\subsection{MSC Reformulated as QUBO}\label{ssec:ourQUBO} 

The first step to reformulate MSC as QUBO is to vectorise the 
objective in Eq.~\eqref{eq_MSC_sum} and convert the constraint into a matrix form. More precisely, we rewrite it as follows:
\begin{equation}
\min_{\mathbf{y}\in\mathbb{B}^{n}, \ \mathbf{z}\in\mathbb{B}^{m}} -\mathds{1}^{T}_{n}\mathbf{y} + \lambda_{1}(\mathds{1}^{T}_{m}\mathbf{z}) \\
\quad s.t \qquad P\mathbf{z} = \mathbf{y} \label{eq16}
\end{equation}
where $\mathds{1}$ denotes a vector of ones (whose length is given as a subscript), $P$ is the preference matrix of size $n\times{m}$,  $\mathbf{y}$ and $\mathbf{z}$ are binary vectors collecting the $y_i$ and $z_j$ variables, respectively. 
To simplify the formulation into a single variable optimisation problem, we can combine the two unknowns into a single variable $\textbf{w}$: 
\begin{equation}
\mathbf{w} = \begin{pmatrix}
 \mathbf{y}\\
 \mathbf{z}\label{eq17}
\end{pmatrix} \in \begin{Bmatrix}
 0,1 \\
\end{Bmatrix}^{n+m}, 
\end{equation}
and rewrite the main objective of \eqref{eq16} 
in terms of variable $\mathbf{w}$:
\begin{align}
   -\mathds{1}^{T}_{n}\mathbf{y} + \lambda_{1}(\mathds{1}^{T}_{m}\mathbf{z}) &= 
   [-\mathds{1}^{T}_{n}, \ \mathds{O}^{T}_{m}]\mathbf{w} + \lambda_{1}
[\mathds{O}^{T}_{n}, \ \mathds{1}^{T}_{m}]\mathbf{w} \nonumber \\
   &= 
   [ -\mathds{1}^{T}_{n}, \  \lambda_{1}\mathds{1}^{T}_{m}] \mathbf{w}
   \label{eq20}
\end{align}
where $\mathds{O}$ represents a vector of zeros, with its length indicated as a subscript. 
Furthermore, the constraint in \eqref{eq16} can also be reformulated in terms of the newly introduced variable $\mathbf{w}$ as follows: 
\begin{equation}
P \mathbf{z} - \mathbf{y} = \mathds{O}_{n}
\ \Leftrightarrow \ 
   [-\mathbb{I}_{n\times{n}}, \  P]\mathbf{w} = \mathds{O}_{n} \label{eq22}
\end{equation}
where $\mathbb{I}_{n\times{n}}$ is an identity matrix of size $n\times{n}$.

After incorporating the constraint from Eq.~\eqref{eq22} as a penalty term into equation \eqref{eq20} we obtain 
\begin{equation}
    \min_{\mathbf{w} \in \mathbb{B}^{n+m}} [ -\mathds{1}^{T}_{n}, \ \lambda_{1}\mathds{1}^{T}_{m}]\mathbf{w} + \lambda_{2}{||[-\mathbb{I}_{n\times{n}}, \  P]\mathbf{w}||}^{2}_{2}.  \label{eq23}
\end{equation}
Now, when comparing \eqref{eq23} with \eqref{eq2}, the following correspondences emerge: 
\begin{equation}
\begin{gathered}
   {Q} = 0, \quad  \mathbf{s}^T=[ -\mathds{1}^{T}_{n}, \ \lambda_{1}\mathds{1}^{T}_{m}], \\  A=[-\mathbb{I}_{n\times{n}}, \  P], \quad \mathbf{b} =\mathds{O}_{n}.  
\end{gathered}
\label{eq24}
\end{equation} 
Note that ${Q}$ equals zero matrix as 
there are no quadratic terms involved. 
Finally, we obtain the target QUBO objective admissible on quantum hardware that can be written in the form of 
Eq.~\eqref{eq3}: 
\begin{equation}
\min_{\mathbf{w}}\;\mathbf{w}^{T}\widetilde{Q}\mathbf{w} + \widetilde{\mathbf{s}}^T\mathbf{w},
\label{eq25}
\end{equation}
where
\begin{equation}
\begin{gathered}
    \mathbf{w} = \begin{pmatrix}
     \mathbf{y}\\
     \mathbf{z}
    \end{pmatrix} \in \begin{Bmatrix}
     0,1 \\
    \end{Bmatrix}^{n+m}, \\
    \widetilde{Q} = \lambda_{2}\begin{pmatrix}
     \mathds{I}_{n \times n}&  -P\\
     -P^{T}& P^{T}P \\
    \end{pmatrix},
    \widetilde{\mathbf{s}} = \begin{pmatrix}
     -\mathds{1}_{n}\\
     \lambda_{1}\mathds{1}_{m} 
     \end{pmatrix}. 
\end{gathered}
\label{eq26} 
\end{equation} 
The QUBO formulation in Eq.~\eqref{eq25} can be optimised by classical global optimisation algorithms such as simulated annealing, or sampled on a quantum annealer. Note that the number of unknowns scales linearly with the number of points and models. 
Finally, we point out that the decomposition principle introduced by Farina \textit{et al.} \cite{Farina2023} can be applied to our QUBO as well, to manage large-scale applications. The core principle is to iteratively break down the input problem into smaller (tractable) sub-problems and aggregate the respective results. 
Algorithm~\ref{algo:derqumf_algorithm} in our supplement provides a summary of our method.

\subsubsection{Selection of {\large $\lambda$}'s} 
The lambda parameters from Eq.~\eqref{eq26} are decisive for the performance of our method. To find suitable values, we use Tree-structured Parzen Estimator (TPE) \cite{bergstra2011algorithms} which employs a Bayesian optimisation strategy utilizing a probabilistic model to steer the search process towards hyperparameter configurations that are more likely to improve the performance metric of interest. 
This model-based approach contrasts sharply with exhaustive grid search, which operates without leveraging prior knowledge or outcomes of evaluations. 
TPE optimises by constructing and refining a probabilistic model based on past evaluation results, thereby smartly converging to optimal hyperparameters through sequential model fitting and utility-based sampling.

\subsection{Implementation Details}\label{ssec:Implementation}

Our QUBO objective can be optimized either on CPU using classical solvers, like Simulated Annealing (SA) \cite{Kirkpatrick1983} and Gurobi \cite{Gurobi2023}, or on QPU (Quantum Processing Unit) via Quantum Annealing (QA).  

For simulated annealing, we use D-Wave's neal package (version 0.6.0), and  we fix the number of samples for SA to $100$ (the same used in Farina \textit{et al.}~\cite{Farina2023}), and we adopt this configuration also for the competing methods. 
 
As regard Gurobi, we rely on version 10.0.3 under academic license with a time limit of $120$ seconds for both RQuMF and the decomposed version De-RQuMF. The same configuration was used for  QuMF  and  De-QuMF \cite{Farina2023}.

Experiments with Quantum Anneling are performed on D-Wave quantum annealer Advantage 5.4. We set the number of anneals to $5k$ for the one-shot version and $2.5k$ for the decomposed version (the same values used in Farina \textit{et al.} \cite{Farina2023}). In total, we used approximately 16 minutes of QPU time for our experiments.
The subproblem size in De-RQuMF is set to $40$ (the same as in Farina \textit{et al.}~\cite{Farina2023}). 
We adopt the maximum chain length criterion for all conducted experiments to calculate chain length for D-Wave experiments. 
This involves initially mapping a logical graph onto a physical graph through the process of minor embedding. 
After determining the final embedding, we calculate the length ($l$) of the longest chain of qubits. 
Subsequently, the chain strength parameter is established by adding a small offset to $l$ specifically, an offset of $0.5$ inline with the previous works \cite{arrigoni2022quantum,BirdalGolyanikAl21}.
See further implementation details of R-QuMF in the released source code.


\section{Experiments} 
\label{sec:Experiments}

We assess the effectiveness of the proposed method  on both synthetic and real datasets.
Specifically, we compare our approach with RanSaCov \cite{Magri2016}, a classical method, and a quantum apporach, namely the recently introduced QuMF~\cite{Farina2023}, as these methods are the closest competitors (see Sec.~\ref{sec:Related_Work}). 
To evaluate the performance of the analysed methods we adopt the \emph{misclassification error}, denoted as $E_{mis}$, which judges the quality of MMF in terms of the segmentations attained. 
Specifically, the misclassification error counts the number of misclassified  points as follows: first,
each point is assigned to a label corresponding to the model it belongs to (outliers are assigned to the label $0$); 
then, the map between ground-truth labels and estimated ones that minimises the overall number of misclassified points is found; a point is deemed as correct if one of its labels corresponds to the ground truth. 
The evaluated multi-model fitting tasks include fitting lines to 2D points (synthetic), fitting planes to 3D points (real) and two-view segmentation on the AdelaideRMF dataset \cite{Wong2011} for fitting fundamental matrices or homographies. 
%

\subsection{Line Fitting on Synthetic Data} 
\begin{figure}[t]
  \centering
  \includegraphics[width=1.0\columnwidth]{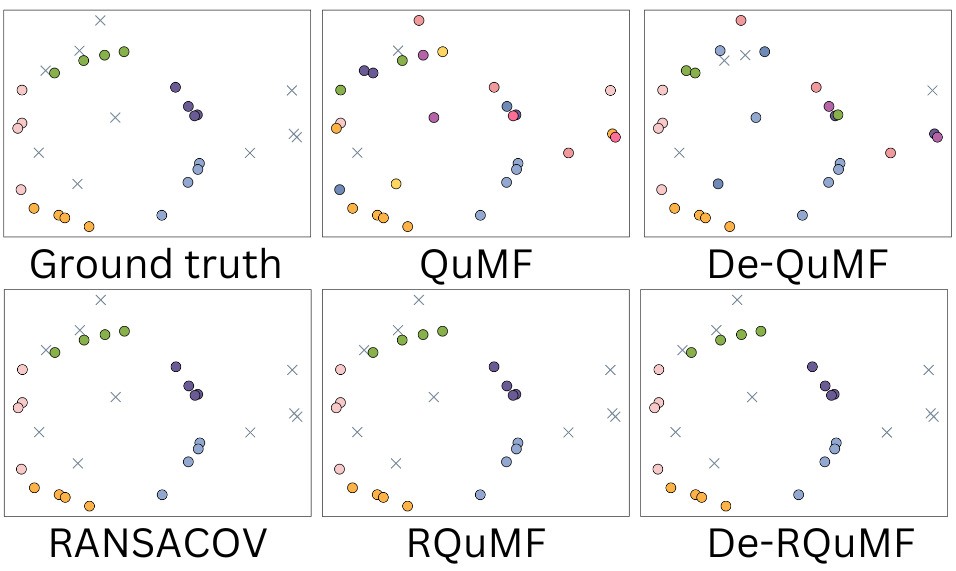}
  \caption{A sample visualization of the synthetic dataset (Ground Truth) and results for various methods for $50$ models and $33\%$ outliers (i.e.~10 outliers out of 30 points).
  } 
  \label{fig:synthetic_data_example}
\end{figure}

We evaluate our method on line fitting problems by creating a synthetic test-bed that comprises five lines arranged into a pentagon (see Fig.~\ref{fig:synthetic_data_example} for some visualisations). Each line fitting problem comprises $30$ data points, divided into outliers and inliers. Outliers are uniformly distributed, whereas inliers (i.e., points belonging to the lines) are perturbed using Gaussian noise with a standard deviation of $0.01$, and they are equally distributed among the lines. 
Each test is repeated $20$ times and the mean misclassification error is reported. 
The preference matrix is the same for all compared methods and true models have always been sampled in the pool of provisional models $\Theta$ (a standard practice). 

\smallskip 

\begin{figure}[t]
    \begin{center}
    \begin{subfigure}{0.7\columnwidth}
        \centering
        \includegraphics[width=\textwidth]{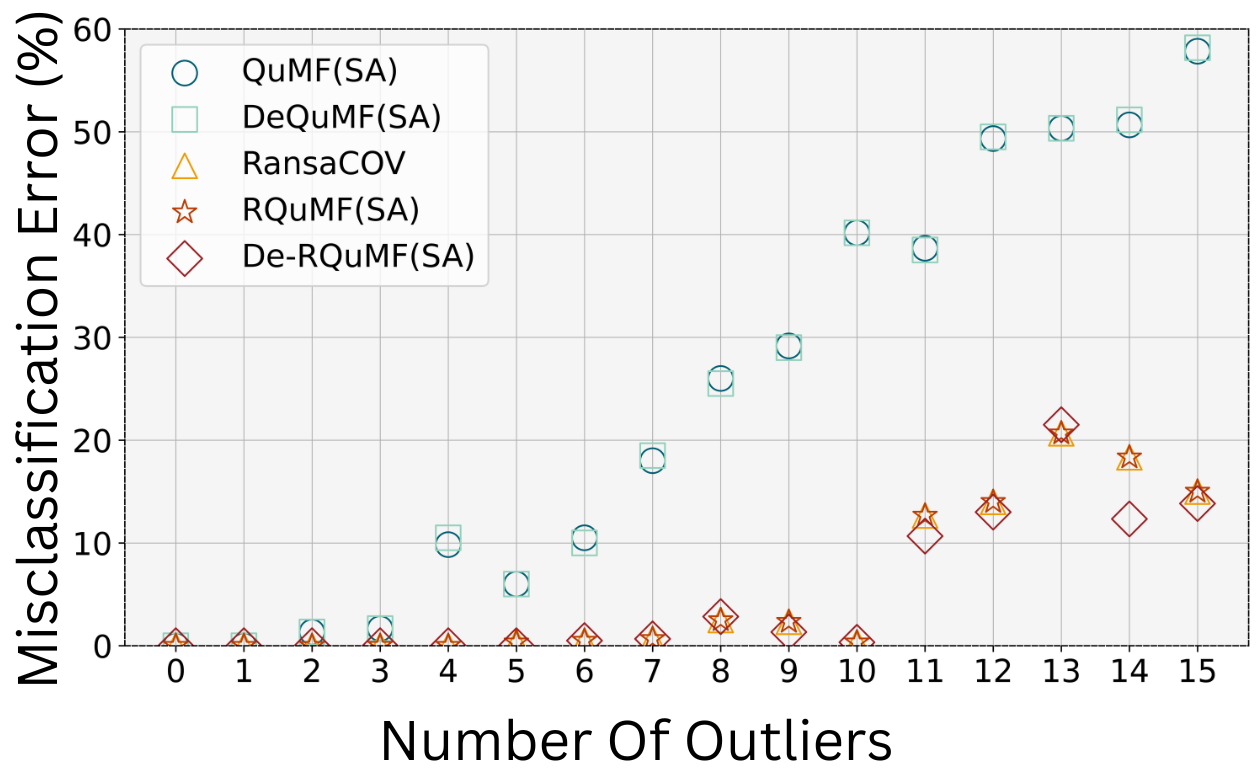}
        \label{fig:synthetic_outlier_increase}   
    \end{subfigure}\\
    \begin{subfigure}{0.7\columnwidth}
        \centering
        \includegraphics[width=\textwidth]{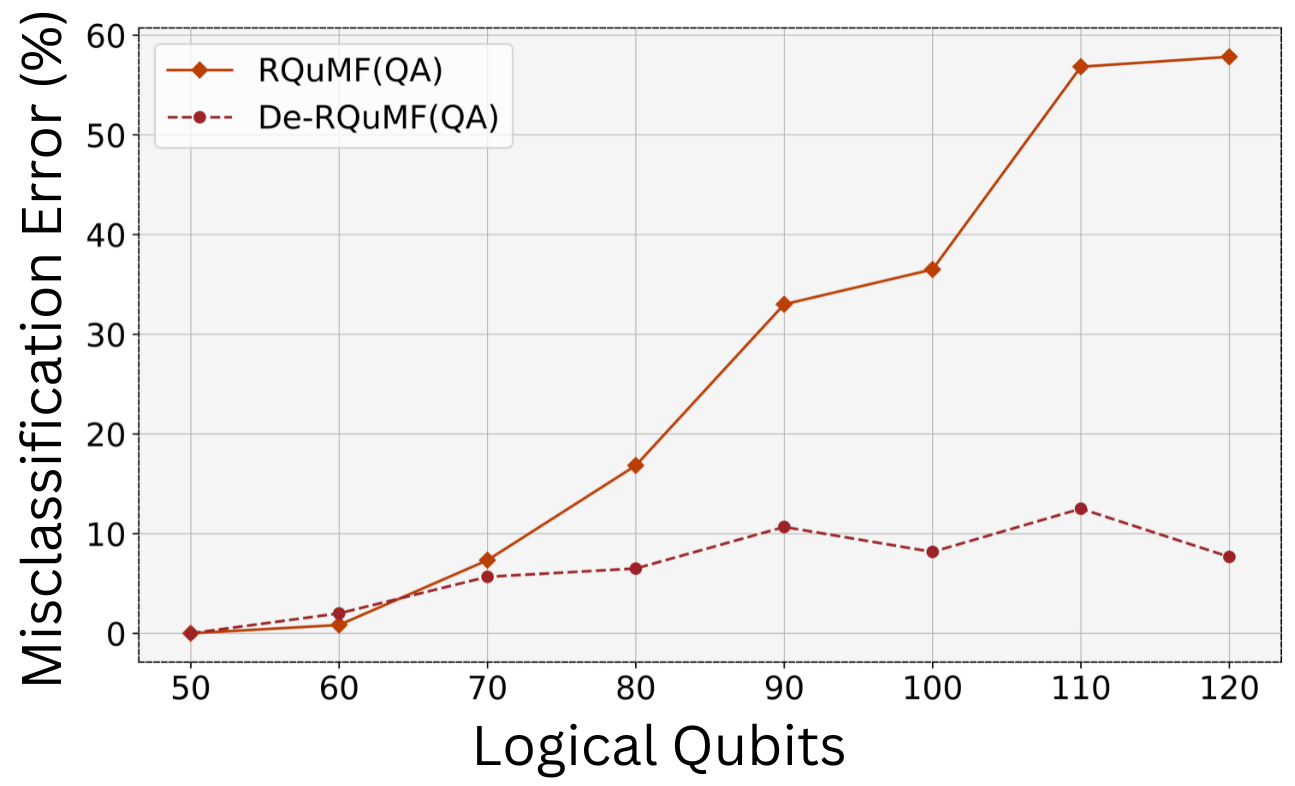}
        \label{fig:quantum_synthetic_increasing_problem_size_experiment}
    \end{subfigure}
    \end{center}
    \caption{\textbf{Top:} Misclassification Error [\%] on synthetic data for $40$ sampled models with increasing outliers ($0$-$50$\%); the problem size is fixed to $70$ ($30$ data points + $40$ models). SA is used for quantum methods. \textbf{Bottom:} Misclassification Error [\%] for synthetic data on quantum hardware with increasing problem size; outlier percentage is fixed to $17\%$. 
    Note that R-QuMF's $E_{mis}$ breaks starting from $120$ qubits. Non-robust quantum methods (i.e. QuMF and De-QuMF) are omitted because they fail in this scenario. 
    }
    \label{fig:synthetic_vizualisation}
\end{figure}

\noindent\textbf{Robustness to Outliers.} 
First, in order to evaluate the robustness of the methods, we gradually increase the outlier percentage from $0\%$ to 50\% while keeping fixed the total number of points (i.e., 30 points). 
The number of sampled models is set equal to 40 models. 
Results of this experiment are reported in Top Fig.~\ref{fig:synthetic_vizualisation}. 
QuMF and De-QuMF are not designed to deal with outliers and, as expected, they do not provide accurate results as the number of outliers increases. 
This phenomenon can also be observed in Fig.~\ref{fig:synthetic_data_example}, where outliers are erroneously detected as belonging to models supported by very few points. 
On the contrary, RanSaCov, RQuMF and De-RQuMF, which deal with outliers by design, are able to cope with a higher percentage of outliers without suffering a huge performance degradation.

\smallskip

\begin{table}[t]
  \centering 
  \begin{adjustbox}{max width=\columnwidth}
  \begin{tabular}{c|c|c|c|c|c}
     \midrule
     \textbf{\# of Models}  & \textbf{QuMF\cite{Farina2023}} & \textbf{De-QuMF\cite{Farina2023}} & \textbf{RanSaCov\cite{Magri2016}} & \textbf{RQuMF} & \textbf{De-RQuMF}\\
     \bottomrule
     20 &  \cellcolor{lightred}1.00 & \cellcolor{lightred}1.00 & \cellcolor{darkred}\textbf{0} & \cellcolor{darkred}\textbf{0} & \cellcolor{darkred}\textbf{0} \\
     \hline
     50 &  \cellcolor{lightred}16.16 & \cellcolor{midred}9.50 & \cellcolor{darkred}\textbf{0.66} & \cellcolor{darkred}\textbf{0.66} & \cellcolor{darkred}\textbf{0.66} \\
     \hline
     100 &  44.83 & \cellcolor{lightred}18.66 & \cellcolor{darkred}\textbf{0} & \cellcolor{midred}1.33 & \cellcolor{darkred}\textbf{0} \\
     \hline
     500 & 86.66 & 36.66 & \cellcolor{midred}14.66 & \cellcolor{lightred}30.99 & \cellcolor{darkred}\textbf{0} \\
     \hline
     1000 & 89.50 & 49.83 & \cellcolor{midred}15.99 & \cellcolor{lightred}35.99 & \cellcolor{darkred}\textbf{3.32} \\
     \bottomrule
  \end{tabular}
  \end{adjustbox}
  
  \caption{Misclassification Error [\%] on synthetic data with varying problem size (i.e.~number of sampled models); the outlier percentage is fixed to $17\%$ (i.e.~five outliers); data size is fixed to $30$ points. SA is used for quantum methods. This experiment is the classical counterpart of the evaluations reported in the bottom Fig.~\ref{fig:synthetic_vizualisation}. All variants of quantum-enhanced methods in this table (Farina et al.~\cite{Farina2023} and ours) use simulated annealing. 
} \label{tab:synthetic_results_varing_problem_size}
\end{table}

\noindent\textbf{Scalability.} 
In our second experiment, we focus on how the performance scales with respect to the dimension of the multi-model fitting problem at hand. Thus, having fixed the outlier ratio at $17\%$, we vary the number $m$ of sampled models between $20$ and $1000$ (with a step size of $10$ until $200$ models are reached, and a step size of $100$ after that). 
Results of this experiment are reported in Tab.~\ref{tab:synthetic_results_varing_problem_size}. 
It can be noted that, when the number $m$ of sampled models increases, the misclassification error achieved by RQuMF worsens, as the problem becomes more difficult to solve. 
On the contrary,  De-RQuMF can handle a higher number of sampled models without impacting the overall misclassification error, thanks to the decomposition principle. 
It can also be observed that our approach outperforms previous quantum work (QuMF and De-QuMF), similarly to the previous experiment. 

\smallskip

\noindent\textbf{Experiments on Quantum Hardware.} 
In order to assess the performance on Quantum Hardware, we also repeat the previous experiment on D-Wave quantum annealer Advantage 5.4. 
We consider a constant outlier ratio of $17\%$ and problem size varying from $50$ to $120$ qubits (which is the sum of the number of points and models, namely 30 data points plus 20 to 140 sampled models). 
Results are reported in Bottom Fig.~\ref{fig:synthetic_vizualisation}. 
We observe that RQuMF cannot handle problems starting from $80$ models (\emph{i.e.}, 110 qubits)--when its error goes beyond $50\%$, while De-RQuMF remains robust for the entire range of problem sizes with $E_{mis}$ predominantly staying under $10\%$.  This confirms the advantages of a decomposed and iterative approach for handling high-dimensional problems.
 The fact that RQuMF can not manage large-scale problems is not surprising since quantum hardware is far from being mature, in agreement with previous work on quantum computer vision \cite{BirdalGolyanikAl21,arrigoni2022quantum}. 
It is worth noticing that results on quantum hardware are worse than the ones with SA reported in Tab.~\ref{tab:synthetic_results_varing_problem_size}, as expected. For example, SA can successfully manage problems with 100 sampled models with $1.33\%$ error whereas QA fails.

\begin{figure}[t]
    \centering
    \begin{subfigure}{0.5\columnwidth}
        \centering
        \includegraphics[width=\textwidth]{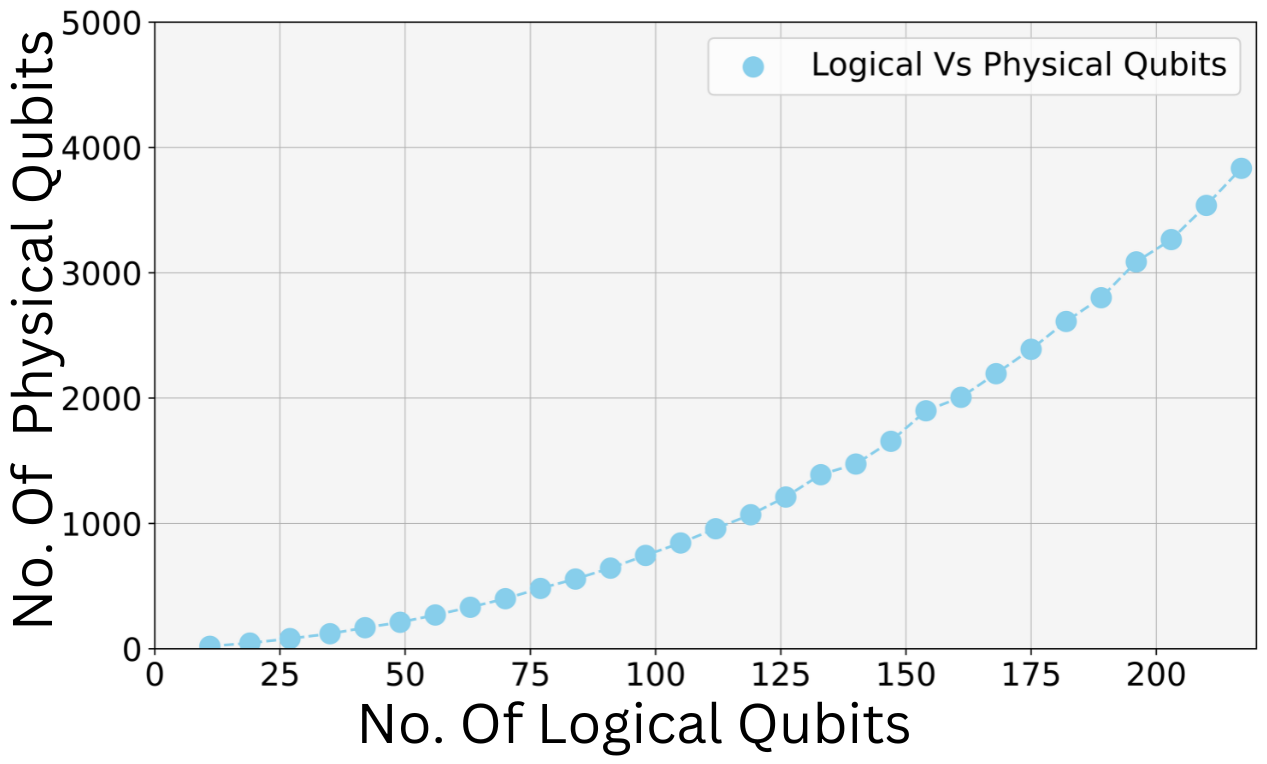}
    \end{subfigure}%
    \begin{subfigure}{0.5\columnwidth}
        \centering
        \includegraphics[width=\textwidth]{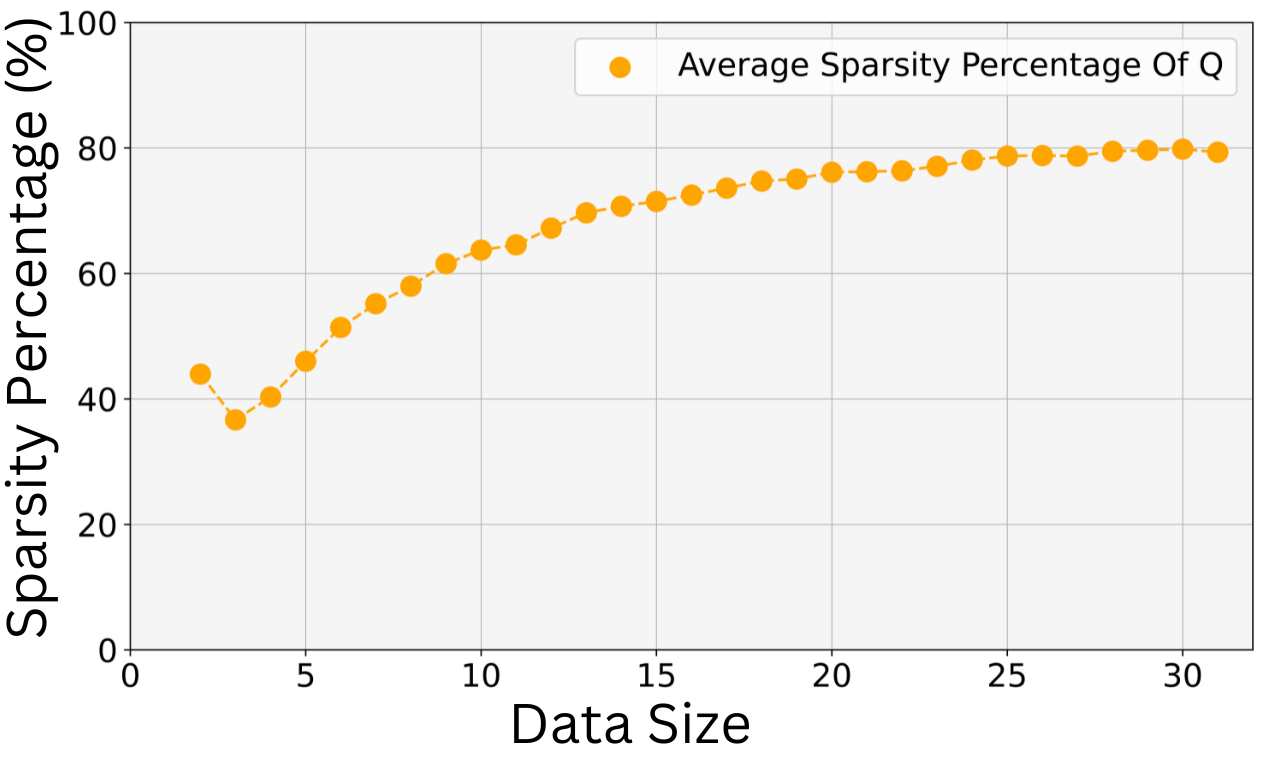}
    \end{subfigure}
    \caption{
    Analysis of R-QuMF runs on our synthetic dataset.
    %
    \textbf{Left:} The number of physical qubits as a function of the number of logical problem qubits for data points varying in the range $[2; 32]$; sampled models are $6$ times the data size. \textbf{Right:} The sparsity of $\widetilde{Q}$ in $\%$ as the function of the input data size. 
    }
    \label{fig:both}
\end{figure}

To further analyze this aspect, we visualize in Left Fig.~\ref{fig:both} how the number of physical qubits (that reflects the effective allocation of QPU resources) scales with respect to the increasing problem size (represented by the number of logical qubits) on sample problems. 
Although $\widetilde{Q}$ in our QUBO is significantly sparse (see Fig.~\ref{fig:both}-(right)), the number of physical qubits still increases superlinearly with respect to the logical qubits, approaching the maximum size that can typically be handled by an adiabatic quantum computer. 
These results are in line with previous quantum work \cite{Farina2023}. 

\subsection{Motion Segmentation on Real Data}
\label{sec:experiments:real}

We consider the {AdelaideRMF dataset \cite{Wong2011}}, which encompasses two distinct types of multi-model fitting tasks: fitting fundamental matrices (15 image pairs with at least two moving objects) and homographies ($16$ image pairs with at least two planes). 
Our evaluation specifically targets the multi-model sequences associated with both types of fitting problems and we do not take into account single-model fitting. 
The outlier percentages for these data are depicted in 
the supplementary material, with nearly all sequences exhibiting an outlier rate exceeding 30\%, and some reaching as high as 68\% (for more detail about the outlier distribution see Fig.~\ref{fig:fm_hm_outlier_percentage} from supplementary material): this presents a substantial challenge for accurate model fitting.

The number of sampled models for each instance is six times the number of points. As before, the preference matrices used as input remain consistent across all evaluated methods. We conducted each experiment 20 times, reporting the average $E_{mis}$.

\begin{table*}[t]
  
  \centering 
  \begin{adjustbox}{max width=\textwidth}
  \begin{tabular}{l|l|c|c|c|c|c||c|c|c|c}
     \midrule
     \textbf{Outliers Settings} & {} & \textbf{QuMF(SA)\cite{Farina2023}} & \textbf{De-QuMF(SA)\cite{Farina2023}} & \textbf{RanSaCov\cite{Magri2016}} & \textbf{RQuMF(SA)} & \textbf{De-RQuMF(SA)}& \textbf{QuMF(G)\cite{Farina2023}} & \textbf{De-QuMF(G)\cite{Farina2023}} & \textbf{RQuMF(G)} & \textbf{De-RQuMF(G)}\\
     \bottomrule
     \multirow{2}{*}{No Outliers} & Mean & \cellcolor{midred}3.61 & \cellcolor{darkred}\textbf{0.84} & 9.79 & \cellcolor{lightred}6.46 & 11.47 & \cellcolor{midred}3.14 & \cellcolor{darkred}1.29  & 12.56 & \cellcolor{lightred}11.96 \\
     & Median & \cellcolor{lightred}2.68 & \cellcolor{darkred}\textbf{0.28} & 7.97 & \cellcolor{midred}2.41 & 10.55 & \cellcolor{midred}1.87 & \cellcolor{darkred}0.93  & 11.35 & \cellcolor{lightred}10.55\\
     \hline
     \multirow{2}{*}{With Outliers} & Mean & 40.37 & 26.19 & \cellcolor{darkred}\textbf{7.22} & \cellcolor{midred}10.46 & \cellcolor{lightred}12.69 & 45.81 & \cellcolor{lightred}26.21  & \cellcolor{midred}13.14 & \cellcolor{darkred}12.84\\
     & Median & 39.82 & 26.94 & \cellcolor{darkred}\textbf{5.76} & \cellcolor{midred}8.33 & \cellcolor{lightred}11.18 & 46.58 & \cellcolor{lightred}27.35  & \cellcolor{darkred}10.96 & \cellcolor{midred}11.33 \\
     \hline
     With Outliers $+$ & Mean & 19.76 & \cellcolor{darkred}\textbf{8.89} & NA & \cellcolor{midred}9.70 & \cellcolor{lightred}12.48 & 25.61 & \cellcolor{darkred}8.94  & \cellcolor{lightred}12.67 & \cellcolor{midred}12.59\\
    Post Processing & Median & 19.67 & \cellcolor{darkred}\textbf{6.55} & NA &  \cellcolor{midred}8.02 & \cellcolor{lightred}11.09 & 27.92 & \cellcolor{darkred}7.12 &  \cellcolor{midred}10.75 & \cellcolor{lightred}11.33 \\
     \bottomrule
  \end{tabular}
  \end{adjustbox}
  
  \caption{Misclassification Error [\%] for the 15 multi-model fundamental matrix sequences from AdelaideRMF \cite{Wong2011} using SA or Gurobi (for quantum methods). Results for RanSaCov without and with outliers are taken from \cite{Farina2023} and \cite{MagriPhDThesis}, respectively. 
  }
   \label{tab:fm_outlier_detection_comparison_SA}
\end{table*}

\begin{table}[t]
    \centering 
  \begin{adjustbox}{max width=\columnwidth}
  \begin{tabular}{l|l|c|c|c|c|c}
     \midrule
     \textbf{Outliers Settings} & {} & \textbf{QuMF\cite{Farina2023}} & \textbf{De-QuMF\cite{Farina2023}} & \textbf{RanSaCov\cite{Magri2016}} & \textbf{RQuMF} & \textbf{De-RQuMF}\\
     \bottomrule
     \multirow{2}{*}{No Outliers} & Mean & 54.10 & \cellcolor{darkred}\textbf{13.94} & - & \cellcolor{lightred}20.77 & \cellcolor{midred}19.11 \\
     & Median & 55.44 & \cellcolor{darkred}\textbf{16.20} & - & \cellcolor{lightred}25.11 & \cellcolor{midred}24.96 \\
     \hline
     \multirow{2}{*}{With Outliers} & Mean & 86.23 & 49.32 & \cellcolor{midred}14.72 & \cellcolor{lightred}17.01 & \cellcolor{darkred}\textbf{14.33} \\
     & Median & 86.25 & 46.35 & \cellcolor{darkred}\textbf{14.38} & \cellcolor{lightred}16.72 & \cellcolor{midred}15.76 \\
     \hline
     With Outliers $+$ & Mean & 51.22 & \cellcolor{lightred}26.57 & - & \cellcolor{midred}16.75 & \cellcolor{darkred}\textbf{14.21} \\
     {Post Processing} & Median & 50.76 & \cellcolor{lightred}20.46 & - & \cellcolor{midred}16.63 & \cellcolor{darkred}\textbf{16.05} \\
     \bottomrule
  \end{tabular}
  \end{adjustbox}
  \caption{Misclassification Error [\%] for the $16$ multi-model homography sequences from AdelaideRMF \cite{Wong2011} using SA (for quantum methods). 
    Results for RanSaCov without outliers are not available whereas those with outliers are taken from \cite{MagriPhDThesis}. 
    De-QuMF with post-processing fails on at least one sequence in $16$ out of $20$ trials. 
    } 
    \label{tab:hm_outlier_detection_comparison_SA}
\end{table}

\begin{figure}[t]
\centering
\begin{subfigure}[b]{0.48\columnwidth}
    \includegraphics[width=\linewidth]{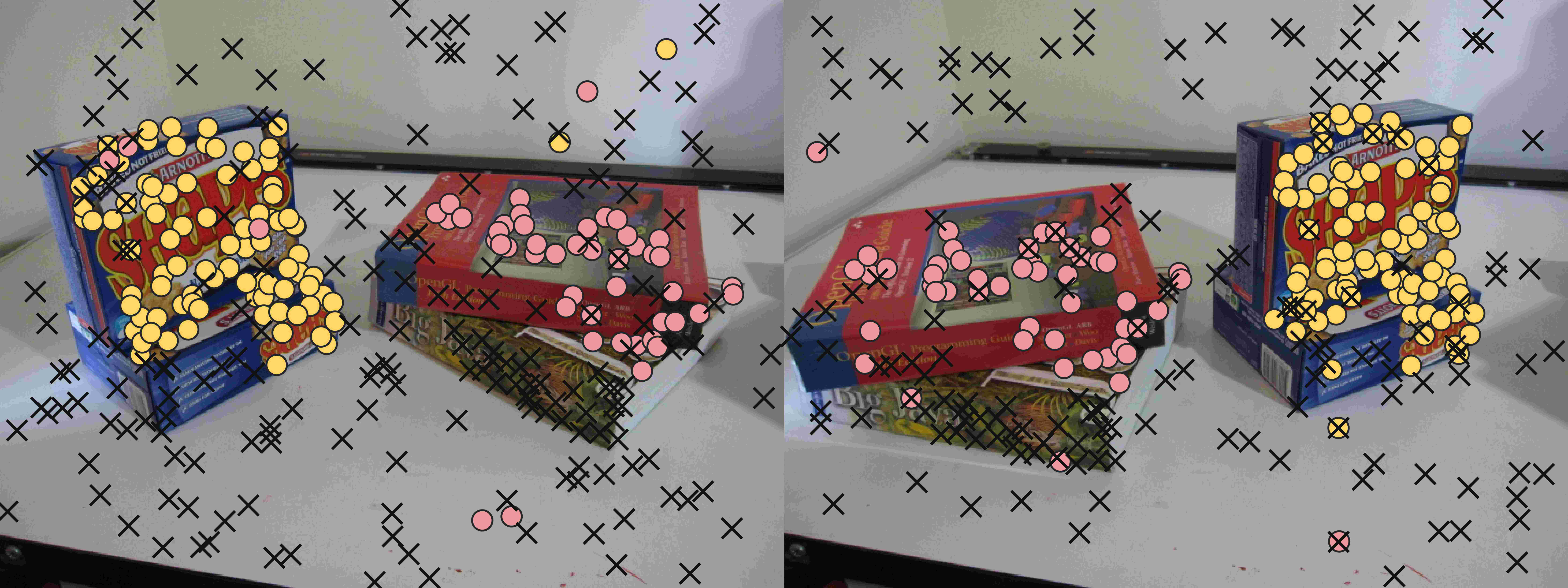}
    \caption{QuMF (Post Processing), $E_{mis} = 16.72\%$}
\end{subfigure}
\begin{subfigure}[b]{0.48\linewidth}
    \includegraphics[width=\linewidth]{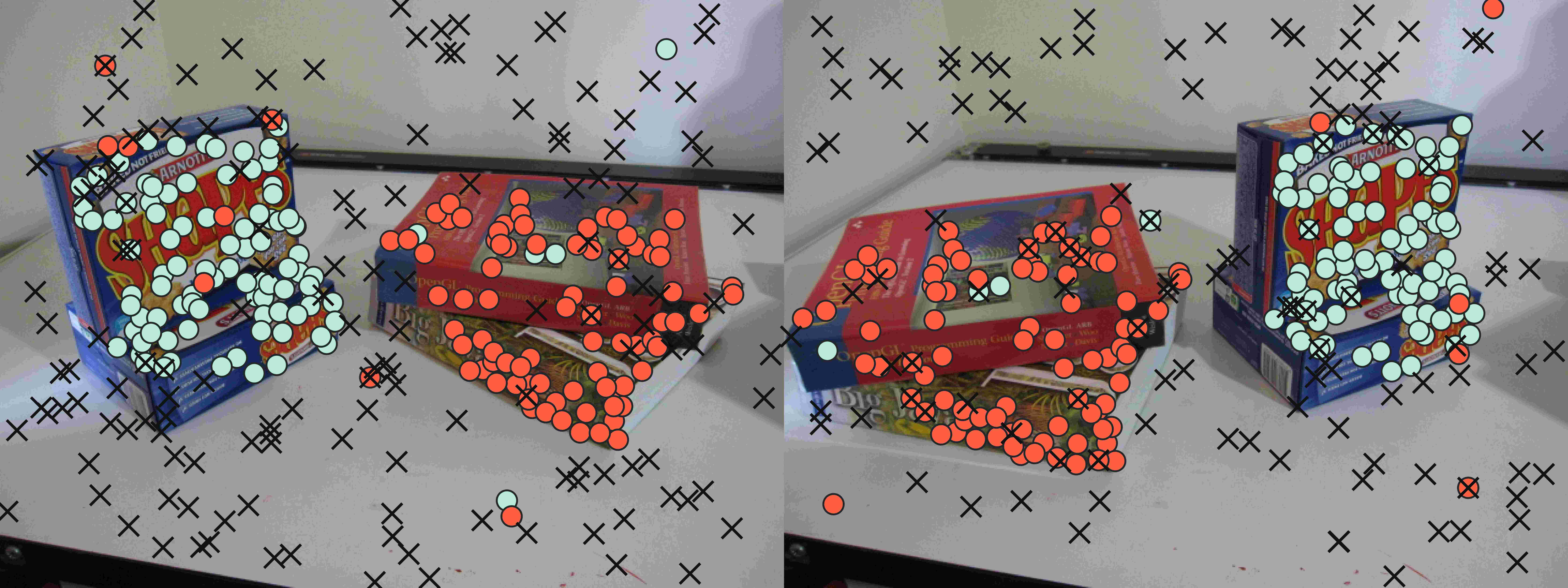}
    \caption{DeQuMF (Post Processing), $E_{mis} = 6.16\%$}
\end{subfigure}
\begin{subfigure}[b]{0.48\linewidth}
    \includegraphics[width=\linewidth]{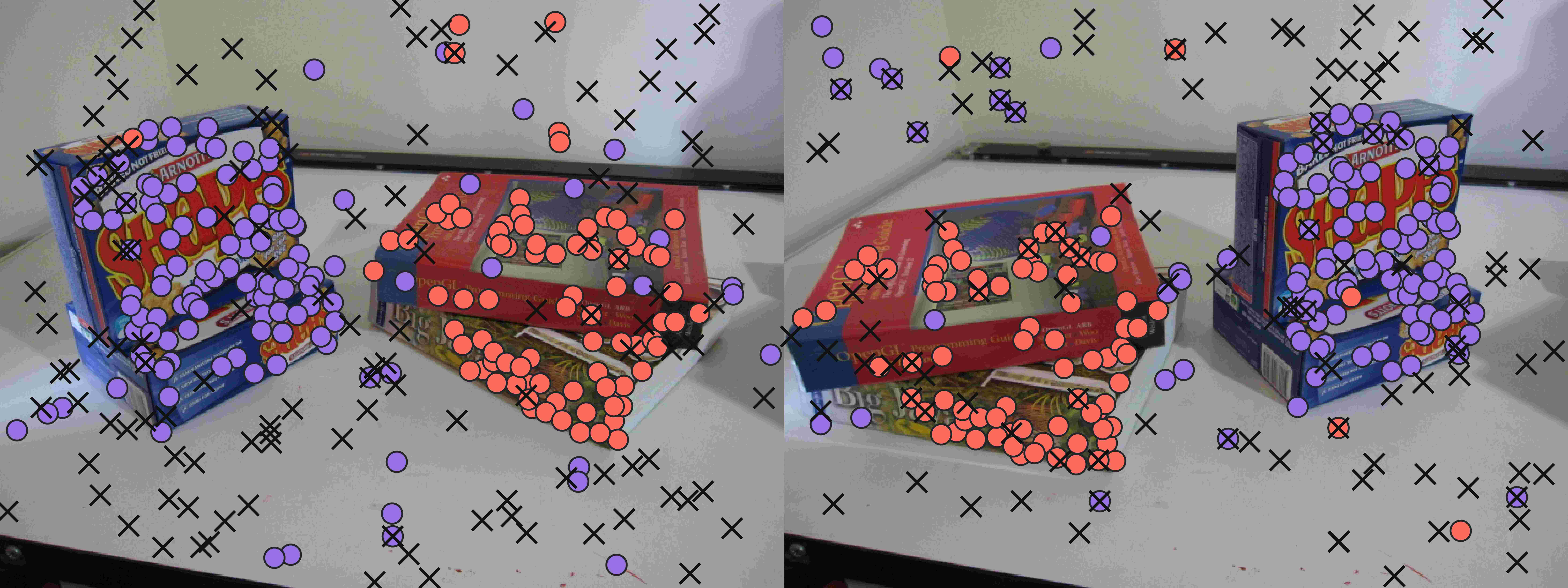}
    \caption{RQuMF, $E_{mis} = 5.87\%$}
\end{subfigure}
\begin{subfigure}[b]{0.48\linewidth}
    \includegraphics[width=\linewidth]{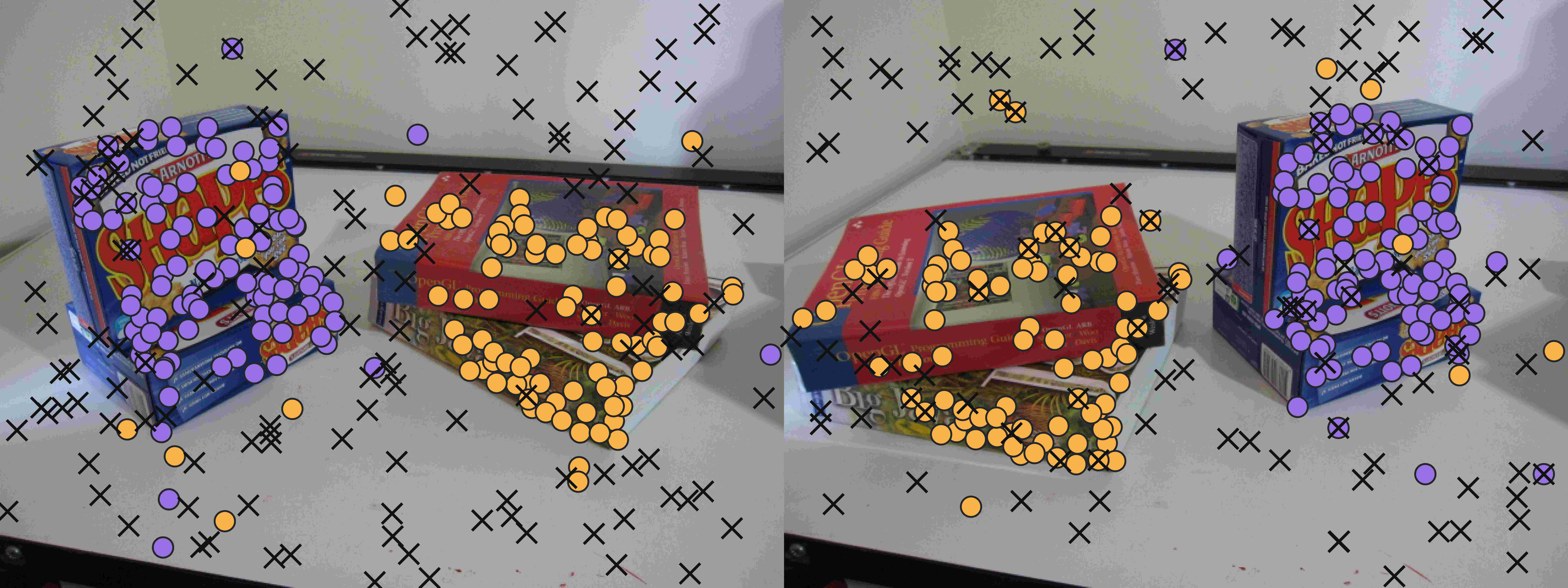}
    \caption{De-RQuMF, $E_{mis} = 5.57\%$}
\end{subfigure}

\caption{Sample results of fundamental matrix fitting on \textit{biscuitbook} using SA. Our method performs as well as QuMF and De-QuMF which use the information about the number of ground-truth models. %
}
\label{fig:fm_visualization_grid}
\end{figure}

\begin{figure}[t]
\centering
\begin{subfigure}[b]{0.48\linewidth}
    \includegraphics[width=\linewidth]{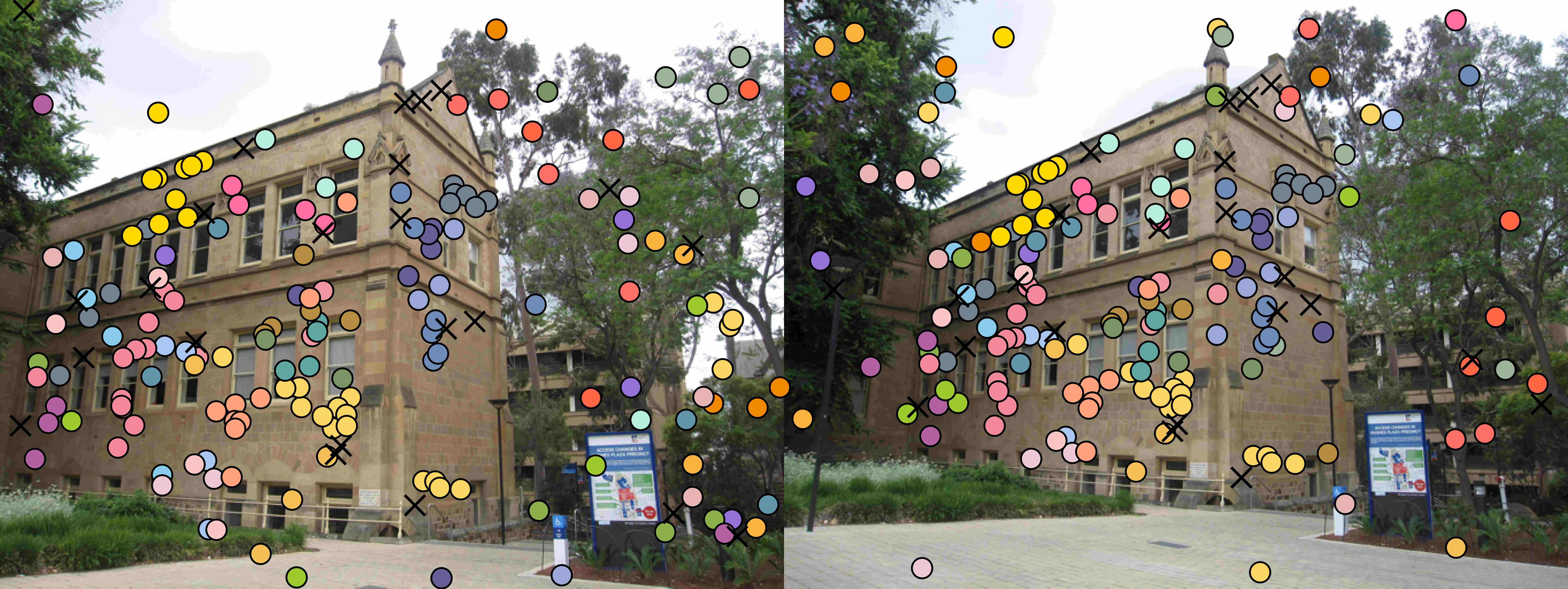}
    \caption{QuMF, $E_{mis} = 93.00\%$}
\end{subfigure}
\begin{subfigure}[b]{0.48\linewidth}
    \includegraphics[width=\linewidth]{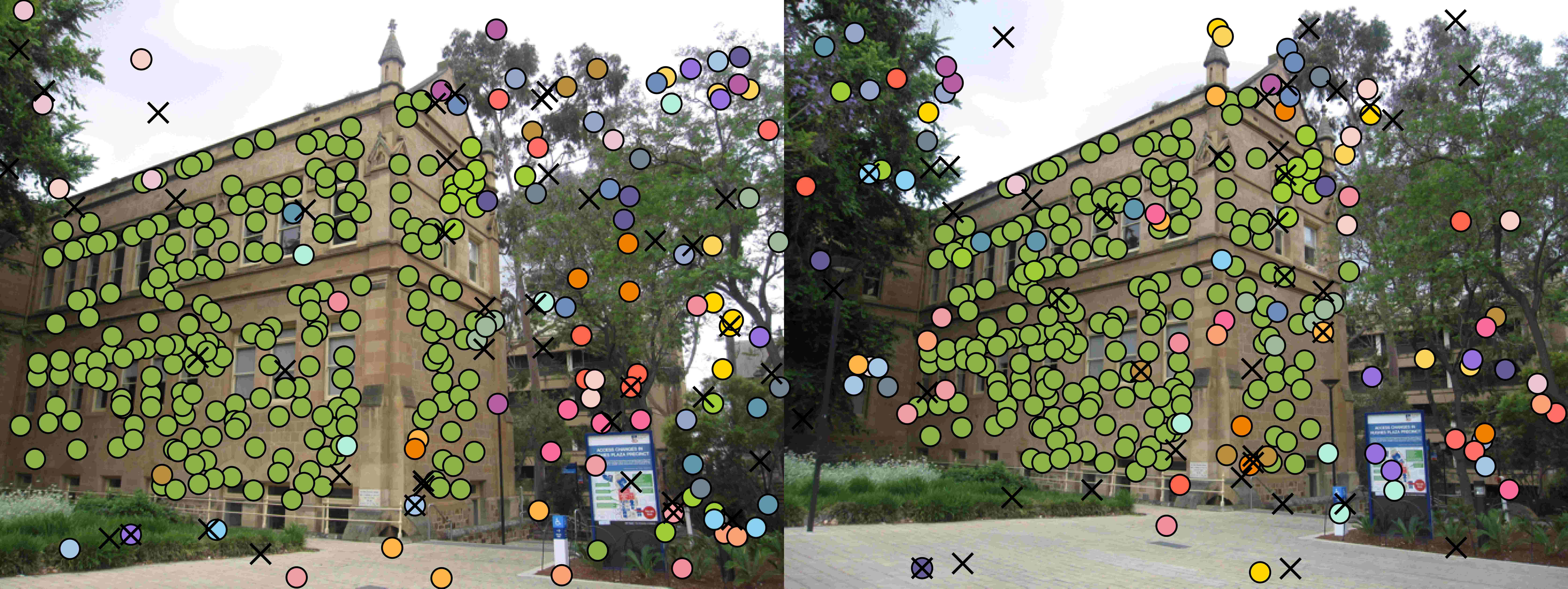}
    \caption{DeQuMF, $E_{mis} = 41.69\%$}
\end{subfigure}
\begin{subfigure}[b]{0.48\linewidth}
    \includegraphics[width=\linewidth]{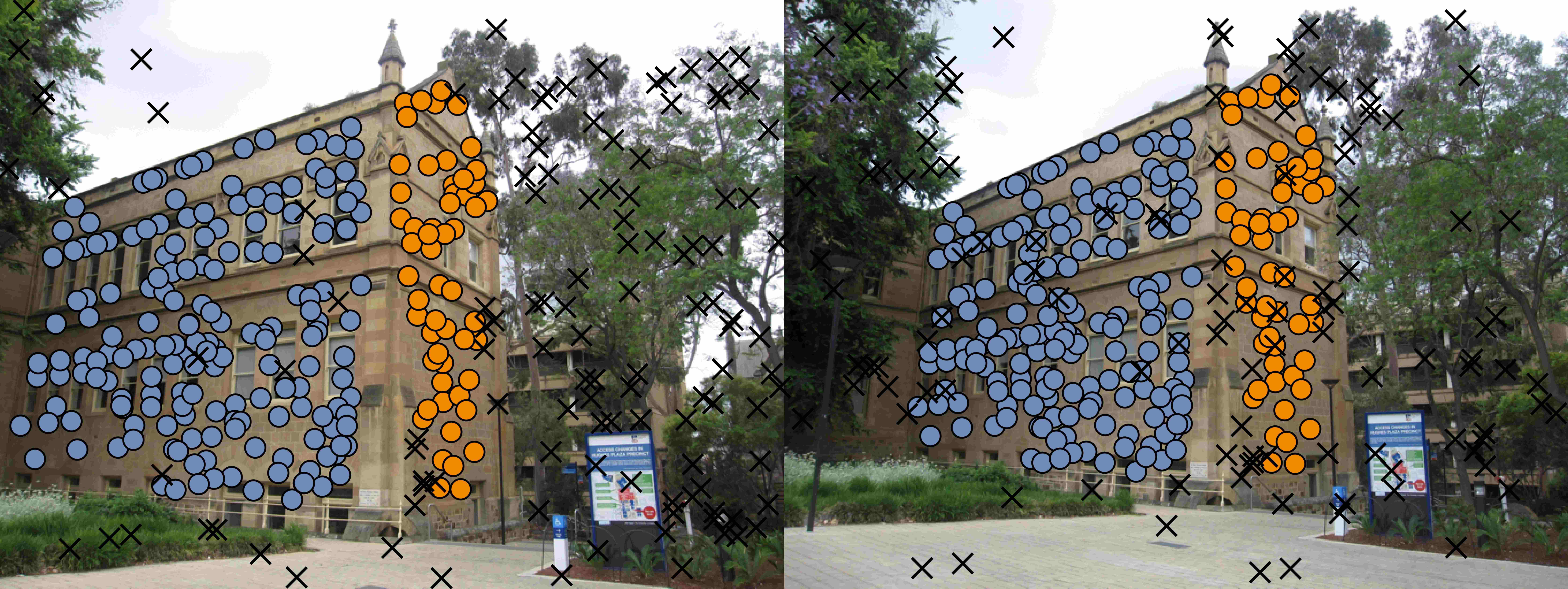}
    \caption{RQuMF, $E_{mis} = 2.9\%$}
\end{subfigure}
\begin{subfigure}[b]{0.48\linewidth}
    \includegraphics[width=\linewidth]{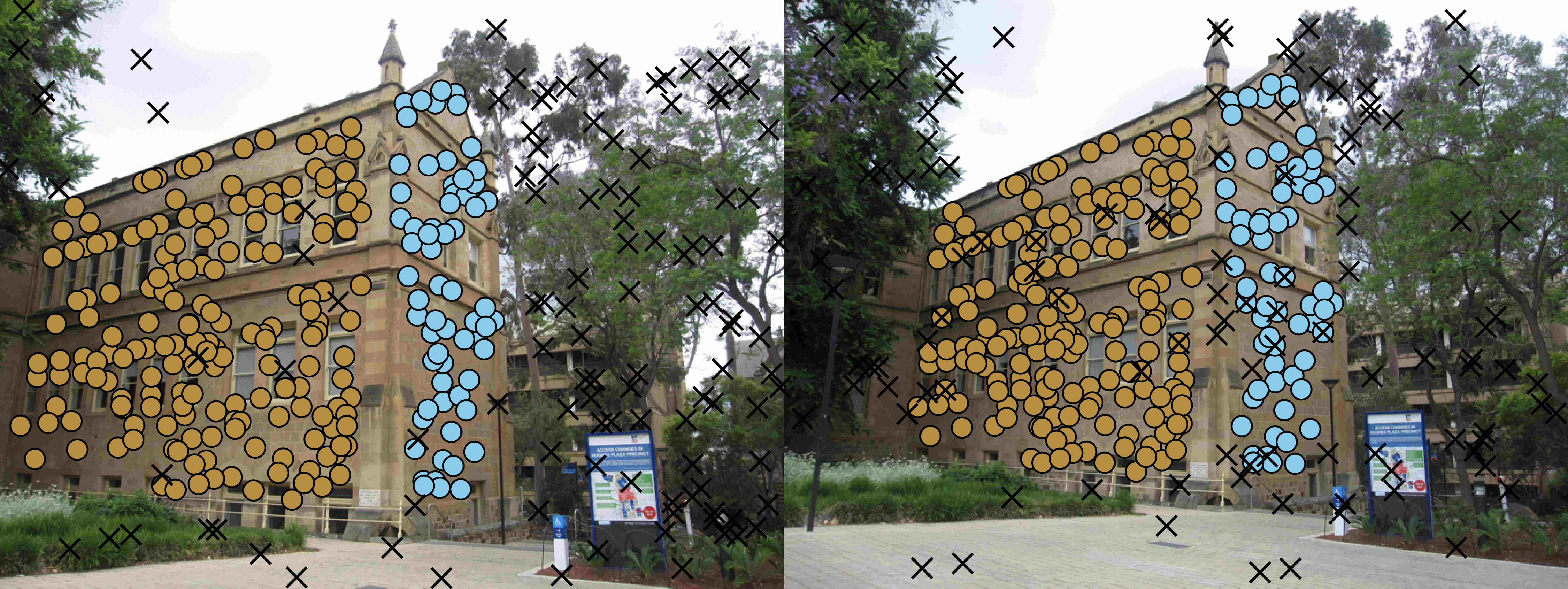}
    \caption{De-RQuMF, $E_{mis} = 0.53\%$}
\end{subfigure}
\caption{Sample result of homography fitting on \textit{oldclassicswing} ($32\%$ outliers) using SA. In the absence of the true number of models both QuMF and De-QuMF fail. Both our proposed methods achieve a near-perfect score. }
\label{fig:hm_visualization_grid}
\end{figure}

Unlike the synthetic experiments, where computational resources are less constrained, for real data we do not report results obtained using QA
due to limited QPU time. 
In addition to SA, we also consider the Gurobi solver in order to enrich the evaluation. 
In addition to the original outlier-contaminated sequences, we also consider the same sequences where outliers have been removed, to study the impact of outliers and diverse behaviour of the considered methods. 
We also analyse the efficacy of post-processing, especially related to QuMF. 
Specifically, we examine whether 1) selecting the top \textit{k} models identified by a method and 2) designating all points not accounted for by these \textit{k} models as outliers, would yield robustness. %
Note that this post-processing does not make sense for RanSaCov, for it enforces hard constraints in its formulation and returns at most  $k$ models.
Aggregated results for fitting fundamental matrices are given in Tab.~\ref{tab:fm_outlier_detection_comparison_SA}, whereas those for homographies are reported in Tab.~\ref{tab:hm_outlier_detection_comparison_SA}. See also Fig.~\ref{fig:fm_visualization_grid} and \ref{fig:hm_visualization_grid} for sample qualitative results. (More visualizations can be referenced from the supplementary material in Fig.~\ref{fig:fm_visualization_grid_supp} and \ref{fig:hm_visualization_grid_supp})

Similar conclusions can be drawn for fitting fundamental matrices and homographies. 
It is not surprising that QuMF outperforms our approach in the outlier-free scenario (as seen in the first row of the aforementioned tables), given that QuMF is specifically designed for such settings. Additionally, the lambda parameters in RQuMF have been fine-tuned for scenarios involving outlier contamination. 

In the case with outliers (second row of the tables), however, QuMF is significantly worse than RQuMF, therefore showing that explicitly modeling outliers is indispensable for achieving robustness. 
Using the post-processing largely improves the performance of QuMF, which, however, is still not comparable to RQuMF. 
Note also that the post-processing assumes the knowledge of the number of true models $k$ which is typically unavailable in practice. 
Our method is not influenced by post-processing, thereby showing that it selects the right number of models in most cases, and it is better than RanSaCov in many scenarios. 
Concerning quantum methods, there are no significative differences between using Gurobi or SA solvers. 
The fact that the decomposed version of our approach does not improve upon using the full QUBO, could be due to the task difficulty in terms of outlier corruption compared to the simplified scenarios with synthetic data. %
From Fig.~\ref{fig:fm_visualization_grid} %
it becomes apparent that QuMF can segment the entire model accurately only with the aid of the post-processing phase. Without post-processing, QuMF
struggles to segment the entire model accurately, often choosing multiple models to explain what essentially constitutes a single true model (see Fig.~\ref{fig:hm_visualization_grid}).

\subsection{Plane Fitting on 3D Point Clouds}
\begin{figure}[h!]
    \centering
    \includegraphics[width=\columnwidth]{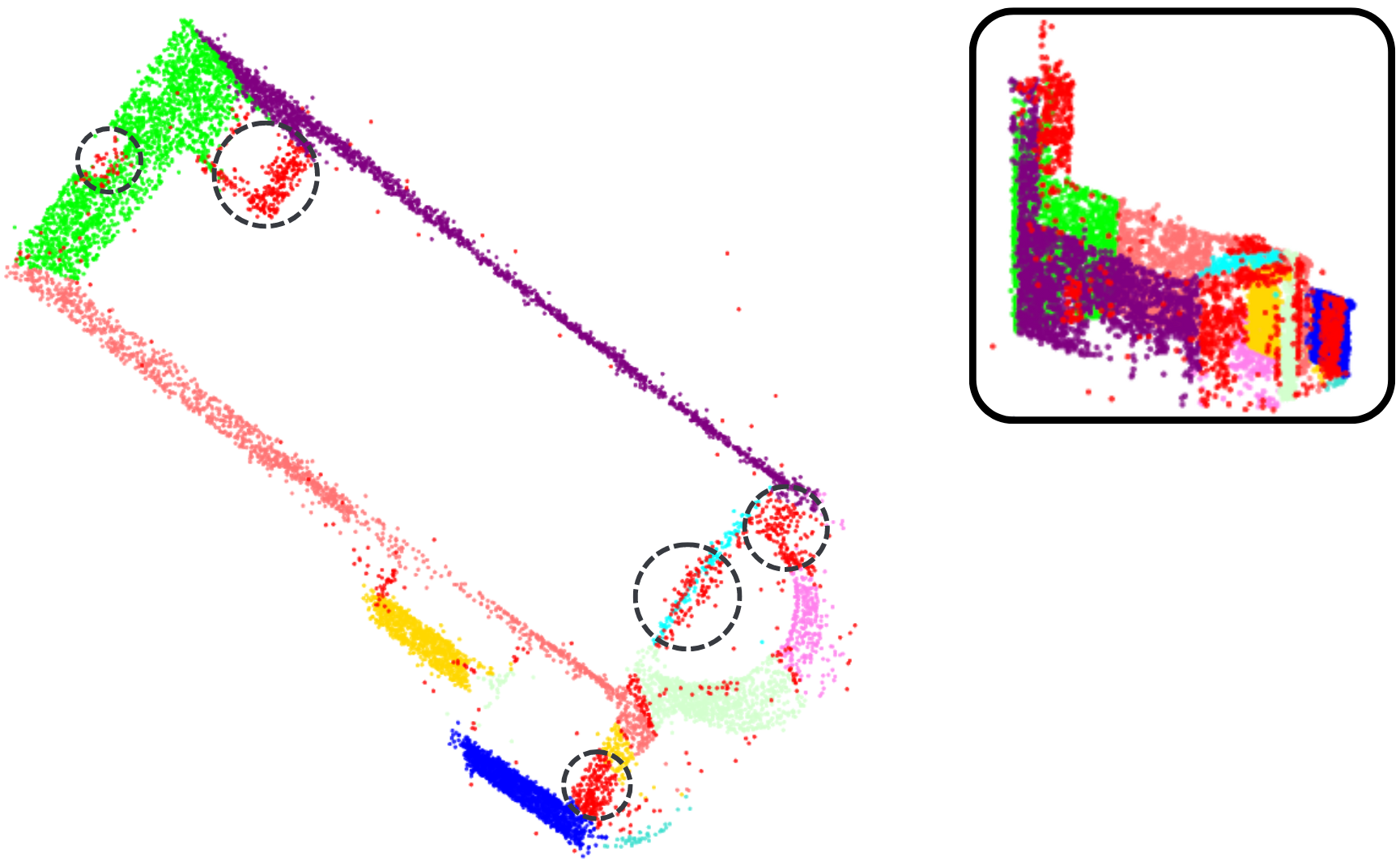}
    \caption{A plane fitting example using the De-RQuMF(SA) method. The encircled dark red points are uncovered and treated as outliers. The inset view on the top-right shows the same result from a different virtual camera perspective.} 
    \label{fig:plane_fitting_drqumf}
\end{figure}

Finally, we illustrate the versatility and practicality of our approach in a 3D plane fitting scenario. %
We consider a 3D point cloud obtained through image-based 3D reconstruction \cite{Samantha2024}. 
The dataset comprises $10812$ points; we sample $2000$ models from those, focusing exclusively on planar structures, with an inlier threshold set to $0.5$ and use the SA solver. 
Fig.~\ref{fig:plane_fitting_drqumf} provides a visual example of a fitting performed using our decomposed method, De-RQuMF. 
The results demonstrate that our method identifies distinct planes within the point cloud. 
As expected, the fitting accuracy for cylindrical sections of the building is lower as our method supports plane sampling exclusively per design. 
%


\section{Conclusion}
\label{sec:conclusion} 

Based on experimental evidence, we conclude that explicitly accounting for outliers in the model significantly lowers the misclassification error across a wide variety of scenarios, compared to all competing quantum-admissible methods. 
With respect to QuMF \cite{Farina2023}, the price to pay in RQuMF for outlier robustness is an increased dimensionality of the $Q$ matrix, which is, however, a minor factor in our iterative version and not a major limitation. 
More importantly, RQuMF does not assume the true number of models explaining the data, which is highly advantageous in practice. 
Although the attained results are promising, one of the limitations of our approach is that performance is unpredictable when outliers exceed 50\% of the data. 
While RQuMF already works on real quantum hardware for small problems, 
we believe its usefulness will increase as the quantum hardware is improving. 
Managing heterogeneous models (e.g.~both planar and cylindrical models) is a promising direction for future extensions.


\paragraph{Acknowledgements.} This paper is partially supported by the PNRR-PE-AI FAIR project funded by the NextGeneration EU program and by Geopride (ID: 2022245ZYB CUP: D53D23008370001) funded by PRIN 2022. 

\newpage
{
    \small
    \bibliographystyle{ieeenat_fullname}
    \bibliography{main,LucaNEW, NostriNEW, FedeNEW}

\begin{thebibliography}{47}
\providecommand{\natexlab}[1]{#1}
\providecommand{\url}[1]{\texttt{#1}}
\expandafter\ifx\csname urlstyle\endcsname\relax
  \providecommand{\doi}[1]{doi: #1}\else
  \providecommand{\doi}{doi: \begingroup \urlstyle{rm}\Url}\fi

\bibitem[{3dflow srl.}(2024)]{Samantha2024}
{3dflow srl.}
\newblock Samantha.
\newblock \url{https://www.3dflow.net/}, 2024.
\newblock online; accessed on the 01.08.2024.

\bibitem[Agarwal et~al.(2005)Agarwal, Lim, Zelnik-manor, Perona, Kriegman, and
  Belongie]{AgarwalJongowooAl05}
Sameer Agarwal, Jongwoo Lim, Lihi Zelnik-manor, Pietro Perona, David Kriegman,
  and Serge Belongie.
\newblock Beyond pairwise clustering.
\newblock In \emph{CVPR}, pages 838--845, 2005.

\bibitem[Arrigoni et~al.(2022)Arrigoni, Menapace, Benkner, Ricci, and
  Golyanik]{arrigoni2022quantum}
Federica Arrigoni, Willi Menapace, Marcel~Seelbach Benkner, Elisa Ricci, and
  Vladislav Golyanik.
\newblock Quantum motion segmentation.
\newblock In \emph{ECCV}, pages 506--523. Springer, 2022.

\bibitem[Barath and Matas(2018)]{BarathMatas17}
Daniel Barath and Jiri Matas.
\newblock Multi-class model fitting by energy minimization and mode-seeking.
\newblock In \emph{ECCV}, pages 229--245. Springer, 2018.

\bibitem[Barath and Matas(2019)]{BarathMatas19}
Daniel Barath and Jiri Matas.
\newblock Progressive-x: Efficient, anytime, multi-model fitting algorithm.
\newblock In \emph{CVPR}, pages 3780--3788, 2019.

\bibitem[Barath et~al.(2023)Barath, Rozumny, Eichhardt, Hajder, and
  Matas]{BarathRozumnyiAl23}
Daniel Barath, Denys Rozumny, Ivan Eichhardt, Levente Hajder, and Jiri Matas.
\newblock Finding geometric models by clustering in the consensus space.
\newblock In \emph{CVPR}, pages 5414--5424, 2023.

\bibitem[Bergstra et~al.(2011)Bergstra, Bardenet, Bengio, and
  K{\'e}gl]{bergstra2011algorithms}
James Bergstra, R{\'e}mi Bardenet, Yoshua Bengio, and Bal{\'a}zs K{\'e}gl.
\newblock Algorithms for hyper-parameter optimization.
\newblock In \emph{NIPS}, 2011.

\bibitem[Bhatia et~al.(2023)Bhatia, Tretschk, Lähner, Benkner, Möller,
  Theobalt, and Golyanik]{bhatia2023CCuantuMM}
Harshil Bhatia, Edith Tretschk, Zorah Lähner, Marcel Benkner, Michael Möller,
  Christian Theobalt, and Vladislav Golyanik.
\newblock Ccuantumm: Cycle-consistent quantum-hybrid matching of multiple
  shapes.
\newblock In \emph{CVPR}, 2023.

\bibitem[Birdal et~al.(2021)Birdal, Golyanik, Theobalt, and
  Guibas]{BirdalGolyanikAl21}
Tolga Birdal, Vladislav Golyanik, Christian Theobalt, and Leonidas Guibas.
\newblock Quantum permutation synchronization.
\newblock In \emph{CVPR}, 2021.

\bibitem[Chin et~al.(2009)Chin, Wang, and Suter]{ChinWangAl09}
Tat{-J}un Chin, Hanzi Wang, and D. Suter.
\newblock Robust fitting of multiple structures: The statistical learning
  approach.
\newblock In \emph{ICCV}, pages 413--420, 2009.

\bibitem[Chin et~al.(2020)Chin, Suter, Ch'ng, and Quach]{Chin2020}
Tat-Jun Chin, David Suter, Shin-Fang Ch'ng, and James~Q. Quach.
\newblock Quantum robust fitting.
\newblock In \emph{ACCV}, 2020.

\bibitem[Date and Potok(2021)]{DatePotok2021}
Prasanna Date and Thomas Potok.
\newblock Adiabatic quantum linear regression.
\newblock \emph{Scientific Reports}, 11:\penalty0 21905, 2021.

\bibitem[Doan et~al.(2022)Doan, Sasdelli, Suter, and Chin]{Doan2022}
Anh-Dzung Doan, Michele Sasdelli, David Suter, and Tat-Jun Chin.
\newblock A hybrid quantum-classical algorithm for robust fitting.
\newblock In \emph{CVPR}, pages 417--427, 2022.

\bibitem[Farhi et~al.(2001)Farhi, Goldstone, Gutmann, Lapan, Lundgren, and
  Preda]{Farhi2001}
Edward Farhi, Jeffrey Goldstone, Sam Gutmann, Joshua Lapan, Andrew Lundgren,
  and Daniel Preda.
\newblock A quantum adiabatic evolution algorithm applied to random instances
  of an np-complete problem.
\newblock \emph{Science}, 292\penalty0 (5516):\penalty0 472--475, 2001.

\bibitem[Farina et~al.(2023)Farina, Magri, Menapace, Ricci, Golyanik, and
  Arrigoni]{Farina2023}
Matteo Farina, Luca Magri, Willi Menapace, Elisa Ricci, Vladislav Golyanik, and
  Federica Arrigoni.
\newblock Quantum multi-model fitting.
\newblock In \emph{CVPR}, pages 13640--13649, 2023.

\bibitem[Fischler and Bolles(1981)]{FischlerBolles1981}
Martin~A. Fischler and Robert~C. Bolles.
\newblock Random sample consensus: a paradigm for model fitting with
  applications to image analysis and automated cartography.
\newblock \emph{Commun. ACM}, 24\penalty0 (6):\penalty0 381–395, 1981.

\bibitem[Golyanik and Theobalt(2020)]{golyanik2020quantum}
Vladislav Golyanik and Christian Theobalt.
\newblock A quantum computational approach to correspondence problems on point
  sets.
\newblock In \emph{CVPR}, pages 9182--9191, 2020.

\bibitem[Govindu(2005)]{Govindu05}
Venu~Madhav Govindu.
\newblock A tensor decomposition for geometric grouping and segmentation.
\newblock In \emph{CVPR}, pages 1150--1157, 2005.

\bibitem[{Gurobi Optimization, LLC}(2023)]{Gurobi2023}
{Gurobi Optimization, LLC}.
\newblock {Gurobi Optimizer Reference Manual}, 2023.

\bibitem[Isack and Boykov(2012)]{IsackBoykov12}
Hossam Isack and Yuri Boykov.
\newblock Energy-based geometric multi-model fitting.
\newblock In \emph{IJCV}, pages 123--147, 2012.

\bibitem[Jain and Govindu(2013)]{JainGovindu13}
Suraj Jain and Venu~Madhav Govindu.
\newblock Efficient higher-order clustering on the grassmann manifold.
\newblock In \emph{ICCV}, pages 3511--3518, 2013.

\bibitem[Kirkpatrick et~al.(1983)Kirkpatrick, Gelatt~Jr, and
  Vecchi]{Kirkpatrick1983}
Scott Kirkpatrick, C~Daniel Gelatt~Jr, and Mario~P Vecchi.
\newblock Optimization by simulated annealing.
\newblock \emph{science}, 220\penalty0 (4598):\penalty0 671--680, 1983.

\bibitem[{Krahn} et~al.(2024){Krahn}, {Sasdelli}, {Yang}, {Golyanik},
  {Kannala}, {Chin}, and {Birdal}]{Krahn2024}
Maximilian {Krahn}, Michelle {Sasdelli}, Fengyi {Yang}, Vladislav {Golyanik},
  Juho {Kannala}, Tat-Jun {Chin}, and Tolga {Birdal}.
\newblock {Projected Stochastic Gradient Descent with Quantum Annealed Binary
  Gradients}.
\newblock In \emph{BMVC}, 2024.

\bibitem[Li and Ghosh(2020)]{LiGhosh2020}
Junde Li and Swaroop Ghosh.
\newblock Quantum-soft qubo suppression for accurate object detection.
\newblock In \emph{ECCV}, 2020.

\bibitem[Magri(2015)]{MagriPhDThesis}
Luca Magri.
\newblock \emph{Multiple Structure Recovery via Preference Analysis in
  Conceptual Space}.
\newblock PhD thesis, University of Milan (Italy), 2015.

\bibitem[Magri and Fusiello(2014)]{MagriFusiello14}
Luca Magri and Andrea Fusiello.
\newblock {T-Linkage}: A continuous relaxation of {J-Linkage} for multi-model
  fitting.
\newblock In \emph{CVPR}, pages 3954--3961, 2014.

\bibitem[Magri and Fusiello(2016)]{Magri2016}
Luca Magri and Andrea Fusiello.
\newblock Multiple models fitting as a set coverage problem.
\newblock In \emph{CVPR}, pages 3318--3326, 2016.

\bibitem[Magri and Fusiello(2017)]{MagriFusiello17}
Luca Magri and Andrea Fusiello.
\newblock Multiple structure recovery via robust preference analysis.
\newblock \emph{Image and Vision Computing}, 67:\penalty0 1--15, 2017.

\bibitem[Magri et~al.(2021)Magri, Leveni, and Boracchi]{MagriLeveniAl21}
Luca Magri, Filippo Leveni, and Giacomo Boracchi.
\newblock Multilink: Multi-class structure recovery via agglomerative
  clustering and model selection.
\newblock In \emph{CVPR}, pages 1853--1862, 2021.

\bibitem[Meli et~al.(2022)Meli, Mannel, and Lellmann]{Meli_2022_CVPR}
Natacha~Kuete Meli, Florian Mannel, and Jan Lellmann.
\newblock An iterative quantum approach for transformation estimation from
  point sets.
\newblock In \emph{CVPR}, pages 529--537, 2022.

\bibitem[Meli et~al.(2025)Meli, Golyanik, Benkner, and Moeller]{meli2025qucoop}
Natacha~Kuete Meli, Vladislav Golyanik, Marcel~Seelbach Benkner, and Michael
  Moeller.
\newblock Qucoop: A versatile framework for solving composite and
  binary-parametrised problems on quantum annealers.
\newblock In \emph{CVPR}, 2025.

\bibitem[Purkait et~al.(2014)Purkait, Chin, Ackermann, and
  Suter]{PurkaitChinAl14}
Pulak Purkait, Tat-Jun Chin, Hanno Ackermann, and David Suter.
\newblock Clustering with hypergraphs: the case for large hyperedges.
\newblock In \emph{ECCV}, pages 672--687, 2014.

\bibitem[Sasdelli and Chin(2021)]{Sasdelli2021QuantumAF}
Michele Sasdelli and Tat-Jun Chin.
\newblock Quantum annealing formulation for binary neural networks.
\newblock In \emph{DICTA}, pages 1--10, 2021.

\bibitem[Schuld et~al.(2016)Schuld, Sinayskiy, and Petruccione]{Schuld2016}
Maria Schuld, Ilya Sinayskiy, and Francesco Petruccione.
\newblock Prediction by linear regression on a quantum computer.
\newblock \emph{Phys. Rev. A}, 94:\penalty0 022342, 2016.

\bibitem[Seelbach~Benkner et~al.(2020)Seelbach~Benkner, Golyanik, Theobalt, and
  Moeller]{seelbach20quantum}
Marcel Seelbach~Benkner, Vladislav Golyanik, Christian Theobalt, and Michael
  Moeller.
\newblock Adiabatic quantum graph matching with permutation matrix constraints.
\newblock In \emph{3DV}, pages 583--592, 2020.

\bibitem[{Seelbach Benkner} et~al.(2021){Seelbach Benkner}, {L\"{a}hner},
  {Golyanik}, {Wunderlich}, {Theobalt}, and {Moeller}]{SeelbachBenkner2021}
Marcel {Seelbach Benkner}, Zorah {L\"{a}hner}, Vladislav {Golyanik}, Christof
  {Wunderlich}, Christian {Theobalt}, and Michael {Moeller}.
\newblock Q-match: Iterative shape matching via quantum annealing.
\newblock In \emph{ICCV}, pages 7586--7596, 2021.

\bibitem[Tepper and Sapiro(2017)]{TepperSapiro17}
Mariano Tepper and Guillermo Sapiro.
\newblock Nonnegative matrix underapproximation for robust multiple model
  fitting.
\newblock In \emph{CVPR}, pages 2059--2067, 2017.

\bibitem[Toldo and Fusiello(2008)]{ToldoFusiello08}
Roberto Toldo and Andrea Fusiello.
\newblock Robust multiple structures estimation with {J-Linkage}.
\newblock In \emph{ECCV}, pages 537--547, 2008.

\bibitem[Vincent and Laganiere(2001)]{VincentLaganiere01}
Esther Vincent and Robert Laganiere.
\newblock Detecting planar homographies in an image pair.
\newblock In \emph{ISPA}, pages 182--187, 2001.

\bibitem[Wang et~al.(2018)Wang, Xiao, Yan, and Suter]{WangGupbao18}
Hanzi Wang, Guobao Xiao, Yan Yan, and David Suter.
\newblock Searching for representative modes on hypergraphs for robust
  geometric model fitting.
\newblock \emph{IEEE TPAMI}, 41\penalty0 (3):\penalty0 697--711, 2018.

\bibitem[Willsch et~al.(2020)Willsch, Madita, Raedt, and
  Michielsen]{Willsch2020}
Dennis Willsch, Willsch Madita, Hans~De Raedt, and Kristel Michielsen.
\newblock {S}upport vector machines on the {D}-{W}ave quantum annealer.
\newblock \emph{Computer physics communications}, 248:\penalty0 107006 --,
  2020.

\bibitem[Wong et~al.(2011)Wong, Chin, Yu, and Suter]{Wong2011}
Hoi~Sim Wong, Tat-Jun Chin, Jin Yu, and David Suter.
\newblock Dynamic and hierarchical multi-structure geometric model fitting.
\newblock In \emph{ICCV}, pages 1044--1051, 2011.

\bibitem[Xu et~al.(1990)Xu, Oja, and Kultanen]{XuOjaAl90}
Lei Xu, Erkki Oja, and Pekka Kultanen.
\newblock A new curve detection method: randomized {H}ough transform ({RHT}).
\newblock \emph{Pattern recognition letters}, 11\penalty0 (5):\penalty0
  331--338, 1990.

\bibitem[Yang et~al.(2024)Yang, Sasdelli, and Chin]{Yang2024}
Frances~Fengyi Yang, Michele Sasdelli, and Tat-Jun Chin.
\newblock Robust fitting on a gate quantum computer.
\newblock In \emph{ECCV}, 2024.

\bibitem[Zaech et~al.(2022)Zaech, Liniger, Danelljan, Dai, and
  Van~Gool]{Zaech_2022_CVPR}
Jan-Nico Zaech, Alexander Liniger, Martin Danelljan, Dengxin Dai, and Luc
  Van~Gool.
\newblock Adiabatic quantum computing for multi object tracking.
\newblock In \emph{CVPR}, pages 8811--8822, 2022.

\bibitem[Zass and Shashua(2005)]{ZassShashua05}
Ron Zass and Amnon Shashua.
\newblock A unifying approach to hard and probabilistic clustering.
\newblock In \emph{ICCV}, pages 294--301, 2005.

\bibitem[Zuliani et~al.(2005)Zuliani, Kenney, and Manjunath]{ZulianiKenneyAl05}
Marco Zuliani, Charles~S Kenney, and BS Manjunath.
\newblock The multi{RANSAC} algorithm and its application to detect planar
  homographies.
\newblock In \emph{ICIP}, pages III--153--6, 2005.

\end{thebibliography}
}


\clearpage
\maketitlesupplementary

This document provides additional details on our Robust Quantum Multi-Model Fitting algorithm and its decomposed version  (Sec.~\ref{sec:AlgDetails}). 
It also provides further details and visualizations for the experiments (Sec.~\ref{sec:AdditionalExperiments}) presented in the main paper.

\section{Algorithmic Details}\label{sec:AlgDetails}
\paragraph{Decomposed R-QuMF.} 
Real-world size problems are currently intractable on a modern AQC, since the amount of physical qubits required to map logical qubits increases super-linearly. %
Our decomposed approach, following Farina \textit{et al.}~\cite{Farina2023}, mitigates this issue by decomposing the preference-consensus matrix $P$ (\emph{i.e.}, consensus set of sampled models) into manageable sub-matrices with at most $s$ columns 
(\emph{i.e.}~sub-problem size) that can be confidently sampled on modern quantum hardware using RQuMF. 
Alg.~\ref{algo:derqumf_algorithm} summarises the decomposed version of our approach that we call De-RQuMF. 
    \begin{algorithm}[h!]
    \caption{De-RQuMF Method 
    }
    \label{algo:derqumf_algorithm}
    \begin{algorithmic}[1]
    \State \textbf{Input:} Point set $X$, problem size $s$, inlier threshold $\epsilon$ 
    \State \textbf{Output:} Predicted labels $l$
    \Statex \underline{Generate preference consensus matrix}
    \State $M \leftarrow k \times |X|$, initialize $P$ as a $|X| \times M$ zero matrix
    \For{$i = 0$ to $M-1$}
        \State Sample points from $X$
        \State fit geometric model
        \State compute residuals $R \forall \{(x_i,y_i)\} \in X$ 
        \State Update $P[:,i] \leftarrow [r < \epsilon  ?  1 : 0 \; \forall r \in R]$
        \State $i \leftarrow i + 1$
    \EndFor
    \Statex \underline{Logical graph mapping}
    \While{$|\text{columns}(P)| > s$}
        \State Partition $P$ into $L$ subproblems $\{P_j\}$ of size $s$
        \For{$j = 0$ to $L$}
            \State $z$ = RQuMF($P_j$)
            \Procedure{RQuMF}{$P_j$}
                \State $A \leftarrow$ concatenate $[-I; P_j]$
                \State $\widetilde{Q} \leftarrow \lambda_2 \times A^TA$
                \State $\widetilde{s} \leftarrow$ concatenate $[-\mathbf{1}_N; \lambda_1 \times \mathbf{1}_M]$
                \State $z$ = solveQUBO($\widetilde{Q}, \widetilde{s}$) \State return $z$
            \EndProcedure
            \State retain $z$ from $P_{j}$ 
            \State $j \leftarrow j + 1$
        \EndFor
    \EndWhile
    \State $z$ = RQuMF($P$)
    \Statex \underline{Final model selection}
    \State return $z$
    \Statex \underline{Generate label assignment}
    \For{each model $z_i$ in $z$}
      \For{each point $x_j$ in consensus set of $z_i$}
       \State  Assign label $i$ to $x_j$\;
       \EndFor
    \EndFor
    \State Solve a linear assignment problem to maximize label coverage across $X$\;
    \State return  predicted labels $l$\;
\end{algorithmic}
\end{algorithm}

The algorithm takes in input a dataset $X$, a sub-problem size $s$, and inlier thresholds $\epsilon$. It returns a set of labels $l$ corresponding to a cover of $X$ according to the retrieved models. The parameter $s$ controls how many sampled models are processed in each iteration of the decomposed method.
 The first step (lines $3{-}10$) consists of generating a pool of $M$ tentative models via random sampling and of computing their consensus set. The number of hypothesis $M$ is defined as a multiple of the number of input points (line $3$). Minimal sample sets are sampled from $X$ using localized sampling (line $5$) and used to fit a geometric model (line $6$). Hence residuals are computed (line $7$). Residuals smaller than the inlier threshold $\epsilon$ define the consensus sets of each sampled model, which are stored as columns in the preference matrix (line $8$). The process is repeated until $M$ models are sampled (line $4$).

 The second step (lines $11-25$) involves the decomposition of the preference matrix $P$ to define the logical graph mapping. Specifically, $P$ is partitioned into $L$ sub-block $P_j$ having at most $s$ columns each (line $12$). 
 Each sub-problem $P_j$ is converted to a QUBO form (lines $13-21$) with its logical graph 
 and hence solved using RQuMF (line $19$). Each solution $z$ represents the selected models, indicated by the corresponding columns of $P_j$. In the pruning phase (line $22$), the process retains only the chosen models, while the rest are eliminated, thereby reducing the dimensionality of  $P_{j}$.
 Once the overall number of retained models falls below $s$, a final execution of the RQuMF method is carried out (line $26$) to derive the final solution. The models selected in this last iteration of RQuMF (line 26) undergo label assignment (line $29$) where the consensus set corresponding to each model is labelled with the $i^{\text{th}}$ index of the model in question. Subsequently, to optimise the coverage of these labels across all data points, a linear assignment problem is tackled (line $33$) yielding the final labels  $l$.%

\section{Additional Details On Experiments}\label{sec:AdditionalExperiments}
We provide here additional details about the experiments performed in Sec.~\ref{sec:Experiments} of the main paper. 

\begin{figure}[h!]
  \centering
  \includegraphics[width=\columnwidth]{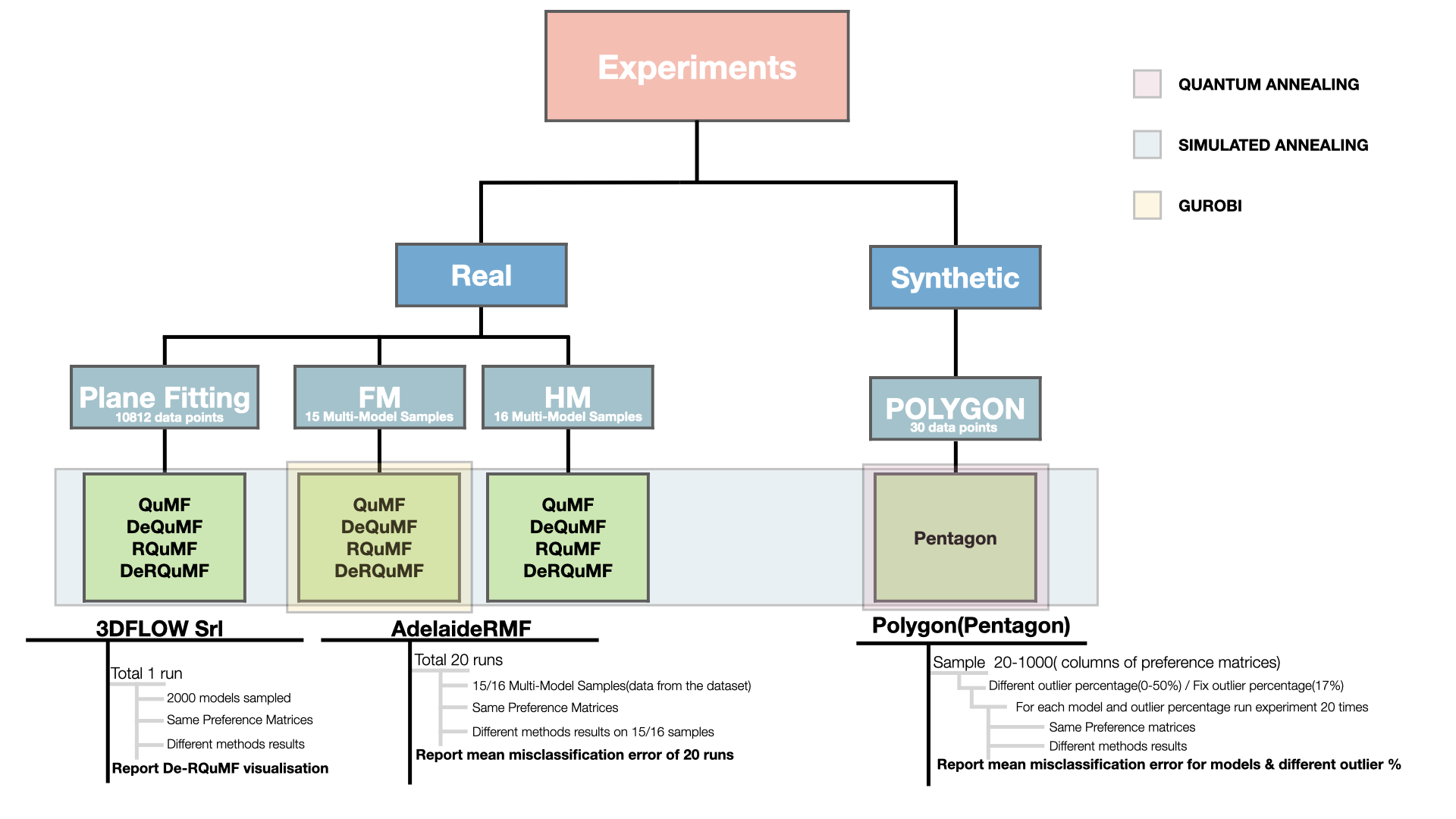}
  \caption{Overview of our experiments. ``FM'' refers to datasets based on the fundamental matrix model whereas ``HM'' refers to the homography model.
  }
  \label{fig:experiment_overview}
\end{figure}

\paragraph{Experiments Overview.} In  Fig.~\ref{fig:experiment_overview} we report a general overview of the experiments  conducted on real and synthetic datasets, highlighting the different solvers used (\emph{i.e.}, QA, SA etc) and the different configurations adopted to test scalability and robustness of our proposed methods.

\begin{figure}[h!]
    \centering
    \begin{subfigure}{0.5\columnwidth}
        \centering
        \includegraphics[width=\textwidth]{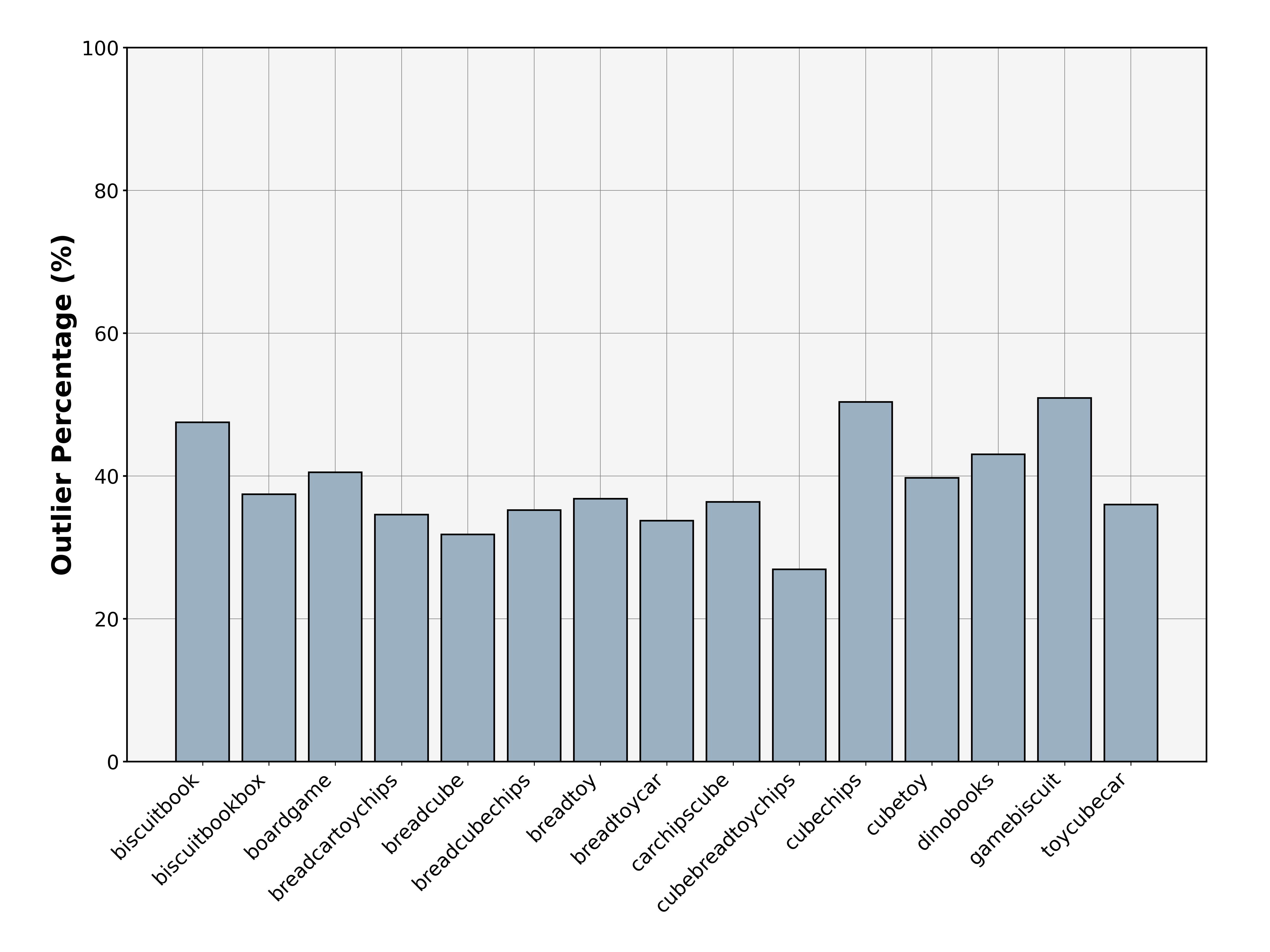}
    \end{subfigure}%
    \begin{subfigure}{0.5\columnwidth}
        \centering
        \includegraphics[width=\textwidth]{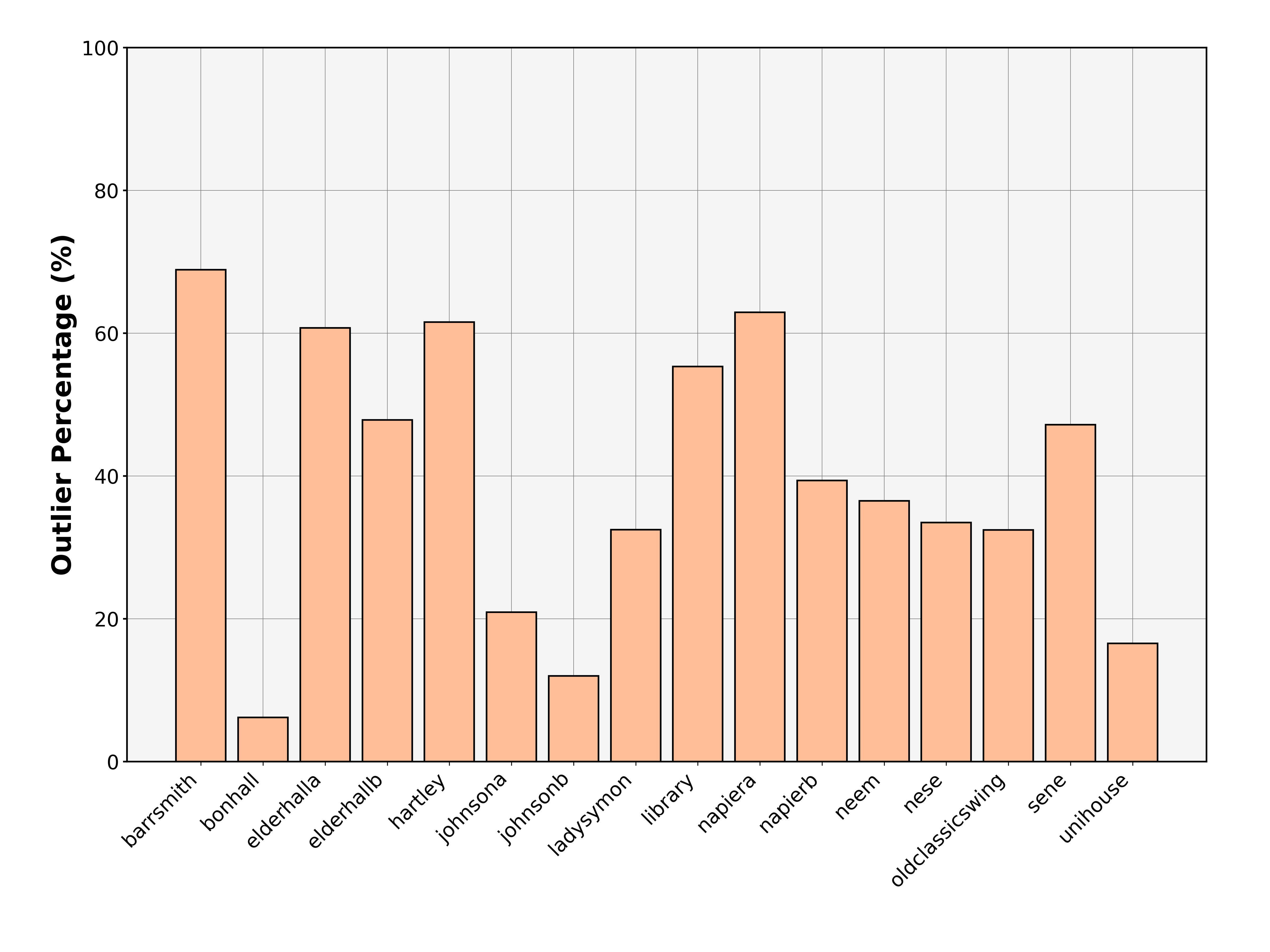}
    \end{subfigure}
    \caption{Outlier Percentage of each sequence in AdelaideRMF dataset \cite{Wong2011}. Left: 15 image pairs for Fundamental Matrices fitting, Right: 16 image pairs for homographies. }
    \label{fig:fm_hm_outlier_percentage}
\end{figure}

\paragraph{Outliers Percentage In Real Data Multi-Model Fitting Tasks.} The plots in Fig.~\ref{fig:fm_hm_outlier_percentage} show the outlier percentage in multiple fundamental matrices and homography fitting problems respectively, as referred in \ref{sec:experiments:real}.  Outliers typically correspond to wrong key-point matches that cannot be described by any model.
It can be appreciated that in most of the pairs related to fundamental matrices (which are related to motion segmentation in two images) the outlier ratio is greater than $30\%$, while for homographies (which are related to plane fitting) we have that in $5$ pairs out of $16$, there are more than $50\%$ of outliers, making the problem particularly challenging. This justifies the higher errors reported in plane fitting with respect to the one attained on motion segmentation.

\begin{figure}[h!]
    \centering
    \begin{subfigure}{0.48\columnwidth}
        \centering
        \includegraphics[width=\textwidth]{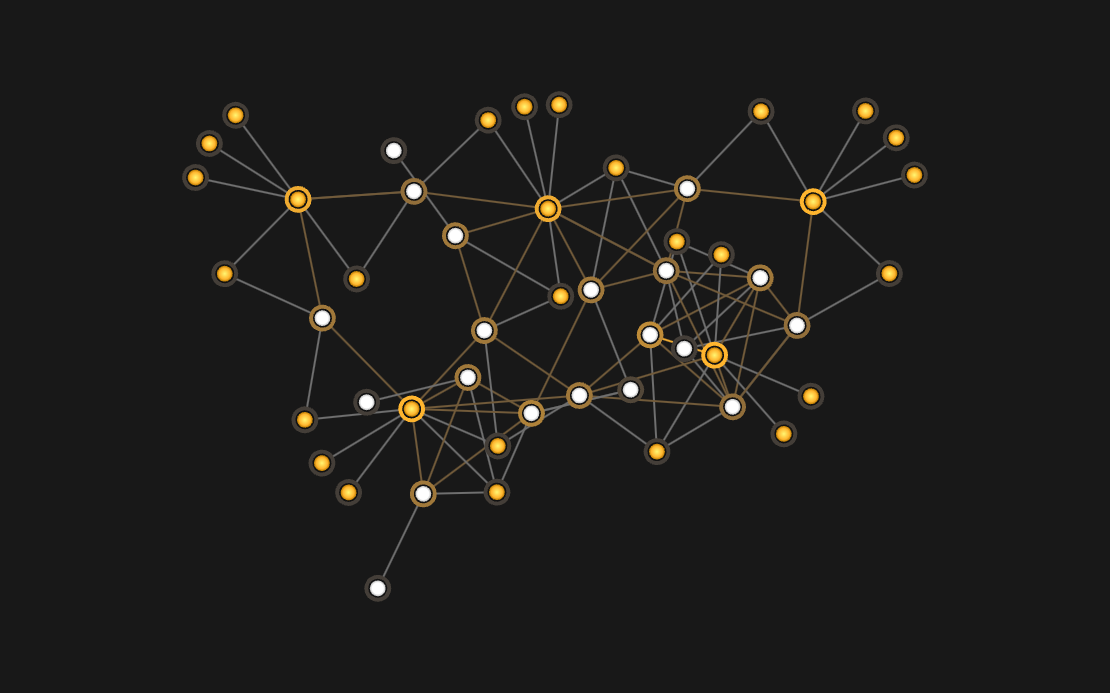}
        \caption{50 Logical Qubits}
    \end{subfigure}
    \hfill
    \begin{subfigure}{0.48\columnwidth}
        \centering
        \includegraphics[width=\textwidth]{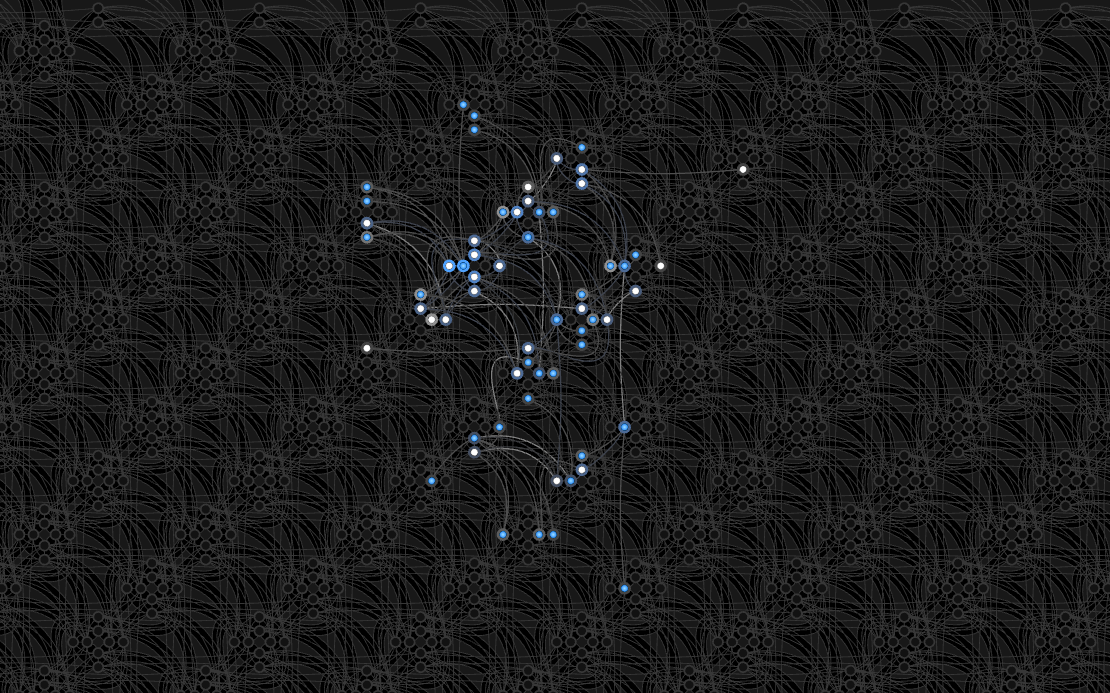}
        \caption{62 Physical Qubits}
    \end{subfigure}
    \begin{subfigure}{0.48\columnwidth}
        \centering
        \includegraphics[width=\textwidth]{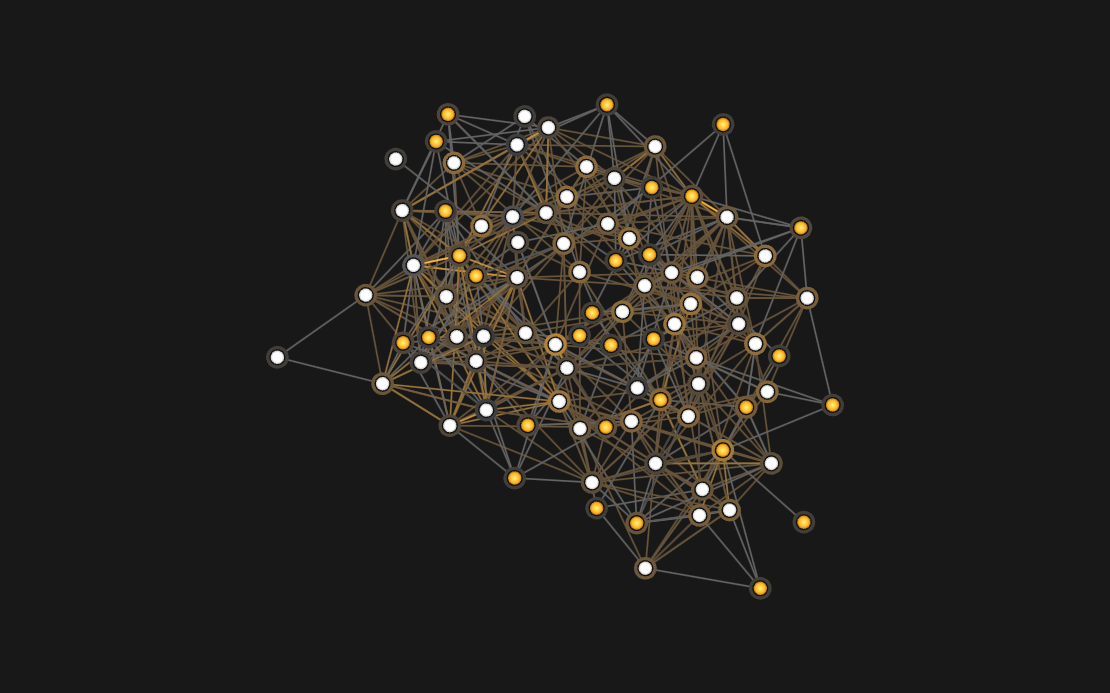}
        \caption{90 Logical Qubits}
    \end{subfigure}
    \hfill
    \begin{subfigure}{0.48\columnwidth}
        \centering
        \includegraphics[width=\textwidth]{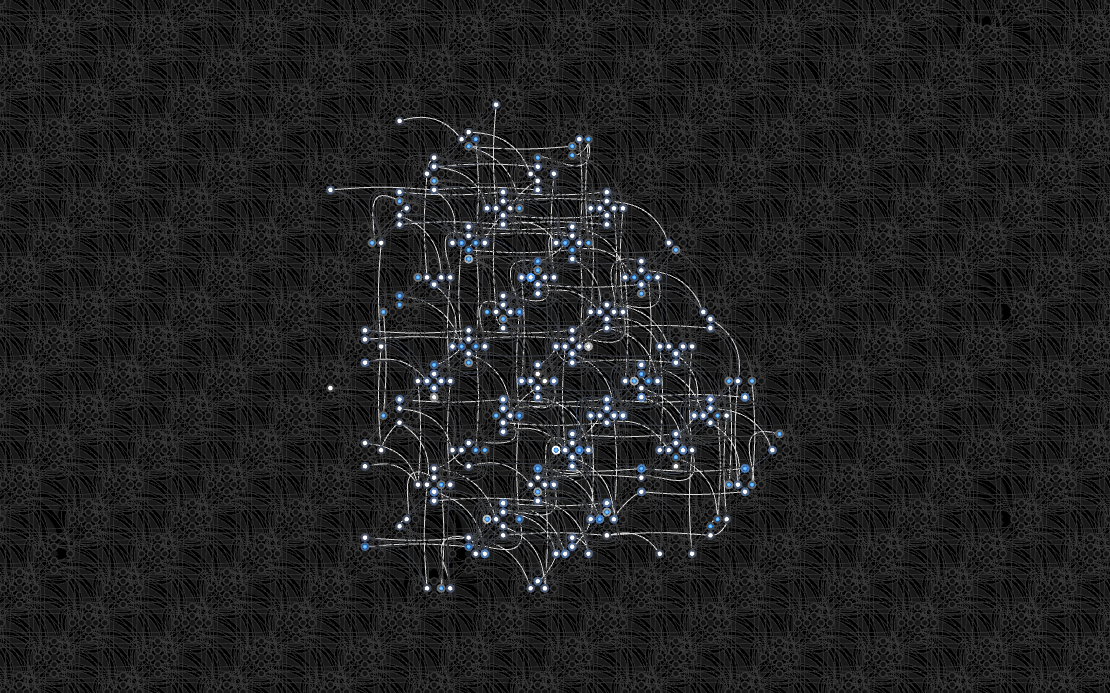}
        \caption{273 Physical Qubits}
    \end{subfigure}
    \begin{subfigure}{0.48\columnwidth}
        \centering
        \includegraphics[width=\textwidth]{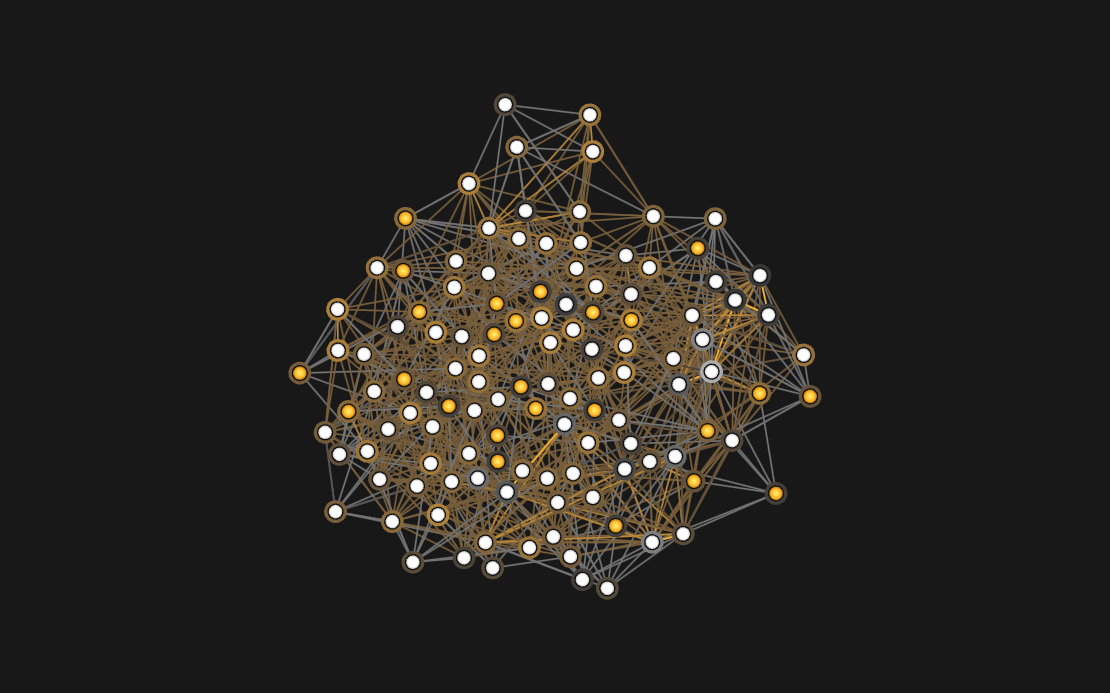}
        \caption{120 Logical Qubits}
        \end{subfigure}
    \hfill
    \begin{subfigure}{0.48\columnwidth}
        \centering
        \includegraphics[width=\textwidth]{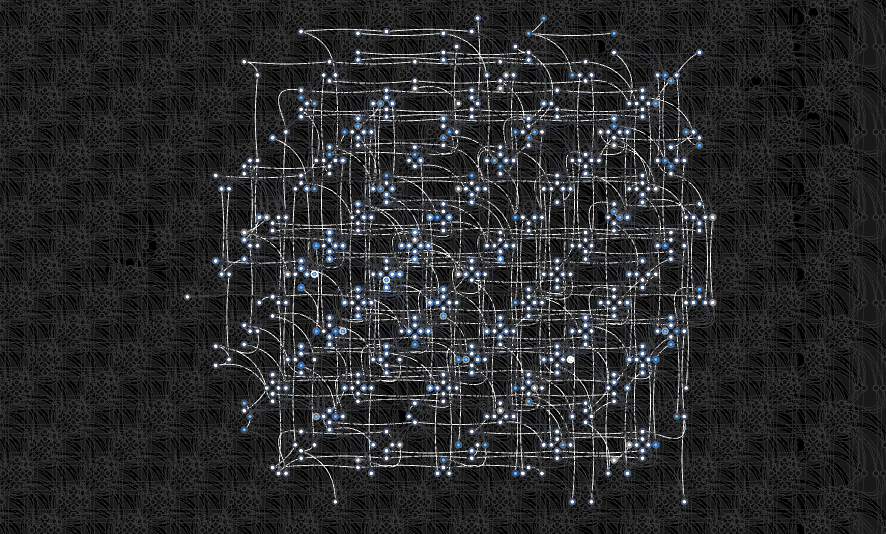}
        \caption{602 Physical Qubits}
    \end{subfigure}
    \caption{\textbf{Left:} These images depict logical graphs for different problem sizes, reported in Fig.~\ref{fig:synthetic_vizualisation} of the main manuscript. \textbf{Right:} The corresponding physical qubit embedding of logical graphs (a, c, d) respectively using Pegasus topology on DWave Advantage5.4 as mentioned in Sec.~\ref{ssec:Implementation} of the main paper. }
    \label{fig:quantum_expermient_logical_to_physical}
\end{figure}

\paragraph{Logical and physical graphs.} In  Fig.~\ref{fig:quantum_expermient_logical_to_physical} {we report the logical and physical graphs} corresponding to sample problems related to the scalability experiment on the synthetic dataset, where we maintain a constant outlier ratio of $17\%$ while expanding the sampled model size from 20 to 140. The left-side images showcase the logical graph representation of the problem, where each node corresponds to a logical qubit and the edges depict the coupling between these qubits. Through minor embedding, these logical qubits are mapped onto physical qubits within the quantum hardware. On the right side, the physical representation of these mappings is displayed, with each node representing a physical qubit. The colour inside the node reveals the measured value in its most stable energy state, while the colour of the node's outer ring indicates the direction of bias, specifically pointing out whether the coefficient of the linear component in the optimization is positive or negative.

\begin{figure}[h!]
  \centering
  \includegraphics[width=\columnwidth]{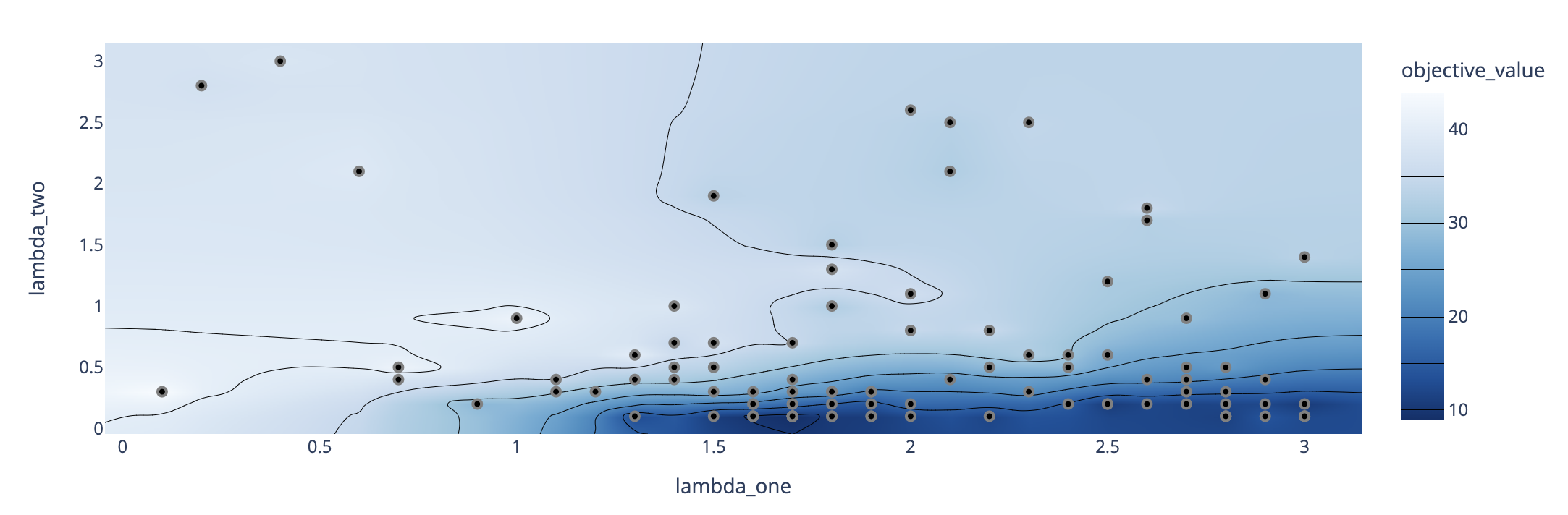}
  \caption{Lambda optimisation contour plot of RQuMF for fundamental matrix data ($\lambda_{1} = 1.7 \, , \lambda_{2} = 0.1 $) using TPE.}
  \label{fig:parameter_tunning}
\end{figure}

\paragraph{Hyperparameter Tuning.}
We studied the effect of hyperparameter tuning in our experiments. Specifically, we studied how the objective value defined in Eq.~\eqref{eq25} changes with respect to $\lambda_1, \lambda_2$. %
In Fig.~\ref{fig:parameter_tunning} we plot the objective landscape for fundamental matrix (FM) fitting problems, using the Tree-structured Parzen Estimator (TPE) discussed in Sec.~\ref{ssec:Implementation}. 
It can be appreciated that the TPE strategy discards suboptimal values of the parameters and concentrates more on the ones that result in lower values (i.e, a cluster of points near the best objective values), contrary to grid search which doesn't consider the previous results for selecting parameters for the future.
Additionally, in relation with the Tab.~\ref{tab:fm_outlier_detection_comparison_SA} the performance gap between QuMF and our method in the outlier-free setting can be narrowed by adjusting the lambda parameters to suit this specific case. For instance, when optimized lambda values ($\lambda_1 = 4.8$ and $\lambda_2 = 0.6$) are used for the fundamental matrix estimation task in the absence of outliers, the misclassification error for RQuMF decreases to $2.05$, while for De-RQuMF, it drops to $6.18$. Conversely, these parameters are not ideal for outlier-prone scenarios, where misclassification rates increase from $10.46$ to $16.95$ for RQuMF and from $12.69$ to $15.75$ for De-RQuMF. Therefore, we recommend adjusting the lambda parameters based on the specific conditions to achieve optimal results.

\begin{figure}[h!]
  \centering
  \includegraphics[width=\columnwidth]{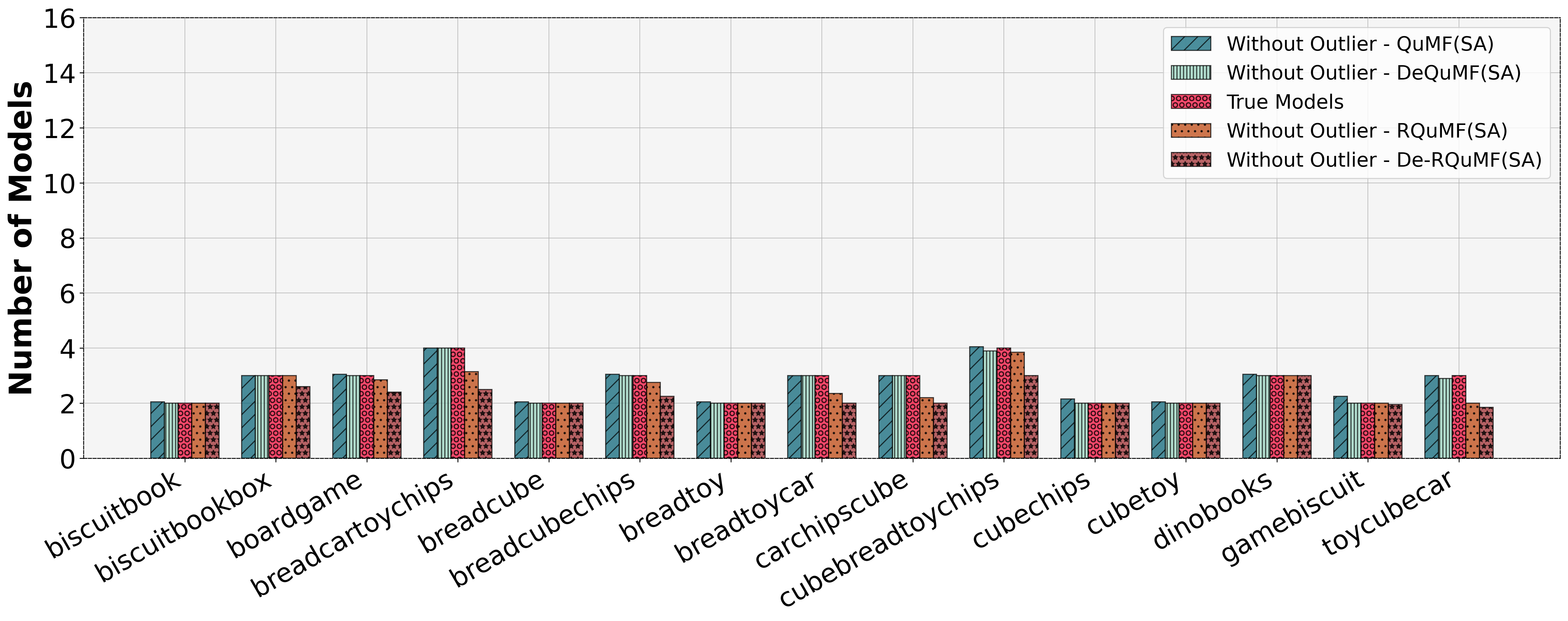}
  \caption{Average number of models selected by different methods for 15 multimodel sequences from AdelaideRMF \cite{Wong2011} dataset for fundamental matrix fitting in the absence of outliers. The middle bar in the grouped bars represents the ground truth number of models, the left two bars represent QuMF, DeQuMF respectively and the right bars represent our proposed methods. }
  \label{fig:nom_without_outliers}
\end{figure}

\begin{figure}[h!]
  \centering
  \includegraphics[width=\columnwidth]{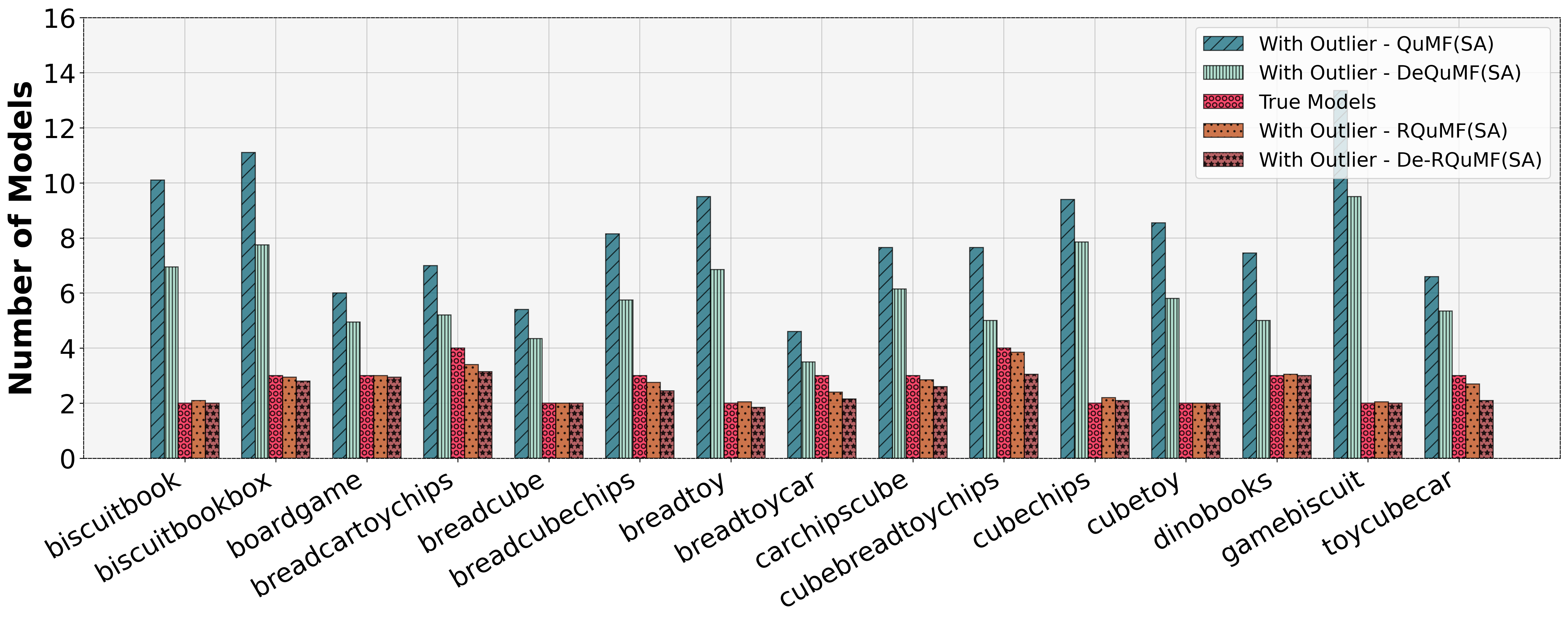}
  \caption{Average number of models selected by different methods for 15 multimodel sequences from AdelaideRMF \cite{Wong2011} dataset for fundamental matrix fitting in the absence of outliers. The middle bar in the grouped bars represents the ground truth number of models, the left two bars represent QuMF, DeQuMF respectively and the right bars represent our proposed methods.} 
  \label{fig:nom_with_outliers}
\end{figure}

\paragraph{Number Of Selected Models.}
Our QUBO formulation does not require knowing in advance the number of models, thus we assess whether the number of estimated models matches the ground truth ones.
Results are reported for fundamental matrix fitting problems in different setups, \emph{i.e.}, without outliers in Fig.~\ref{fig:nom_without_outliers} and with outliers in Fig.~\ref{fig:nom_with_outliers}.  We compare the estimated number of models attained by QuMF, DeQuMF, RQuMF, and De-RQuMF. The plots show that our methods mostly select the true number of models, in contrast to previous methods which estimate the right number of models in the outlier-free scenario,  but, {without any kind of post-processing} are prone to over-estimation in the presence of outliers. This is expected as, being based on set-cover rather than on maximum coverage, QuMF and DeQuMF try to maximize the number of inliers at the cost of hallucinating more models in the solution. It is worth noting, that even when coupled with post-processing, \emph{i.e.}, after providing the right number of models, QuMF failed to achieve competitive results compared to our methods, highlighting the fact that these methods, in the presence of outliers, can not segment the data at the first place. See also Tab.~\ref{tab:hm_outlier_detection_comparison_SA} from the main paper.


 \paragraph{Execution Times.}
 Tab.~\ref{tab:methods_benchmarking} presents the average execution time per sample of different methods on the AdelaideRMF \cite{Wong2011} dataset. It can be noticed that  DeQuMF by far outperforms in terms of execution time among all the methods confirming the advantages of the decomposed approach. Our formulation is less efficient as the dimension of the $Q$ matrix has to encode also the number of points, while in the previous approach, $Q$ scales with the number of sampled models. %
 For reference,  \textit{biscuitbook} of the fundamental matrix fitting dataset has 341 data points. In one of the runs, the corresponding $Q$ matrix dimension and node count in the logical graph are $ (2046, 2046), 1793282$  for QuMF, $ (40, 40), 744$ for DeQuMF,  $(2387, 2387), 1890341$ for RQuMF,  $(381, 381), 3059$ for DeRQuMF respectively. Thus De-RQuMF has to solve a problem of almost $\times 5$ bigger than DeQuMF (in terms of the logical graph). 
 It's crucial to emphasise that the extended execution time of our method significantly enhances the reliability of the results, contrary to the DeQuMF approach, which exhibits limitations as detailed in Tab.~\ref{tab:methods_benchmarking} of the main paper, our proposed method delivers reliable performance without failure. 
 Additionally, we omit the reporting of methods execution time on quantum hardware, as the anneal time remains constant at $20\mu s$, independent of problem size. Note that, with the advent of stable Adiabatic Quantum Computers (AQC), our approach is not only expected to become significantly faster but also stay reliable.
\begin{table}[h!]
    \centering
    \begin{tabular}{l|c|c}
        \toprule
        \textbf{Method} & \textbf{FM}  & \textbf{HM} \\
        \midrule
        QuMF(SA) \cite{Farina2023}   & 45.88  & 81.11\\
        DeQuMF(SA) \cite{Farina2023}   & \textbf{4.40}  & \textbf{4.93} \\
        RQuMF(SA)(ours)  & 51.61  & 130.10 \\
        DeRQuMF(SA)(ours) & 21.48 & 77.99 \\
        \bottomrule
    \end{tabular}
    \caption{Execution time (in seconds) of methods on different real datasets using Apple Silicon M1 machine with 8GB RAM.} 
    \label{tab:methods_benchmarking}
\end{table}

    \begin{figure*}
        \centering
        \begin{subfigure}{0.48\textwidth}
            \centering
            \includegraphics[width=\textwidth]{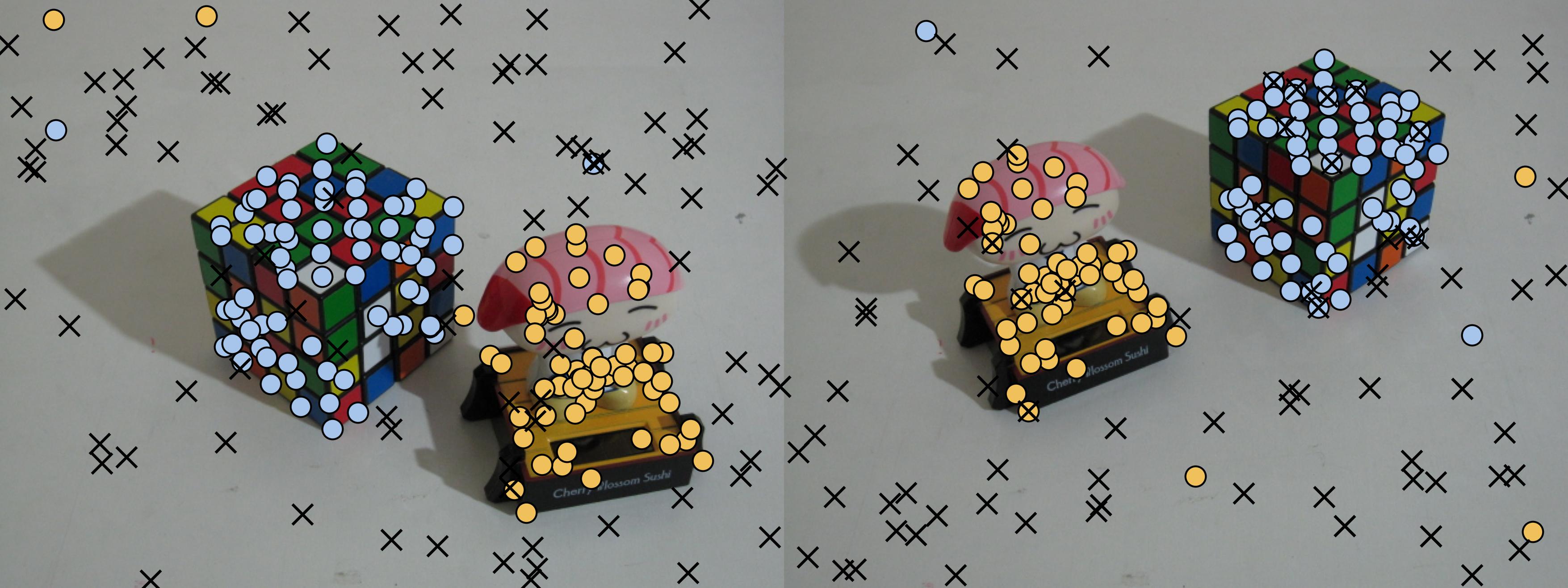}
        \end{subfigure}
        \hfill
        \begin{subfigure}{0.48\textwidth}
            \centering
            \includegraphics[width=\textwidth]{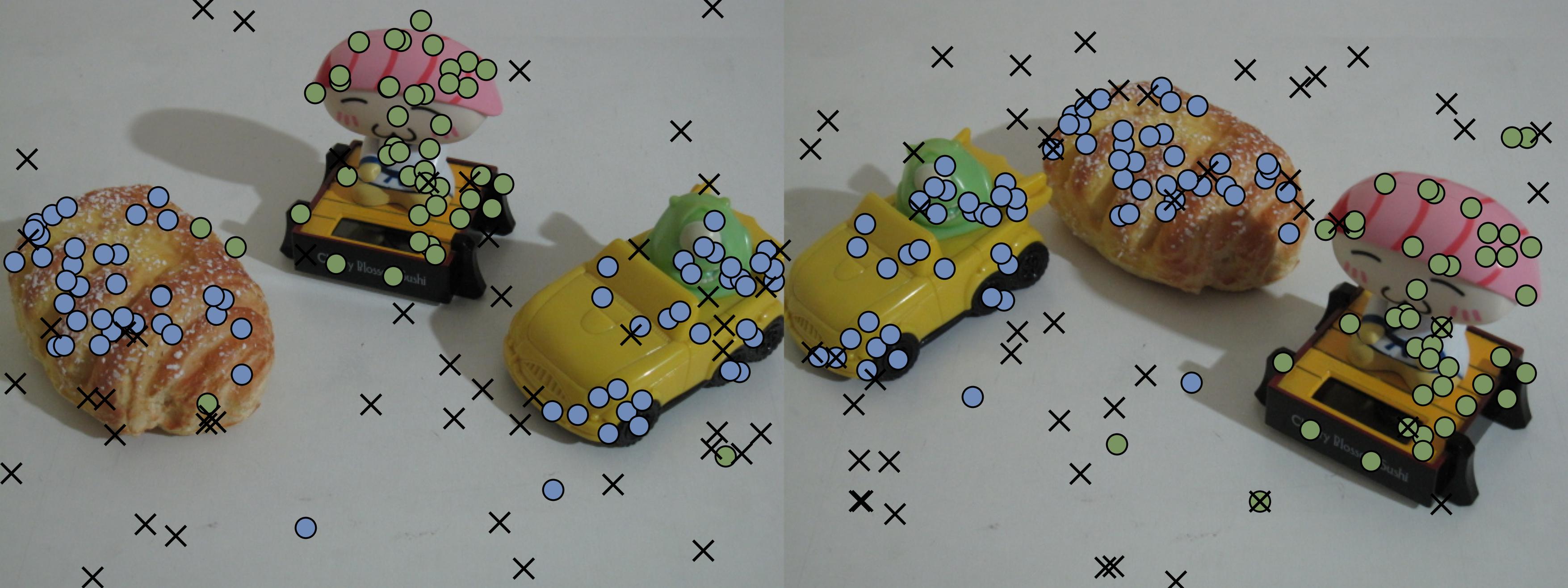}
        \end{subfigure}
        \caption{\textbf{Left:}  A sample of the \textit{best-case} for RQuMF on the \textit{cubetoy} (average $E_{mis} = 3.73\%$) sequence of the AdelaideRMF \cite{Wong2011} dataset for fundamental matrix (for the same sample average $E_{mis}$ for De-RQuMF, QuMF and DeQuMF is $4.13\%, 42.95\%, 23.71\%$ respectively).
        \textbf{Right:}  A sample of the \textit{worst-case} for RQuMF on the \textit{breadtoycar} (average $E_{mis} = 21.02\%$) sequence of the AdelaideRMF \cite{Wong2011} dataset for fundamental matrix (for the same sample average $E_{mis}$ for De-RQuMF, QuMF and DeQuMF is $26.23\%, 35.81\%, 21.95\%$ respectively).} 
        \label{fig:qualitative_supp1}
    \end{figure*}
    
    \begin{figure*}
        \centering
        \begin{subfigure}{0.48\textwidth}
            \centering
            \includegraphics[width=\textwidth]{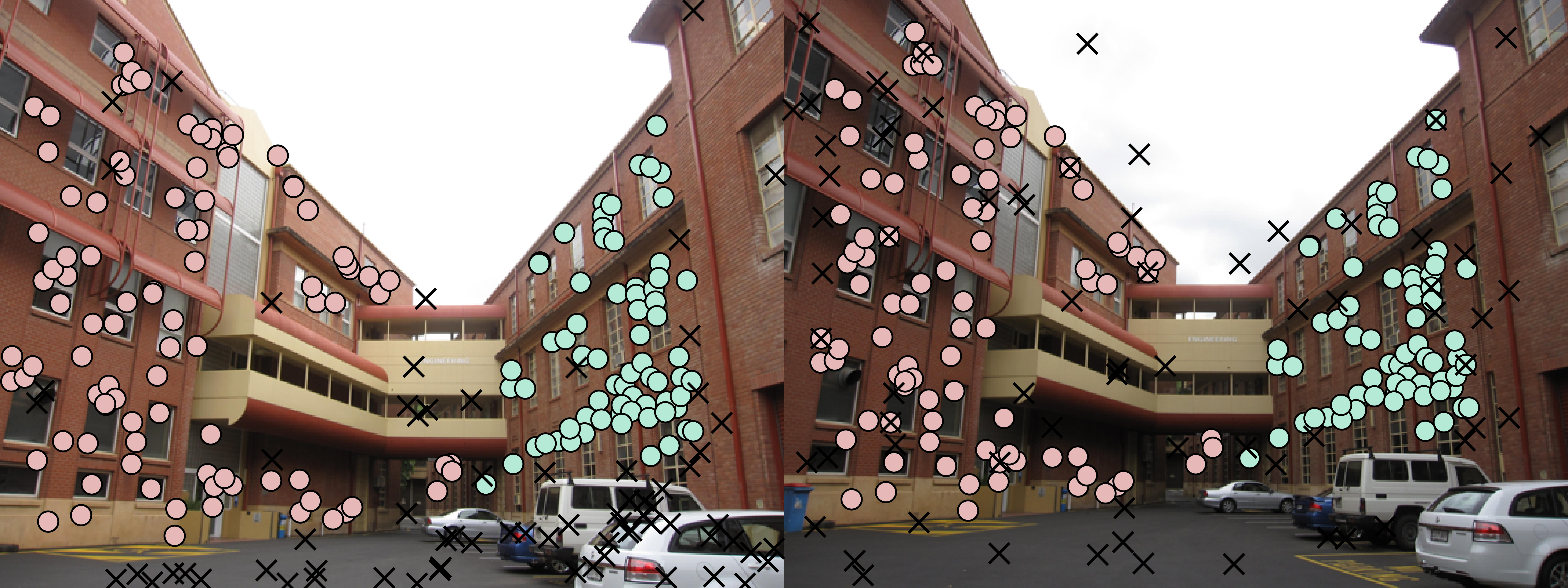}
        \end{subfigure}
        \hfill
        \begin{subfigure}{0.48\textwidth}
            \centering
            \includegraphics[width=\textwidth]{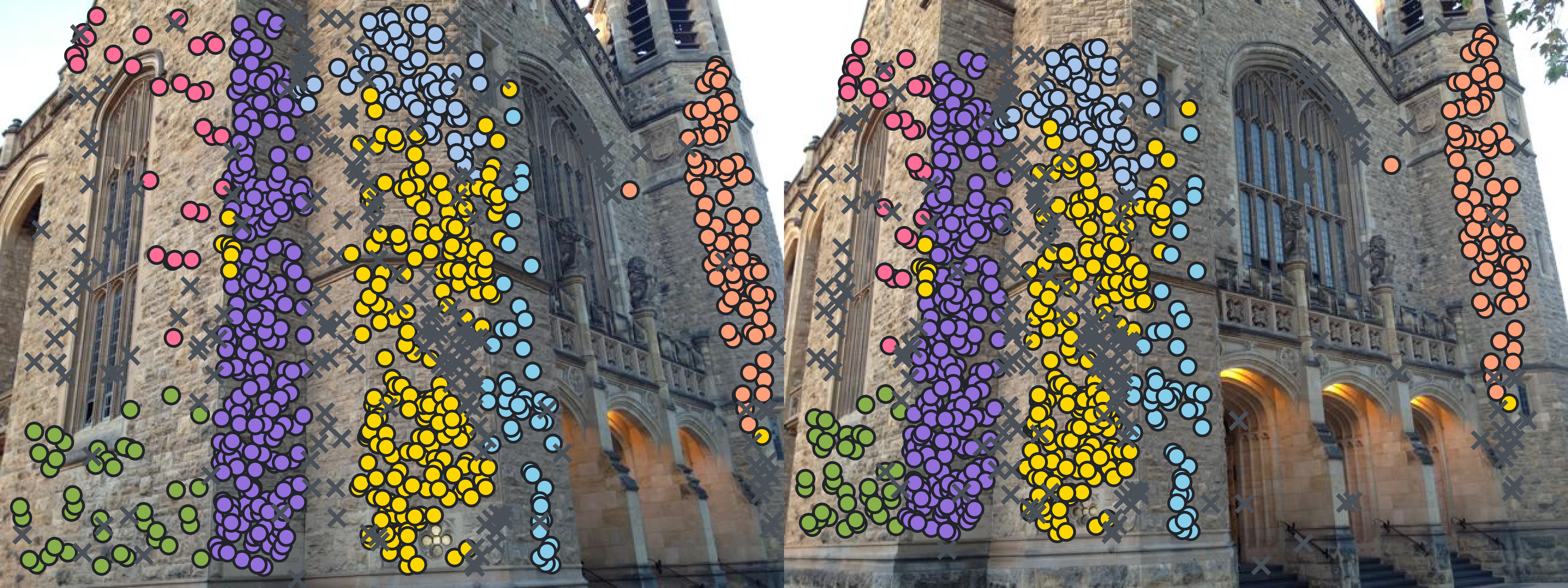}
        \end{subfigure}
        \caption{\textbf{Left:}  A sample of the \textit{best-case} for RQuMF on the \textit{nese} (average $E_{mis} = 1.92\%$) sequence of the AdelaideRMF \cite{Wong2011} dataset for homography matrix (for the same sample average $E_{mis}$ for De-RQuMF, QuMF and DeQuMF is $2.14\%, 77.70\%, 28.29\%$ respectively).
        \textbf{Right:}  A sample of the \textit{worst-case} for RQuMF on the \textit{bonhall}(average $E_{mis} = 41.60\%$) sequence of the AdelaideRMF \cite{Wong2011} dataset for homography matrix (for the same sample average $E_{mis}$ for De-RQuMF, QuMF and DeQuMF is $23.13\%, 74.20\%, 25.63\%$ respectively).} 
        \label{fig:qualitative_supp2}
    \end{figure*}

\begin{figure*}[htbp!]
        \centering
        \begin{subfigure}[b]{0.48\linewidth}
            \includegraphics[width=\linewidth]{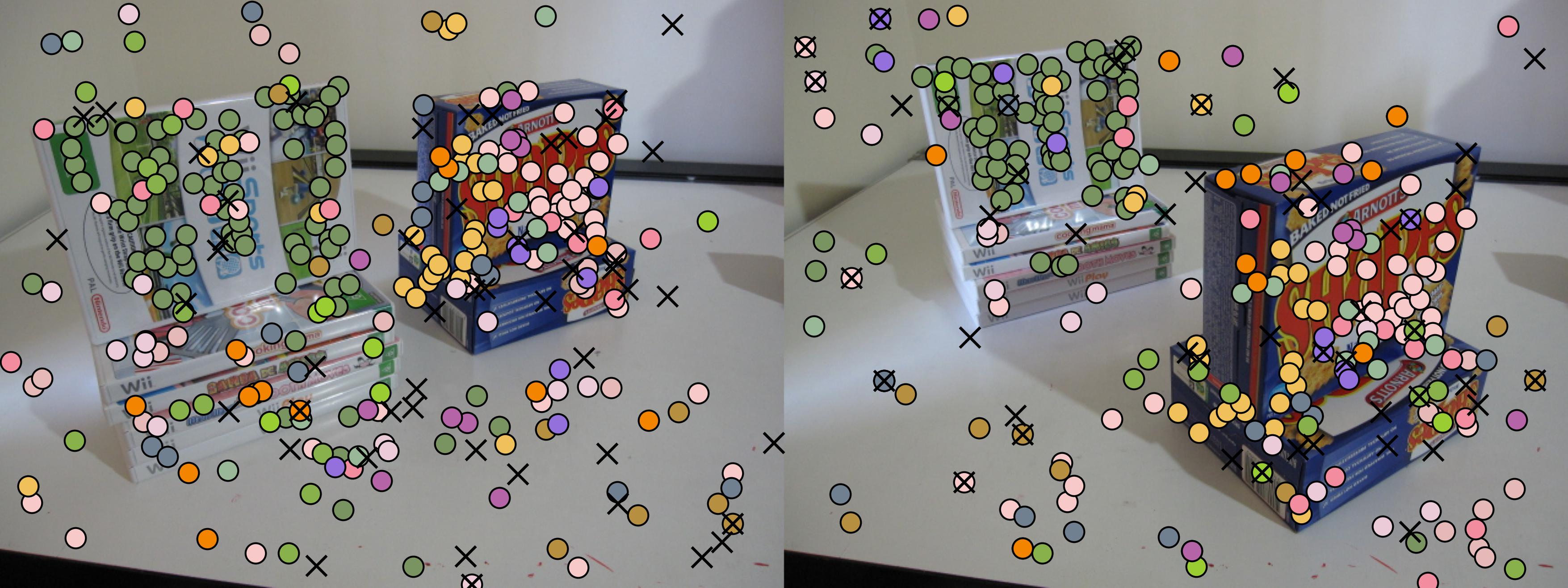}
            \caption{QuMF, $E_{mis} = 53.35\%$}
        \end{subfigure}
        \begin{subfigure}[b]{0.48\linewidth}
            \includegraphics[width=\linewidth]{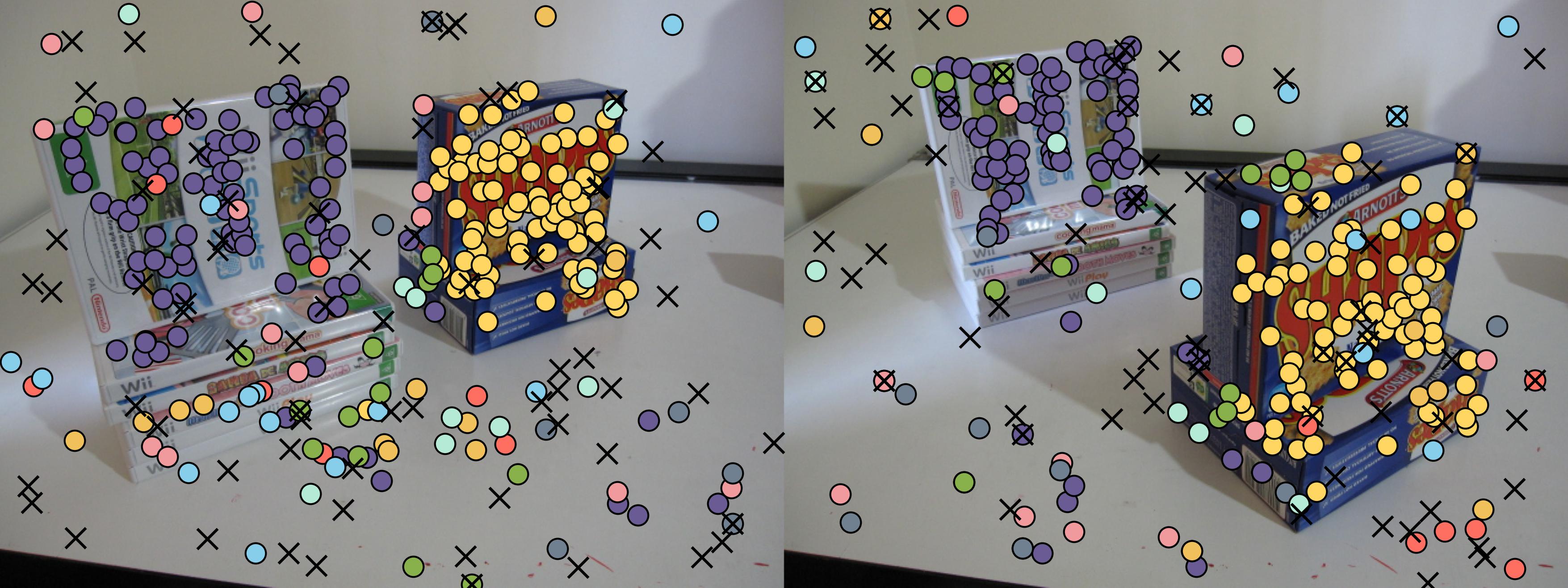}
            \caption{DeQuMF, $E_{mis} = 35.58\%$}
        \end{subfigure}
        \begin{subfigure}[b]{0.48\linewidth}
            \includegraphics[width=\linewidth]{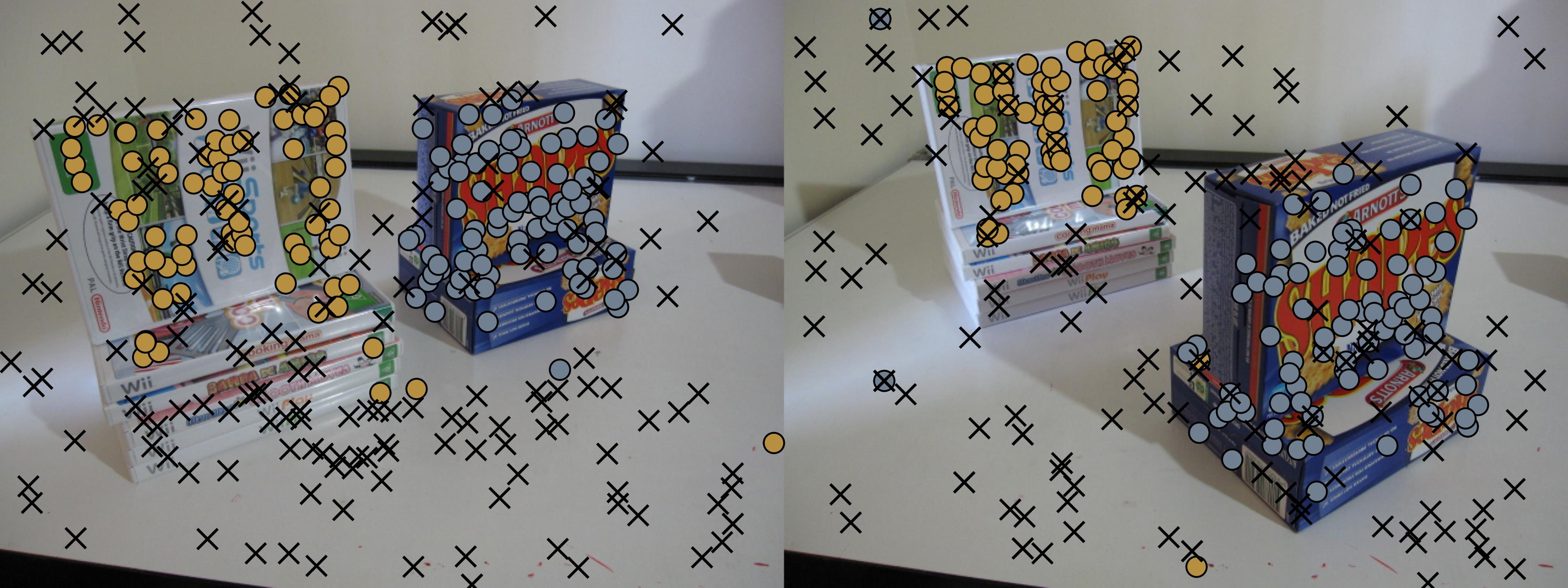}
            \caption{RQuMF, $E_{mis} = 6.12\%$}
        \end{subfigure}
        \begin{subfigure}[b]{0.48\linewidth}
            \includegraphics[width=\linewidth]{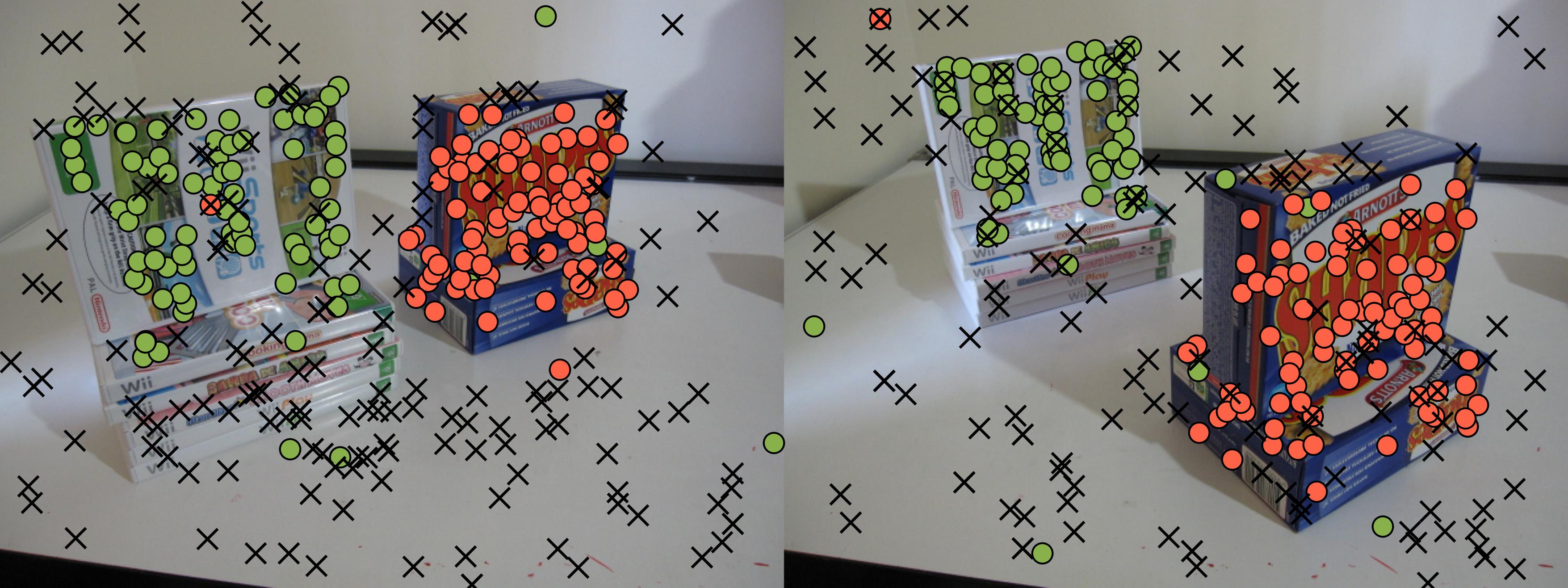}
            \caption{De-RQuMF, $E_{mis} = 6.95\%$}
        \end{subfigure}
        \begin{subfigure}[b]{0.48\linewidth}
            \includegraphics[width=\linewidth]{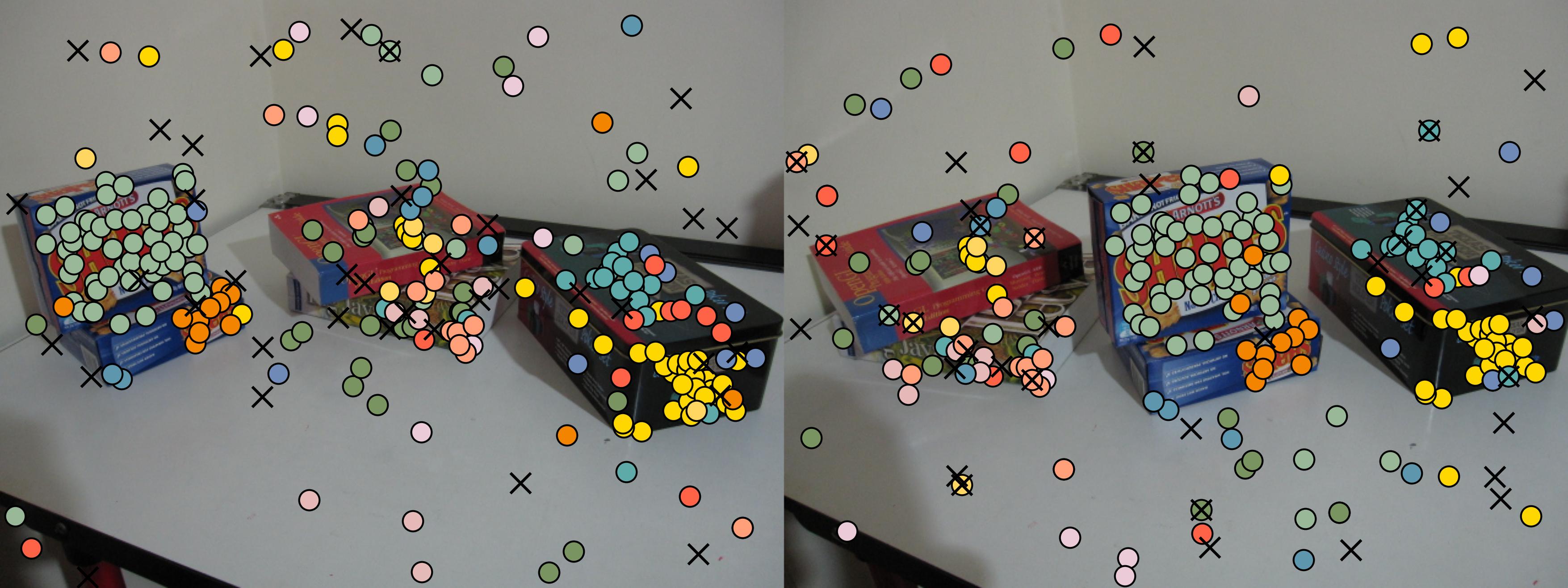}
            \caption{QuMF, $E_{mis} = 45.48\%$}
        \end{subfigure}
        \begin{subfigure}[b]{0.48\linewidth}
            \includegraphics[width=\linewidth]{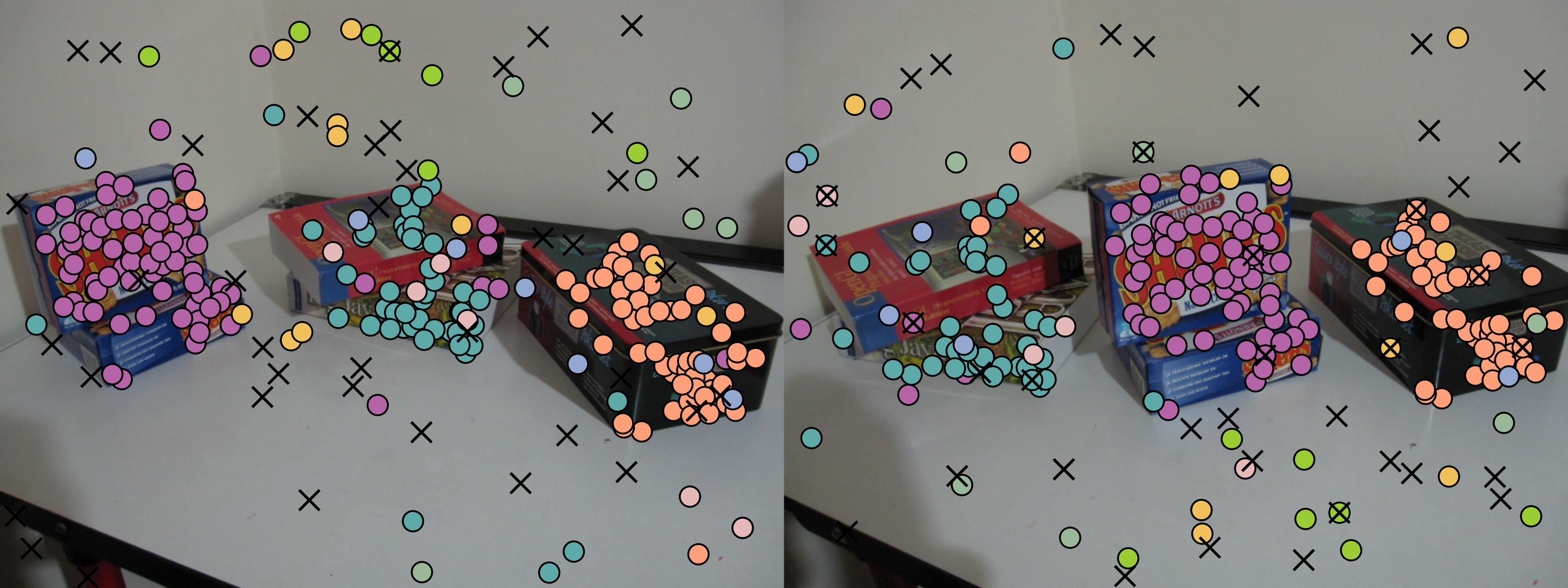}
            \caption{DeQuMF, $E_{mis} = 26.94\%$}
        \end{subfigure}
        \begin{subfigure}[b]{0.48\linewidth}
            \includegraphics[width=\linewidth]{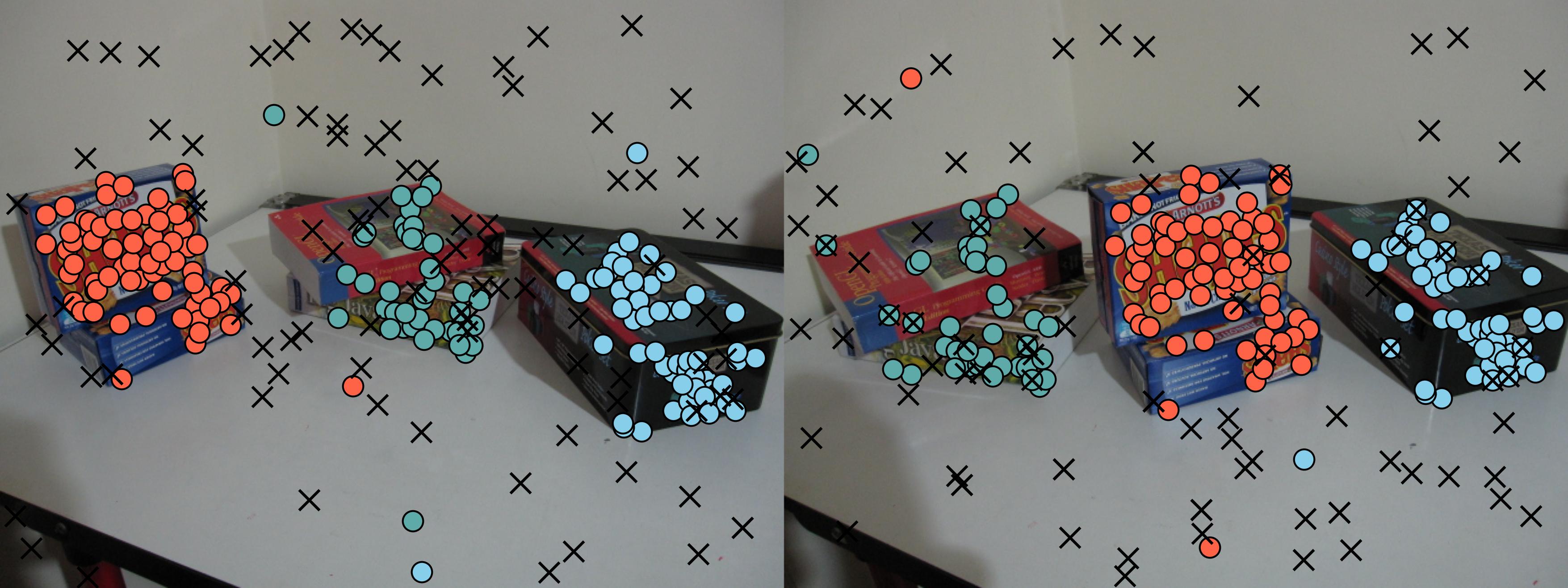}
            \caption{RQuMF, $E_{mis} = 4.86\%$}
        \end{subfigure}
        \begin{subfigure}[b]{0.48\linewidth}
            \includegraphics[width=\linewidth]{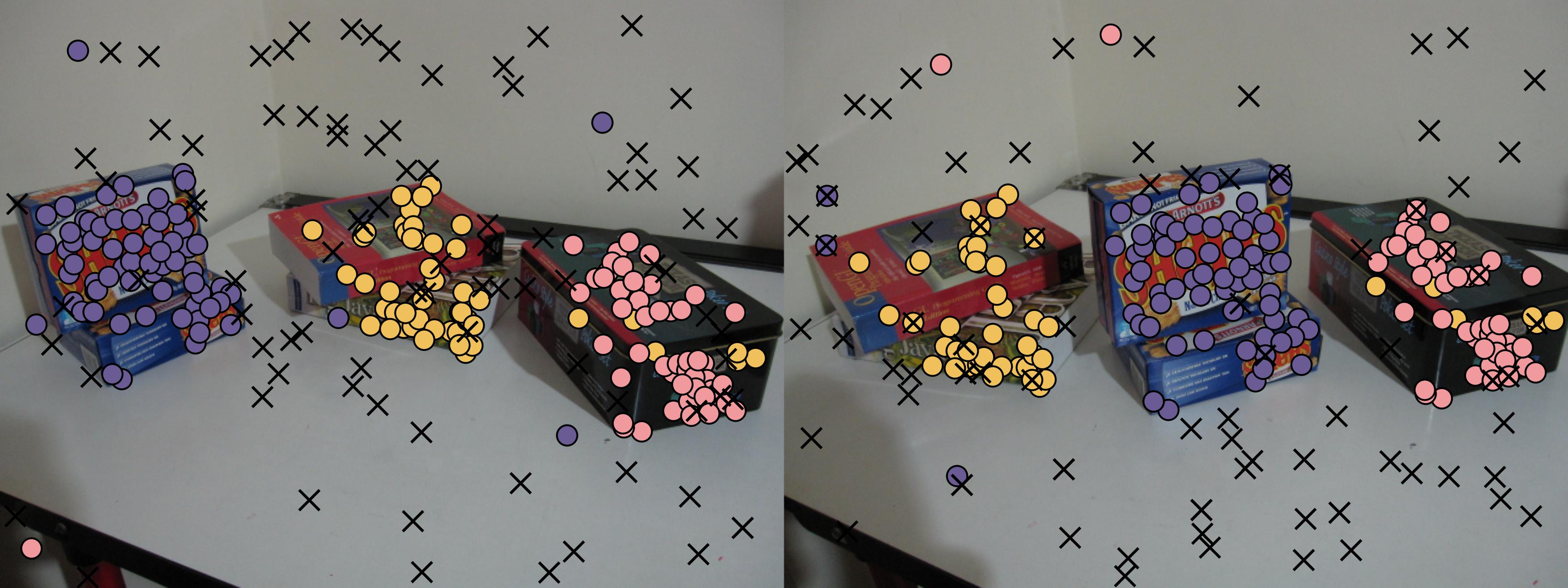}
            \caption{De-RQuMF, $E_{mis} = 8.20\%$}
        \end{subfigure}
        \begin{subfigure}[b]{0.48\linewidth}
            \includegraphics[width=\linewidth]{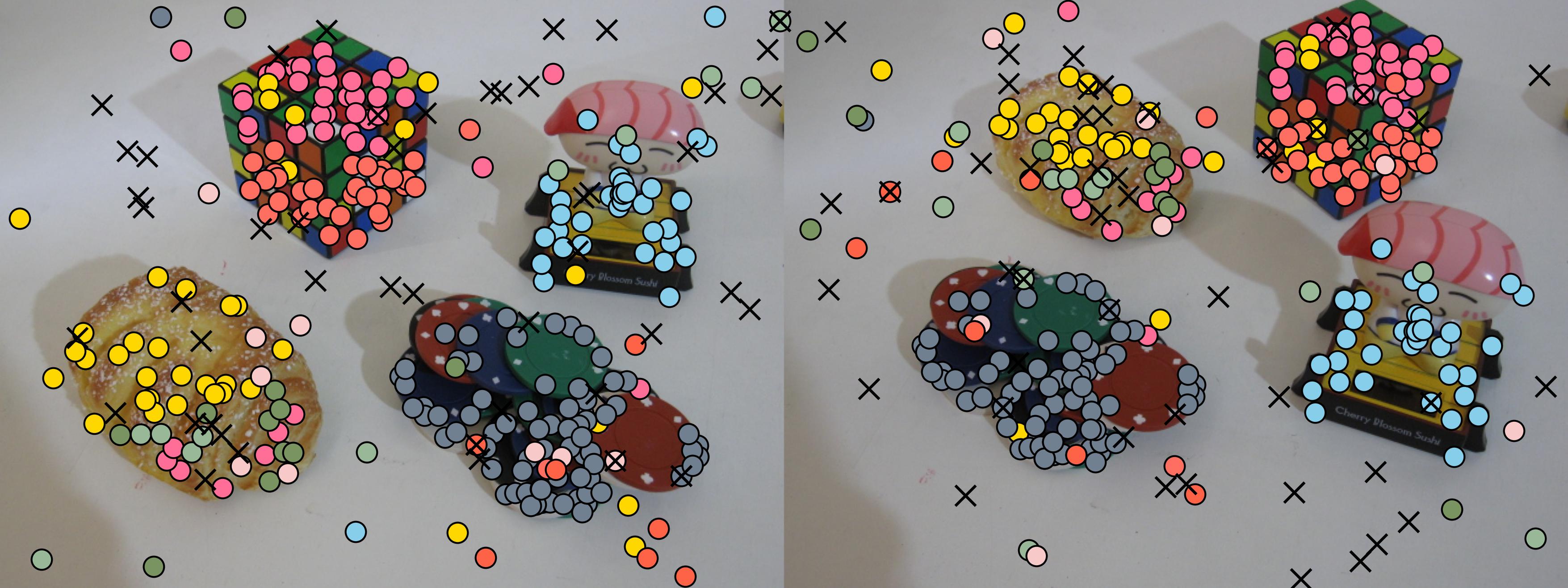}
            \caption{QuMF, $E_{mis} = 34.09\%$}
        \end{subfigure}
        \begin{subfigure}[b]{0.48\linewidth}
            \includegraphics[width=\linewidth]{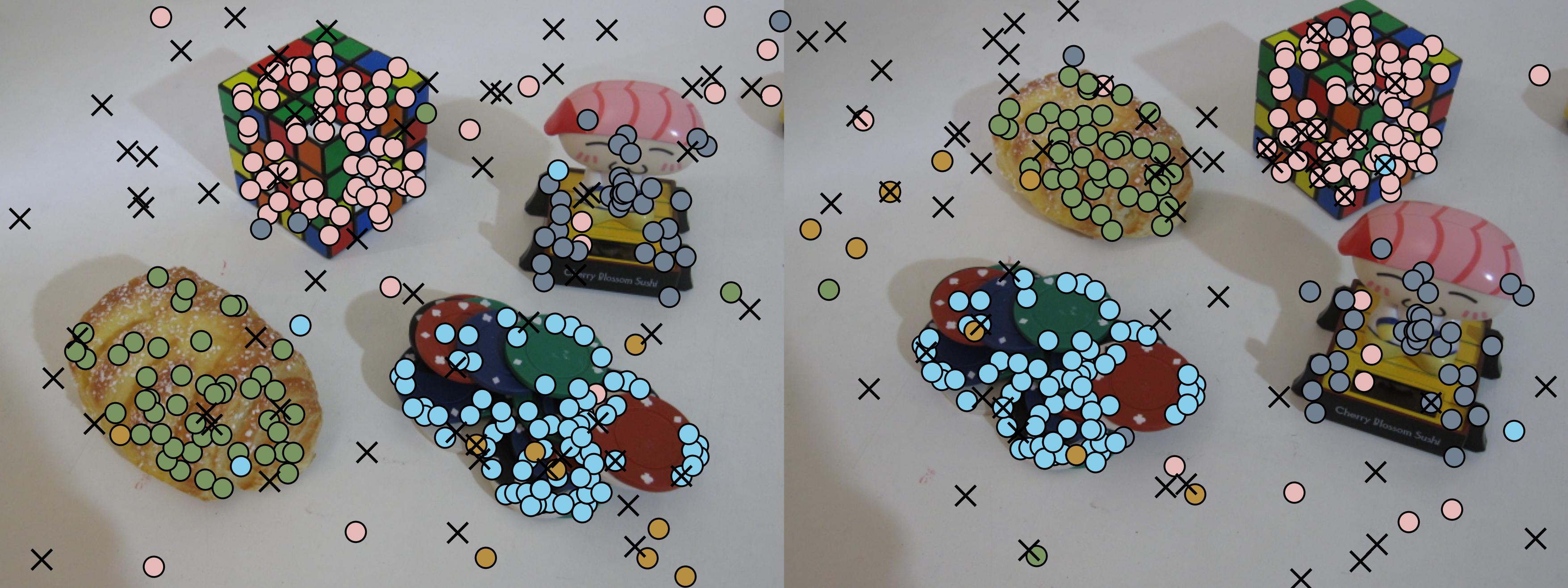}
            \caption{DeQuMF, $E_{mis} = 12.87\%$}
        \end{subfigure}
        \begin{subfigure}[b]{0.48\linewidth}
            \includegraphics[width=\linewidth]{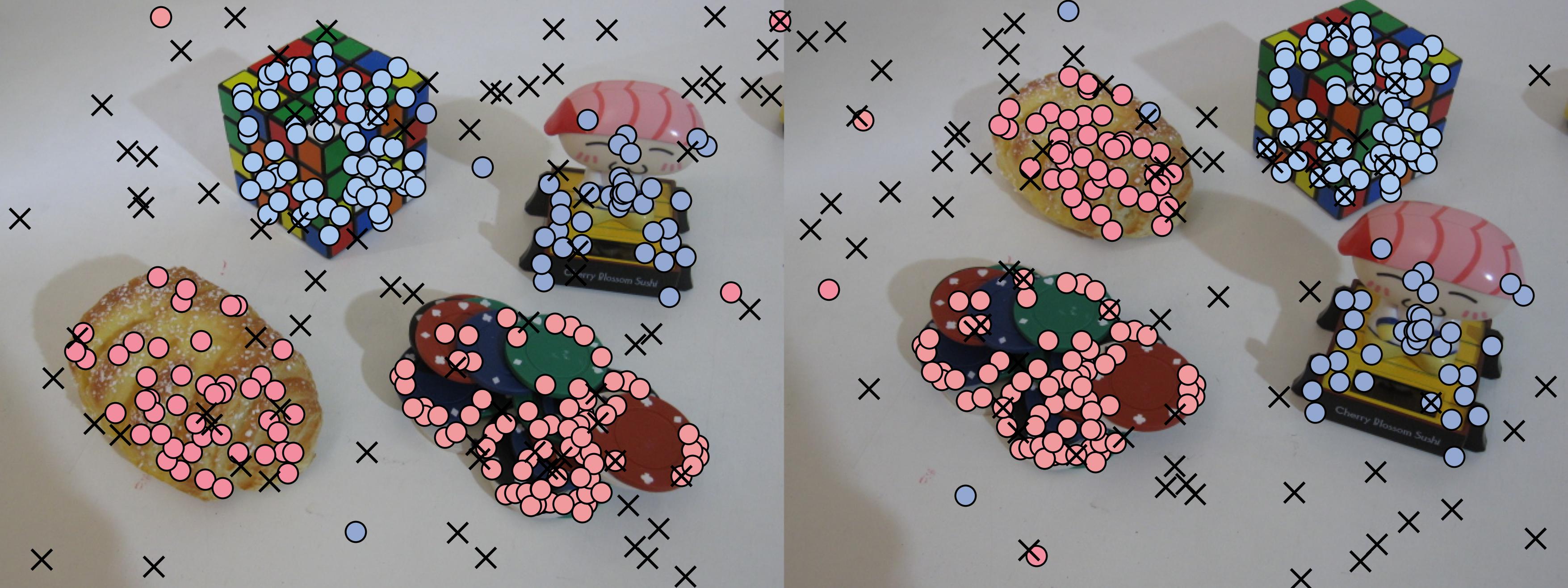}
           \caption{RQuMF, $E_{mis} = 7.79\%$}
        \end{subfigure}
        \begin{subfigure}[b]{0.48\linewidth}
            \includegraphics[width=\linewidth]{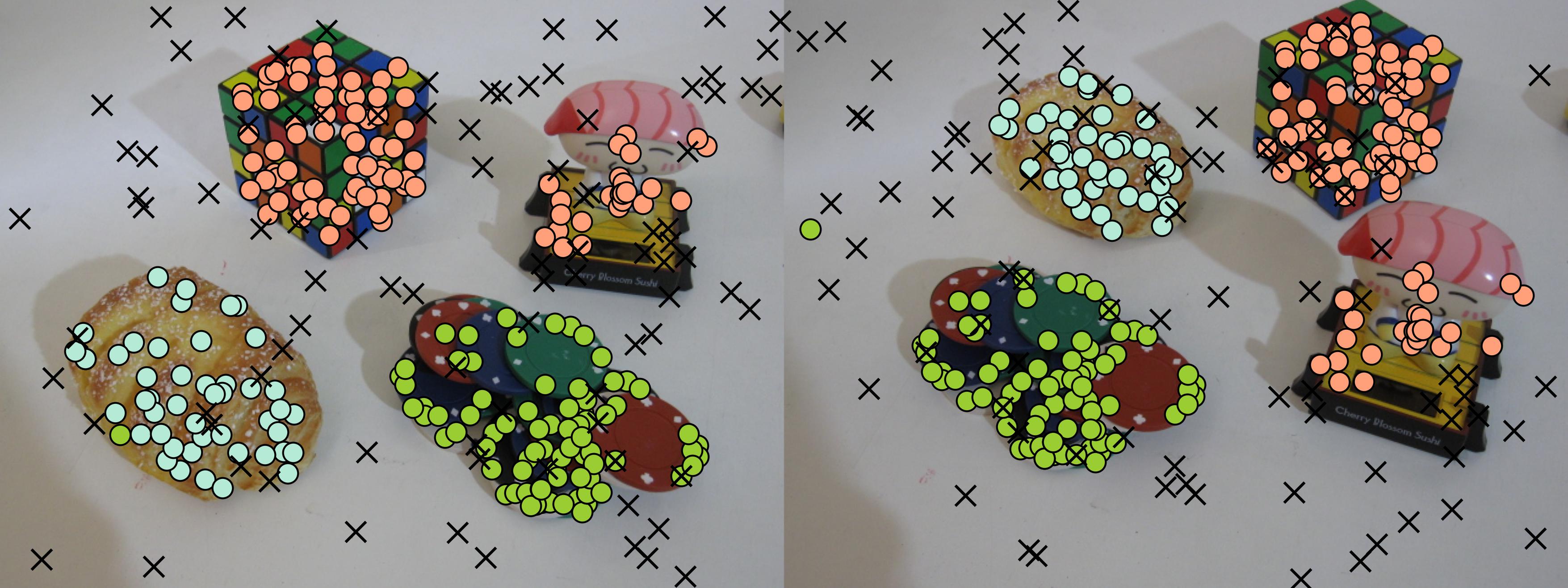}
            \caption{De-RQuMF, $E_{mis} = 16.91\%$}
        \end{subfigure}
        
        \caption{Average $E_{mis}$ on some samples of AdelaideRMF \cite{Wong2011} dataset for fundamental matrix fitting in the presence of outliers. one of our proposed methods outperforms the previous methods every time.}
        \label{fig:fm_visualization_grid_supp}
    \end{figure*}

\begin{figure*}[htbp!]
        \centering
        \begin{subfigure}[b]{0.48\linewidth}
            \includegraphics[width=\linewidth]{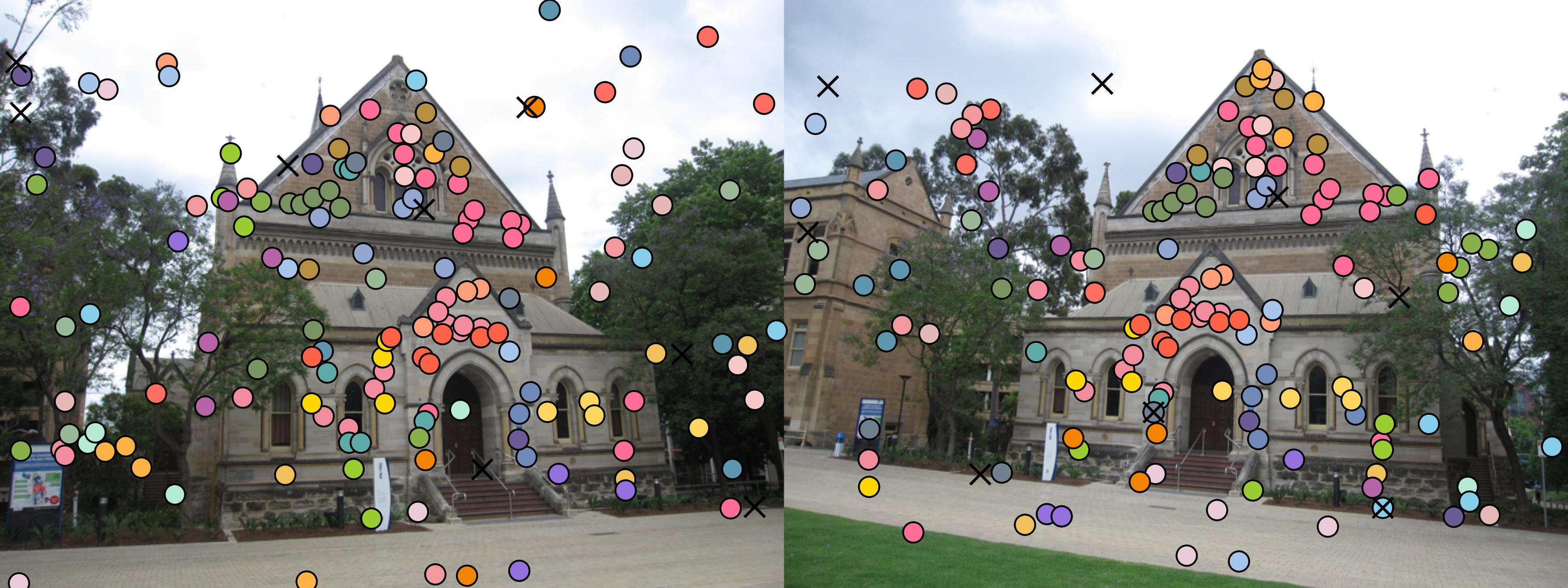}
            \caption{QuMF, $E_{mis} = 83.99\%$}
        \end{subfigure}
        \begin{subfigure}[b]{0.48\linewidth}
            \includegraphics[width=\linewidth]{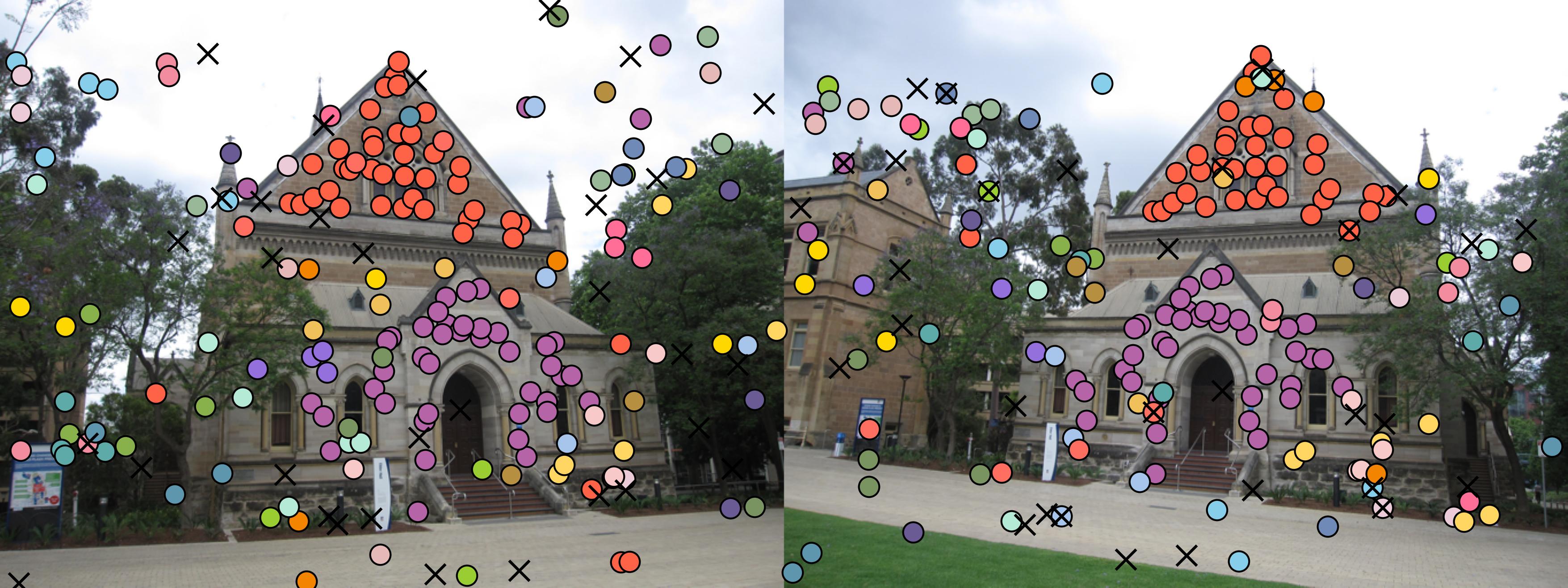}
            \caption{DeQuMF, $E_{mis} = 49.88\%$}
        \end{subfigure}
        \begin{subfigure}[b]{0.48\linewidth}
            \includegraphics[width=\linewidth]{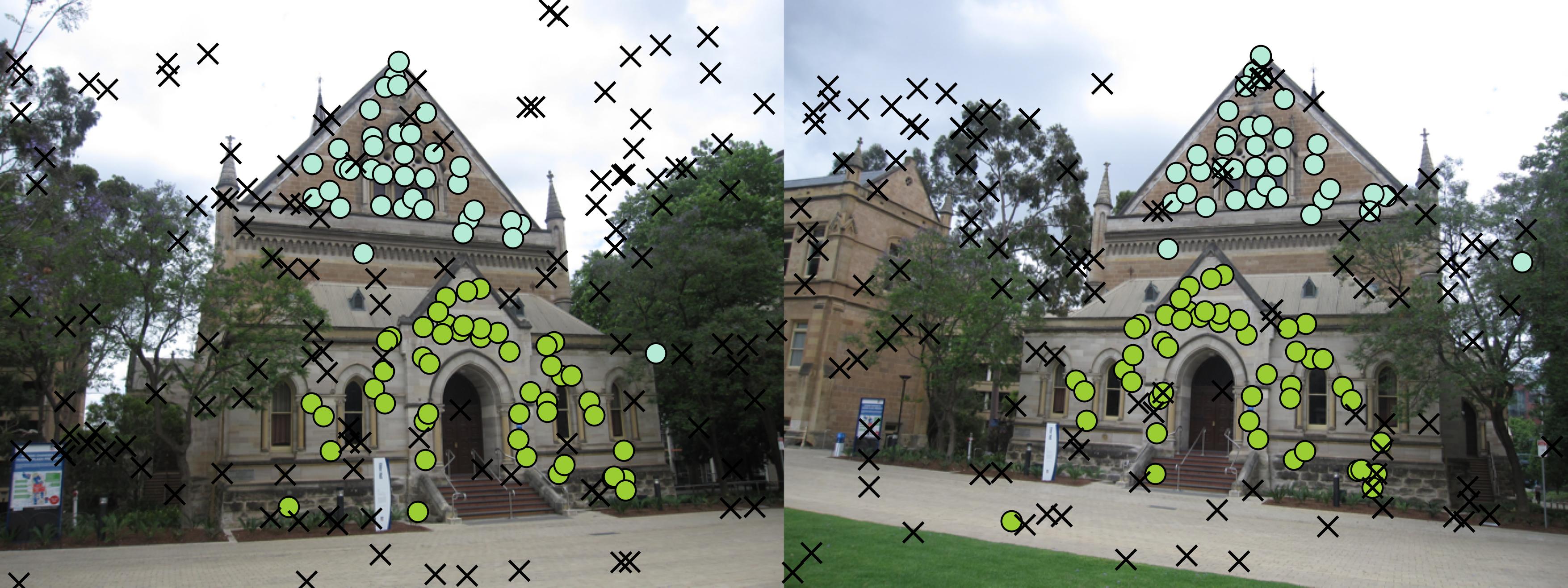}
            \caption{RQuMF, $E_{mis} = 2.86\%$}
        \end{subfigure}
        \begin{subfigure}[b]{0.48\linewidth}
            \includegraphics[width=\linewidth]{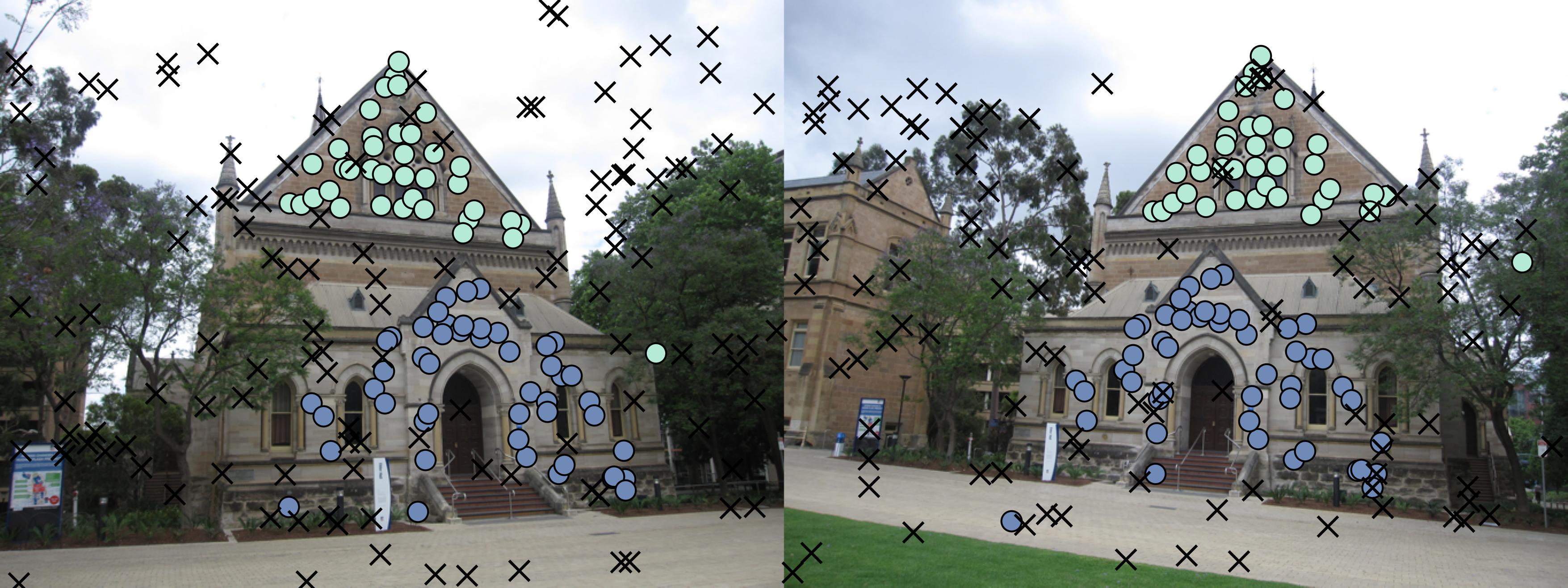}
            \caption{De-RQuMF, $E_{mis} = 2.80\%$}
        \end{subfigure}
        \begin{subfigure}[b]{0.48\linewidth}
            \includegraphics[width=\linewidth]{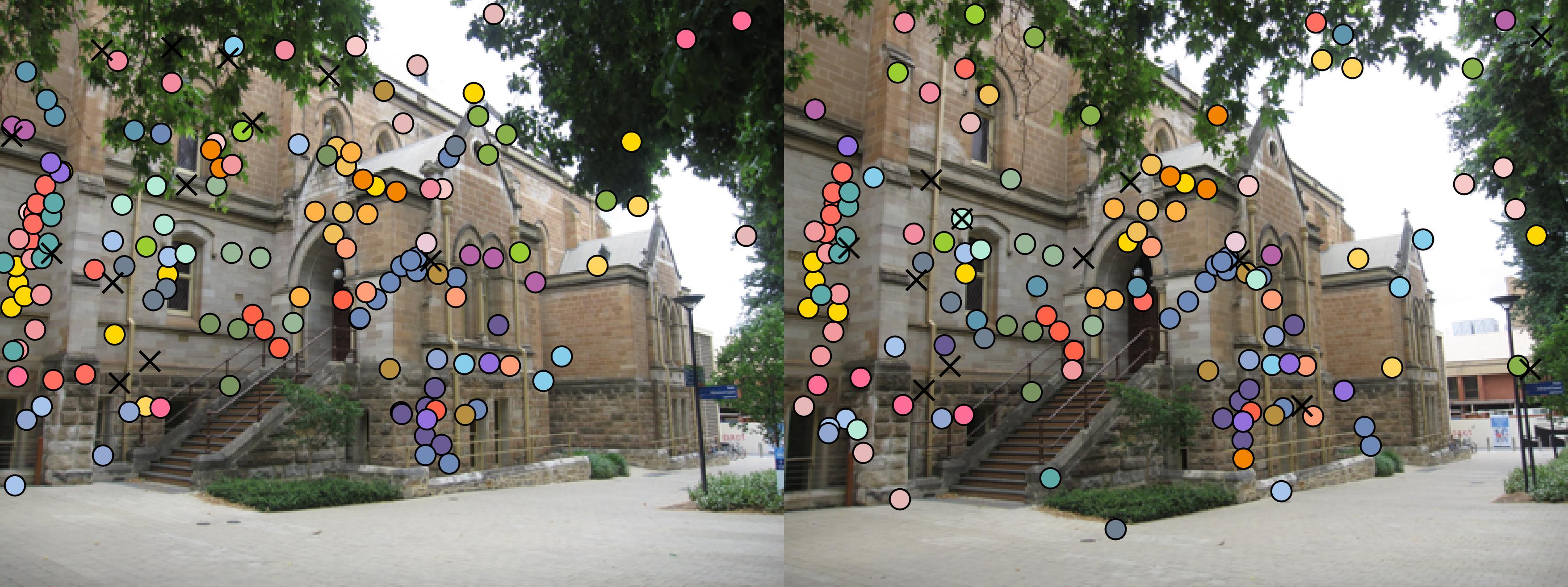}
            \caption{QuMF, $E_{mis} = 86.51\%$}
        \end{subfigure}
        \begin{subfigure}[b]{0.48\linewidth}
            \includegraphics[width=\linewidth]{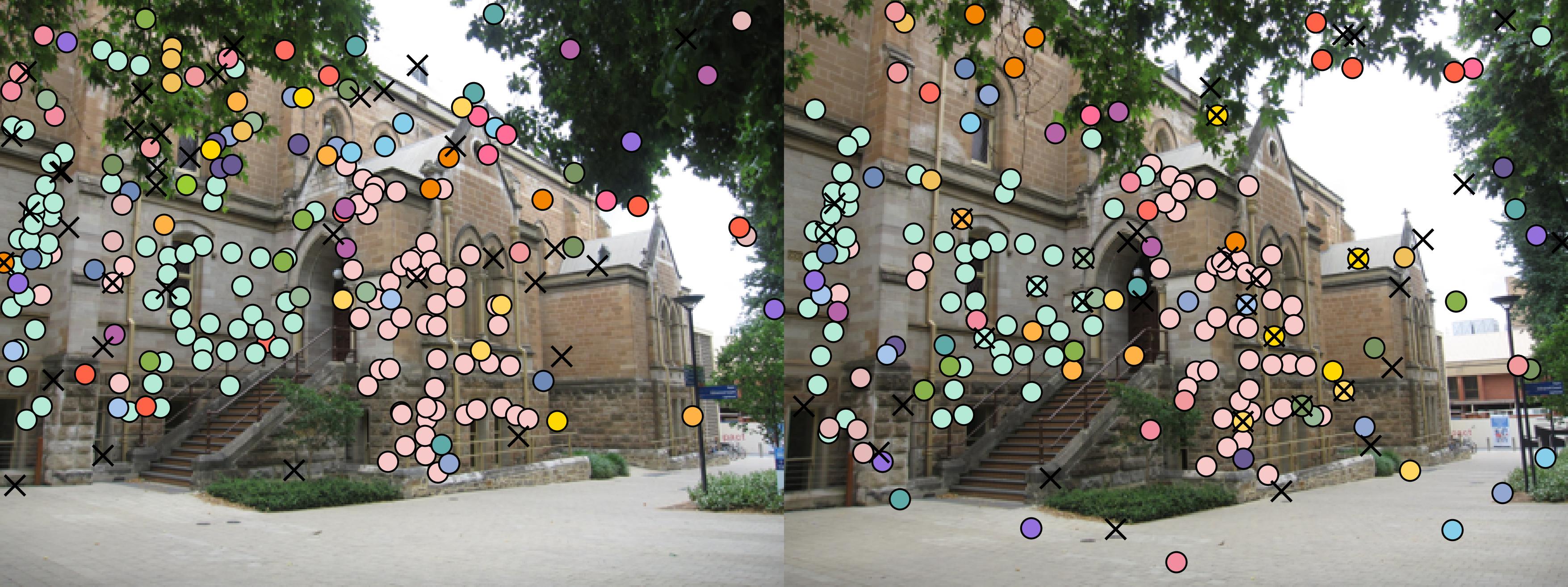}
            \caption{DeQuMF, $E_{mis} = 52.59\%$}
        \end{subfigure}
        \begin{subfigure}[b]{0.48\linewidth}
            \includegraphics[width=\linewidth]{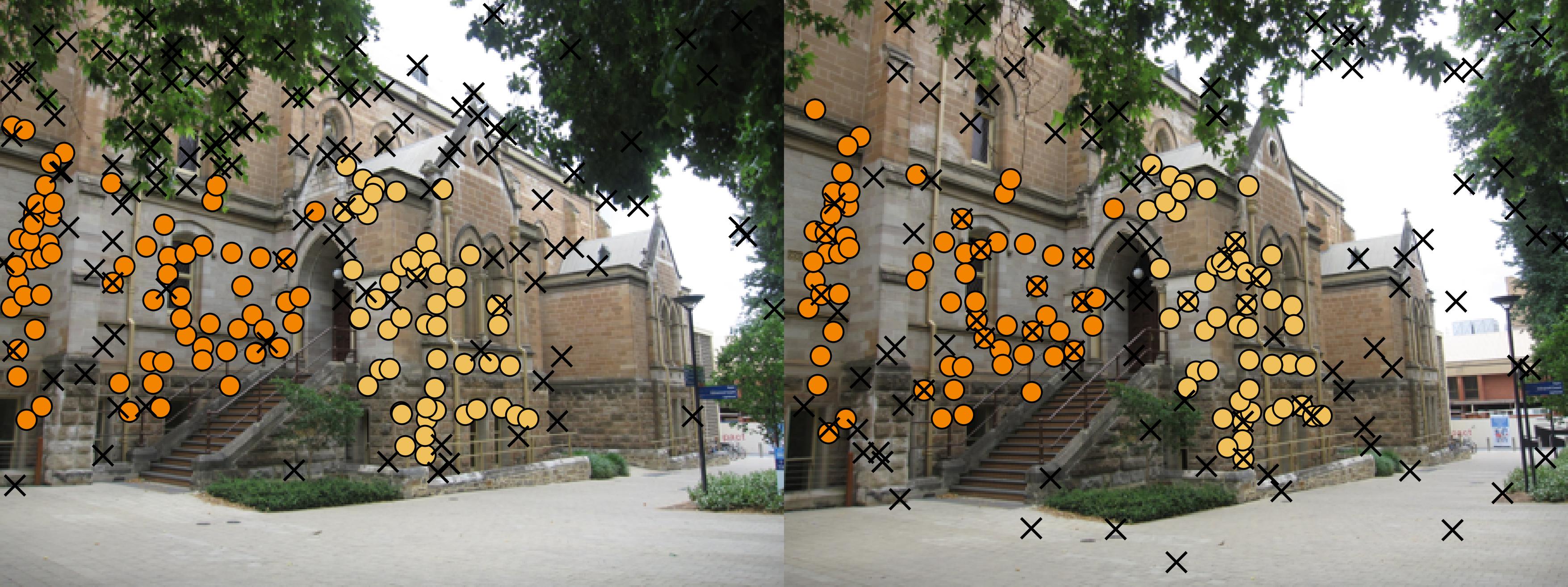}
            \caption{RQuMF, $E_{mis} = 20.53\%$}
        \end{subfigure}
        \begin{subfigure}[b]{0.48\linewidth}
            \includegraphics[width=\linewidth]{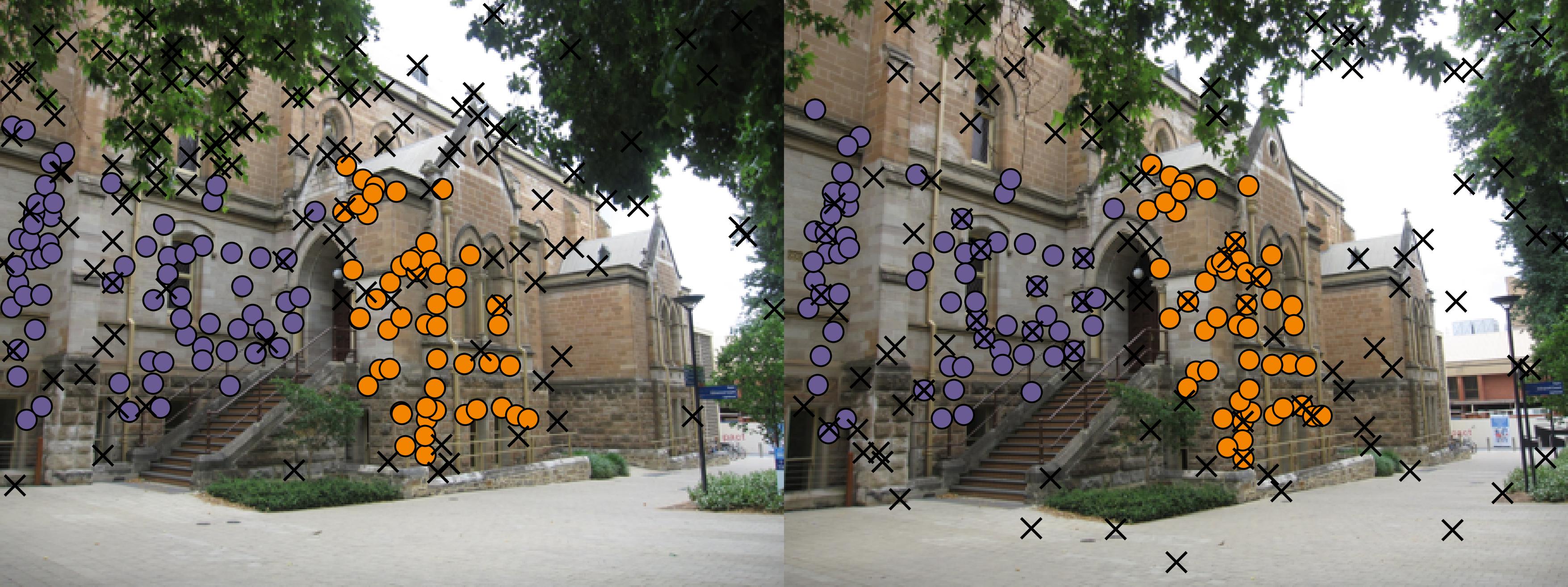}
            \caption{De-RQuMF, $E_{mis} = 24.75\%$}
        \end{subfigure}
        \begin{subfigure}[b]{0.48\linewidth}
            \includegraphics[width=\linewidth]{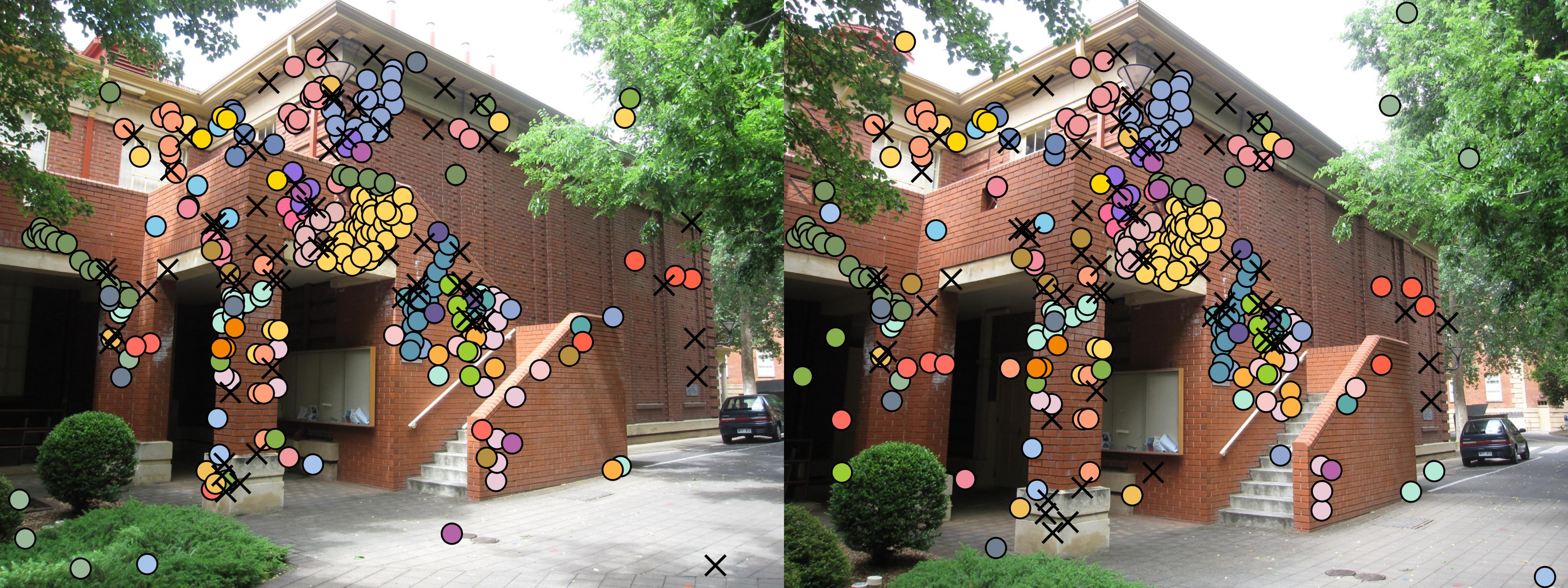}
            \caption{QuMF, $E_{mis} = 80.27\%$}
        \end{subfigure}
        \begin{subfigure}[b]{0.48\linewidth}
            \includegraphics[width=\linewidth]{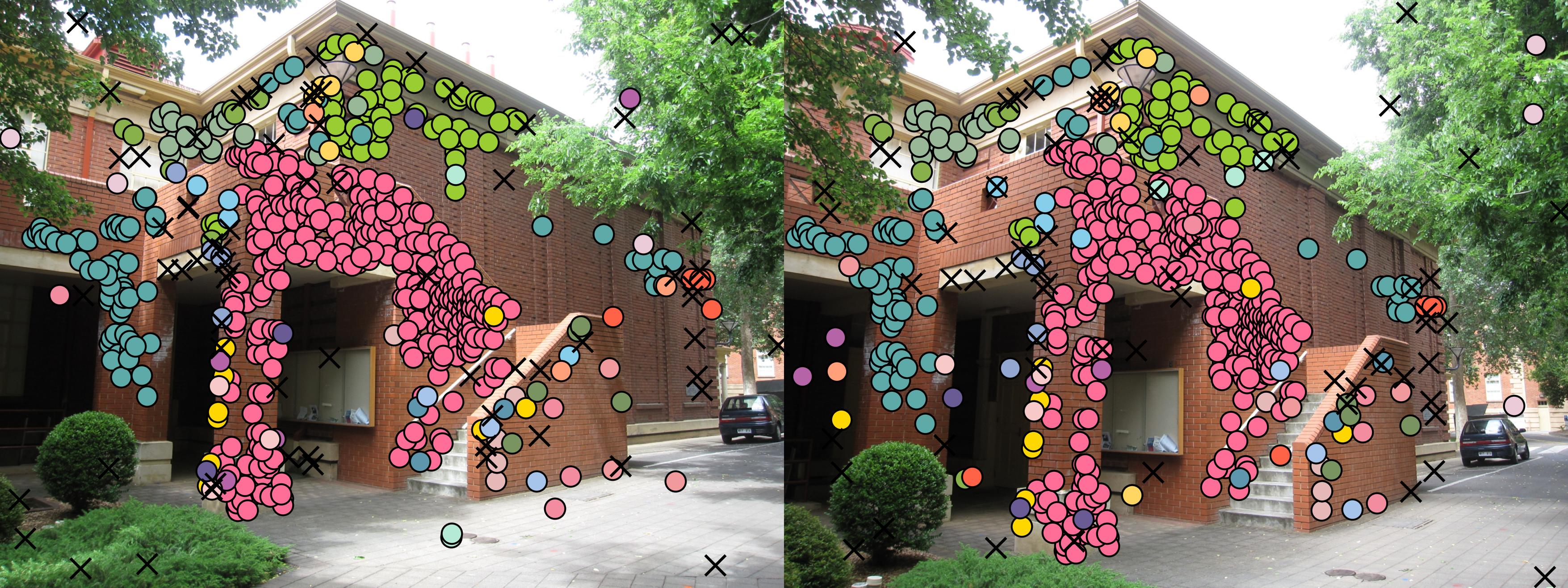}
            \caption{DeQuMF, $E_{mis} = 27.40\%$}
        \end{subfigure}
        \begin{subfigure}[b]{0.48\linewidth}
            \includegraphics[width=\linewidth]{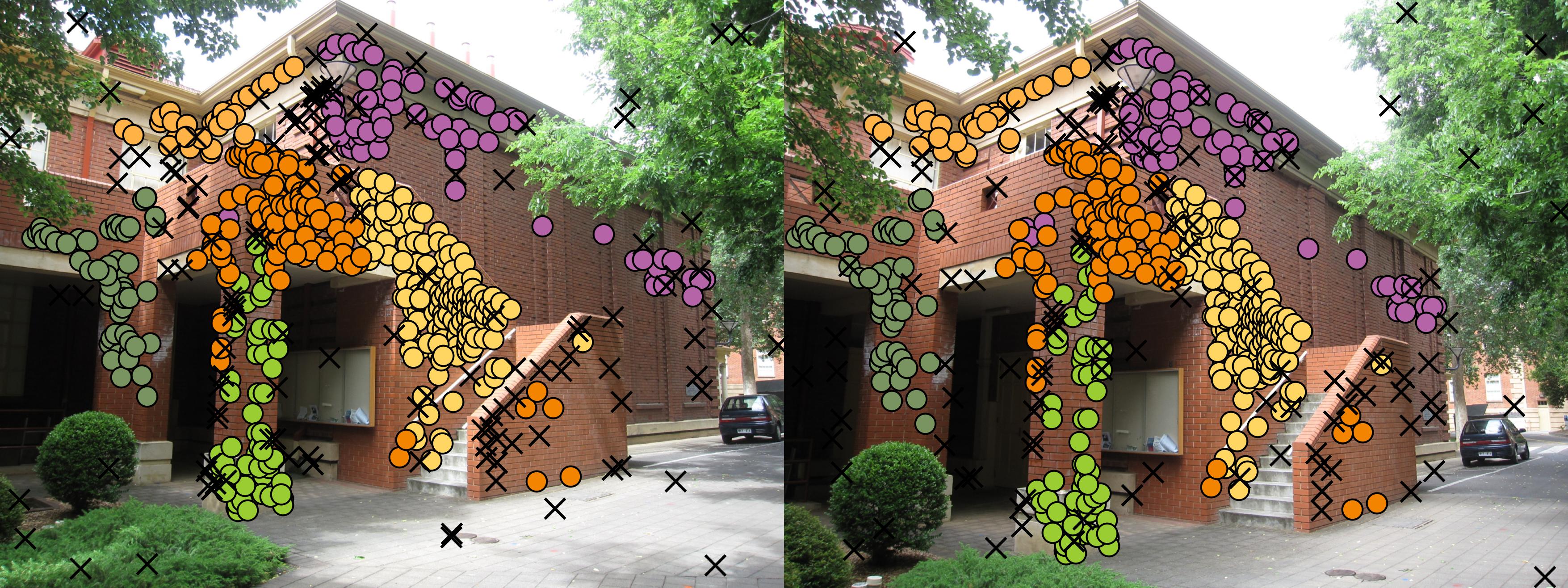}
            \caption{RQuMF, $E_{mis} = 35.67\%$}
        \end{subfigure}
        \begin{subfigure}[b]{0.48\linewidth}
            \includegraphics[width=\linewidth]{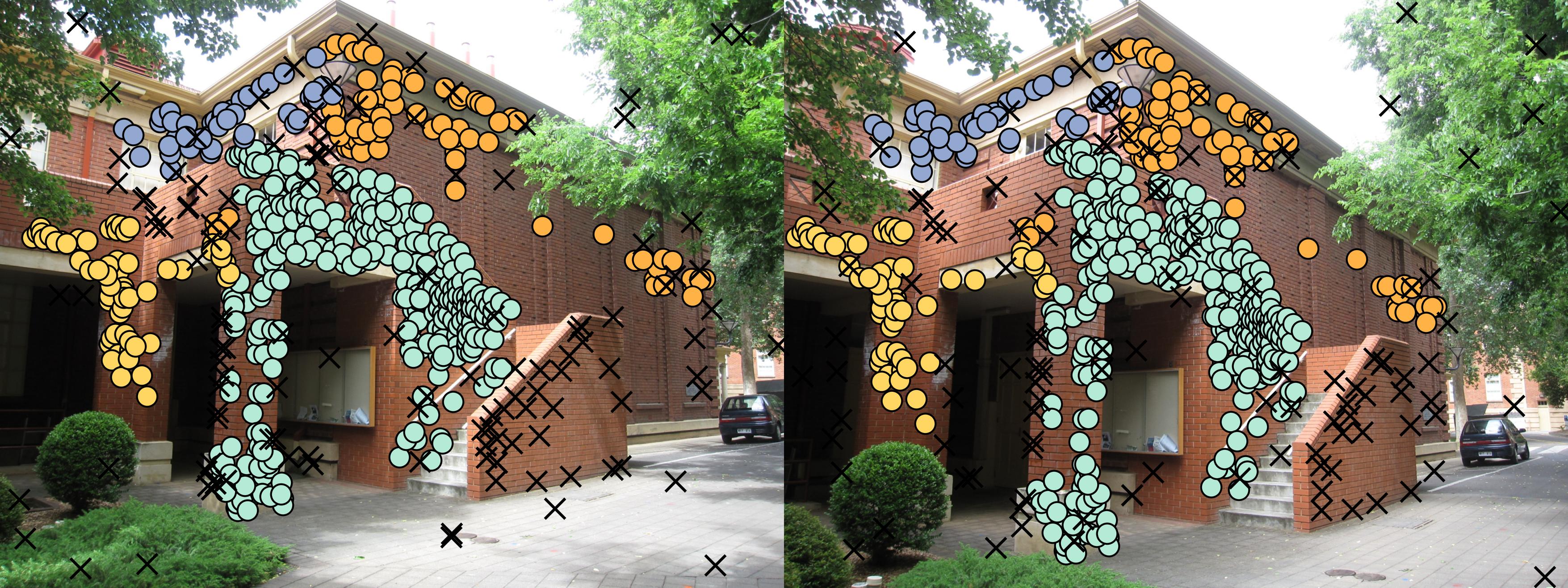}
            \caption{De-RQuMF, $E_{mis} = 22.49\%$}
        \end{subfigure}
        
        \caption{Average $E_{mis}$ on some samples of AdelaideRMF \cite{Wong2011} dataset for homography fitting in the presence of outliers. One of our proposed methods outperforms the previous methods every time in addition to being reliable.}
        \label{fig:hm_visualization_grid_supp}
    \end{figure*}

\paragraph{Qualitative Results.} We report the best and worst results for our RQuMF method for a few sequences of the Adelaide dataset in Fig.~\ref{fig:qualitative_supp1} and \ref{fig:qualitative_supp2}. Notably, our RQuMF method consistently surpasses the previous QuMF method, its non-decomposed counterpart, in performance across all scenarios, including both best and worst cases. Moreover, it excels beyond all other methods in three out of the four instances as depicted in Fig.~\ref{fig:qualitative_supp1} and ~\ref{fig:qualitative_supp2}. Additional visualizations are given in Fig.~\ref{fig:fm_visualization_grid_supp} and \ref{fig:hm_visualization_grid_supp}, confirming previous considerations.


\end{document}